\documentclass[letterpaper]{article}
\usepackage{uai2019}
\usepackage[margin=1in]{geometry}

\usepackage{times}

\usepackage{graphicx}
\usepackage{booktabs} 
\usepackage{amsmath}
\usepackage{amssymb}
\usepackage{amsthm}
\usepackage{mathrsfs}
\usepackage{color}
\usepackage{float} 
\usepackage{caption}
\usepackage{subcaption}

\usepackage[round]{natbib}

\newtheorem{theorem}{Theorem}
\newtheorem{lemma}{Lemma}
\newtheorem{claim}{Condition}
\newtheorem{corollary}{Corollary}

\newtheorem{assumption}{Assumption}

\graphicspath{{./figures/}}

\title{Towards a Better Understanding and Regularization \\of GAN Training Dynamics}

\author{ {Weili Nie} \\ Rice University \\
\texttt{wn8@rice.edu}
\And {Ankit B. Patel} \\ Rice University \& Baylor College of Medicine \\
\texttt{abp4@rice.edu}
} 

\begin{document}

\maketitle

\begin{abstract}
Generative adversarial networks (GANs) are notoriously difficult to train and the reasons underlying their (non-)convergence behaviors are still not completely understood. By first considering a simple yet representative GAN example, we mathematically analyze its local convergence behavior in a non-asymptotic way. Furthermore, the analysis is extended to general GANs under certain assumptions. We find that in order to ensure a good convergence rate, two factors of the Jacobian in the GAN training dynamics should be \textit{simultaneously} avoided, which are (i) the \textit{Phase Factor}, i.e., the Jacobian has complex eigenvalues with a large imaginary-to-real ratio, and (ii) the \textit{Conditioning Factor}, i.e., the Jacobian is ill-conditioned. Previous methods of regularizing the Jacobian can only alleviate one of these two factors, while making the other more severe. Thus we propose a new \textit{JA}cobian \textit{RE}gularization (JARE) for GANs, which simultaneously addresses both factors by construction. Finally, we conduct experiments that confirm our theoretical analysis and demonstrate the advantages of JARE over previous methods in stabilizing GANs.
\end{abstract}

\section{\MakeUppercase{Introduction} }

Generative adversarial networks (GANs) \citep{goodfellow2014generative} have achieved great success at generating realistic samples, with extensive applications \citep{ho2016generative,zhu2017unpaired,karras2019style}. The goal of GANs is to generate samples that are indistinguishable from real data and hence have essentially learned the underlying data distribution. However, they are notoriously difficult to train and as such many heuristics have been developed \citep{radford2015unsupervised, salimans2016improved, brock2018large}. 
Meanwhile, a lot of theoretical work has focused on stabilizing the GAN training by replacing the Jensen-Shannon (JS) divergence implicit in the vanilla GAN \citep{goodfellow2014generative} with alternative divergences, such as $f$-divergence (i.e. $f$-GAN) \citep{nowozin2016f} and Wasserstein distance (i.e. WGAN) \citep{arjovsky2017wasserstein}. Much of the related work has introduced various regularizers for better approximating these divergences \citep{gulrajani2017improved,roth2017stabilizing,miyato2018spectral}. 

But the training dynamics of GANs are still not completely understood. Typically, the training of GANs is achieved by solving a zero-sum game via simultaneous gradient descent (SimGD)
\citep{goodfellow2014generative,nowozin2016f,arjovsky2017wasserstein}.
The original work \citep{goodfellow2014generative} showed that SimGD converges to an equilibrium if the updates are made in the function space. In practice, with the generator and discriminator being parametrized by two distinct neural networks, the updates in the parameter space are no longer guaranteed to converge due to the highly non-convex properties of the loss surface \citep{goodfellow2016nips}.

In this work, we conduct a non-asymptotic analysis of local convergence in GAN training dynamics by evaluating the eigenvalues of its Jacobian near the equilibrium and analyzing the convergence rate. We first consider a simple yet representative GAN example, and then extend the analysis to the general GANs, where we find that the number of iterations needed to achieve $\epsilon$-error may be unexpectedly large due to the \textit{Phase Factor} (i.e., the Jacobian has complex eigenvalues with a large imaginary-to-real ratio) and \textit{Conditioning Factor} (i.e., the Jacobian is ill-conditioned) of the Jacobian. 
We later show that previous methods of regularizing the Jacobian 
can only alleviate one of these two factors, while making the impact of the other factor more severe. 
Based on our analysis, we propose a new \textit{JA}cobian \textit{RE}gularization (JARE) for GANs and show theoretically that it can alleviate these two factors simultaneously. Finally, experiments confirm our theoretical analysis and demonstrate the advantages of JARE over recently proposed methods. 

\section{\MakeUppercase{Related Work}}

\paragraph{Global convergence of GANs.}
By assuming the GAN objectives to be convex-concave, many works have provided the global convergence behaviors of GANs \citep{nowozin2016f, yadav2017stabilizing, gidel2018variational}. However, as shown in Section \ref{sec_simple_gan}, the convex-concave assumption is too unrealistic to hold true even in a simple GAN example. Also, \cite{li2018limitations} showed the global convergence of GANs by assuming a parametrized mixture of two Gaussians as the generator. Nevertheless, their theoretical results only work for GANs provided an optimal discriminator. These unrealistic assumptions together make an inevitably large gap between their theory and the actual training dynamics of GANs. 
Instead, we focus on the local convergence of GANs, which is a necessary condition of global convergence but more analytically tractable as it eschews such strong assumptions.

\paragraph{Local convergence of GANs.}
Recently, \cite{nagarajan2017gradient} showed that under some mild assumptions, the GAN dynamics are locally convergent. Furthermore, \cite{mescheder2018training} pointed out that if the assumptions in \cite{nagarajan2017gradient} are not satisfied,  in particular when data distributions are not absolutely continuous, the GAN dynamics are not always convergent, unless some regularization techniques are applied, such as zero-centered gradient penalties \citep{roth2017stabilizing} and consensus optimization (ConOpt) \citep{mescheder2017numerics}. 
However, these theoretical results are established in an asymptotic limit of vanishing step size where SimGD approximates a continuous-time dynamic system. In practice, we are more concerned about the characterization of the non-asymptotic convergence rate and the choice of the finite step size. 
This is because even though the continuous-time dynamic system is convergent, its discrete-time counterpart might still suffer from a poor convergence behavior. To this end, \cite{liang2018interaction} analyzed the non-asymptotic local convergence of GANs and revealed that the off-diagonal interaction term in the Jacobian can serve as both a blessing and a curse. Our theoretical results can serve a complementary to \cite{liang2018interaction} in terms of better understanding the local convergence of GANs.


\paragraph{General differentiable games. }
Another line of related work has focused on analyzing the general differentiable games with GANs being a specific use case \citep{balduzzi2018mechanics, daskalakis2018limit, letcher2018stable}. In particular,
\cite{balduzzi2018mechanics} decomposed the game dynamics into two components and proposed the Symplectic Gradient Adjustment (SGA) to find stable fixed points in general games. Interestingly, SGA shares some similarities with JARE in form although we are motivated from completely different perspectives. The major difference between SGA and JARE is that SGA needs an exclusive sign alignment during training which JARE does not require, and we argue that a better understanding and improvement of GAN dynamics should take the GAN properties into account, which is missing in this line of related work.

\vspace{-3pt}

\section{\MakeUppercase{Background}}
\label{back}
\vspace{-3pt}

\subsection{\MakeUppercase{GAN as a minimax game}}
\vspace{-3pt}

Despite many variants, the GAN is best described as a minimax game in which the two players, usually named the generator and discriminator, are maximizing and minimizing the same objective function, respectively. The GAN game can be formulated as follows:
\begin{align} \label{gan_form}
\begin{split}
&\min_{\phi}\max_{\theta } \; f(\phi, \theta) \\
f(\phi, \theta) \triangleq \mathbb{E}_{x \sim P_r}  & [g_1(D_{\theta}(x))]  + \mathbb{E}_{z \sim P_{0}}[g_2(D_{\theta}(G_{\phi}(z)))]
\end{split}
\end{align}
where  $\phi \in \Phi \subseteq \mathbb{R}^m$ and $\theta \in \Theta \subseteq \mathbb{R}^n$ denote the parameters of the generator $G_{\phi}: \mathcal{Z} \to \mathcal{X}$ and discriminator $D_{\theta}: \mathcal{X} \to \mathbb{R}$, respectively, $P_r$ and $P_0$ represent the true data distribution with support $\mathcal{X} \subseteq \mathbb{R}^d$ and latent distribution with support $\mathcal{Z}$. We also denote by $P_{\phi}$ the generated data distribution. Note that in our definition, the output of the discriminator $D_{\theta}$ is a real-valued logit rather than a probability. Therefore, by relating the objective in (\ref{gan_form}) to different $f$-divergences and Wasserstein distance between $P_r$ and $P_{\phi}$, 
$g_1, g_2: \mathbb{R} \to \mathbb{R}$
are both \textit{concave} functions, which is similar to \cite{nagarajan2017gradient}. For example, we can recover vanilla GAN  with $g_1(t) = g_2(-t) = -\log(1+e^{-t})$, WGAN with $g_1(t) = g_2(-t) = t$ and reverse Kullback-Leibler (KL) divergence in $f$-GAN with $g_1(t) = -e^{-t}, g_2(t) = 1-t$\footnote{Normally, WGAN requires the discriminator parameter space $\Theta$ to be an $K$-Lipschitz functional space while for $f$-divergences, we can simply set $\Theta = \mathbb{R}^n$.}. 

For training the minimax GAN game (\ref{gan_form}), SimGD is the most commonly used algorithm, in which the parameter updates are alternatively given as 
\begin{align} \label{simGD}
\begin{split}
\phi^{(k+1)} &= \phi^{(k)} - \eta \nabla_{\phi} f(\phi^{(k)}, \theta^{(k)})\\
\theta^{(k+1)} &= \theta^{(k)} + \eta \nabla_{\theta} f(\phi^{(k)}, \theta^{(k)})\\
\end{split}
\end{align}
where $\eta>0$ is the step size, $\phi^{(k)}$ and $\theta^{(k)}$ denote the corresponding parameters in the $k$-th iteration. 
Due to the non-convex properties of the GAN objective \citep{goodfellow2016nips}, it is difficult to analyze its global convergence in general. To gain key insights into the training instabilities in GANs, we focus on the local convergence of points near the equilibrium \citep{nagarajan2017gradient,mescheder2018training,mescheder2017numerics,heusel2017gans}. 


%
%
%
%

\vspace{-3pt}
\subsection{\MakeUppercase{Asymptotic vs. non-asymptotic convergence analysis}}
\vspace{-3pt}

The asymptotic convergence analysis is defined as applying the ``ordinary differential equation (ODE) method'' to analyze the convergence properties of dynamic systems. For example, consider a discrete-time system characterized by the gradient descent $v^{(t+1)} = v^{(t)} + \eta h(v^{(t)})$ for the gradient $h(\cdot) : \mathbb{R}^n \to \mathbb{R}^n$ and step size $\eta > 0$, the asymptotic convergence analysis assumes the step size $\eta$ is infinitely small such that the discrete-time system can be approximated by an ODE $\dot{v}(t) = h(v(t))$.
According to the Linearization Theorem \citep{arrowsmith1992dynamical}, if the Jacobian of the dynamic system $A \triangleq \frac{\partial h(v)}{\partial v}$ evaluated at a stationary point $v^*$ is Hurwitz, namely, $\text{Re} \{ \lambda_i(A) \} < 0, \forall i = 1, \cdots, n$, the equivalent ODE will converge to  $v^*$ for all points in its neighborhood. 

In the non-asymptotic convergence analysis, however, we consider the discrete system directly to obtain the number of iterations needed to achieve an $\epsilon$-error solution with a finite step size. Particularly, given the Jacobian $A$, to ensure the non-asymptotic convergence, we first provide an appropriate range of step size $\eta$ by solving the inequalities $| 1 + \lambda_i(A) | < 1, \forall i=1,\cdots,n$. Based on the constraint of the step size, we get the minimum value of $| 1 + \lambda_i(A) |$, and thus are able to evaluate the minimum number of iterations for an $\epsilon$-error solution, which characterizes the convergence rate. Therefore, the non-asymptotic analysis could more precisely reveal the convergence performance of the dynamic system than the asymptotic analysis.

\vspace*{-3pt}
\section{\MakeUppercase{A simple GAN example}} \label{sec_simple_gan}
\vspace*{-3pt}


For illustration, we first consider a simple GAN example, in which the true data distribution is an isotropic Gaussian with a nonzero mean, i.e. $x \sim \mathcal{N}(v, \sigma^2 I)$ where $x \in \mathbb{R}^n$ (assuming $d = n$) and latent distribution is also a Gaussian with the same shape but a zero mean, i.e. $z \sim \mathcal{N}(0, \sigma^2 I)$ where $z \in \mathbb{R}^n$. 
Basically, the problem becomes whether the generator could translate the latent Guassian to match the real Gaussian.
To this end, we can assume the generator and discriminator are both \textit{linear}, i.e. $D_{\theta}(x) = \theta^Tx$ (assuming $m=n$) and $G_{\phi}(z) = \phi+z$, which are both provably expressive enough to learn the true data distribution. 
Thus, the GAN game objective in (\ref{gan_form}) can be rewritten as
\begin{align} \label{obj_linear}
\begin{split}
f(\phi, \theta)  =& \mathbb{E}_{x \sim \mathcal{N}(v, \sigma^2 I)}[g_1({\theta}^Tx)] \\ 
&+ \mathbb{E}_{z \sim \mathcal{N}(0, \sigma^2 I)}[g_2({\theta}^T(\phi + z))]
\end{split}
\end{align}
It is easy to verify that the equilibrium exists, which is $(\phi^*, \theta^*) = (v, 0)$. 
Before proceeding to the analysis, we show that this simple GAN example is in fact a \textit{concave-concave} game, essentially different from the previous convex-concave assumption in GANs \citep{nowozin2016f, yadav2017stabilizing, gidel2018variational}.
\begin{lemma} \label{lem_concave_concave}
    The objective $f(\phi, \theta)$ in (\ref{obj_linear}) is concave-concave w.r.t. ($\phi, \theta$).
\end{lemma}
\emph{Proof:} See Appendix \ref{proof_lemma1}. \hfill $\square$

By considering a small open neighborhood of $(\phi^*, \theta^*)$ of radius $\delta$, denoted by $B_{\delta}(\phi^*, \theta^*)$, we introduce the local properties in this simple GAN example as follows.
\begin{lemma} \label{lemma2}
	The second-order derivative of $f(\phi, \theta)$ in (\ref{obj_linear}) w.r.t. $(\phi,\theta) \in B_{\delta}(\phi^*,\theta^*)$ is given by
		\begin{align} \label{hessian}
		    \begin{split}
		        &\nabla^2f(\phi,\theta) \triangleq \begin{bmatrix}
			\nabla^2_{\phi\phi}f(\phi,\theta) & \nabla^2_{\phi\theta}f(\phi,\theta) \\
			\nabla^2_{\theta\phi}f(\phi,\theta) & \nabla^2_{\theta\theta}f(\phi,\theta)
			\end{bmatrix} \\
			&\approx \begin{bmatrix}
				0 & g'_2(0)I \\
				g'_2(0)I  & \left(g_1''(0) + g_2''(0)\right) \left(\sigma^2 I+vv^T\right) 
			\end{bmatrix}
		    \end{split}
		\end{align}
\end{lemma}
\emph{Proof:} See Appendix \ref{proof_lemma2}. \hfill $\square$


Without loss of generality, we focus on the vanilla GAN objective, i.e. $g_1(t) = g_2(-t) = -\log(1+e^{-t})$, in the rest of the paper, since the analysis in general applies to different GAN objectives. 
To simplify notations, we let 
$w \triangleq (\phi - v, \theta)$
so the equilibrium becomes $w^{*} = 0$ and the SimGD updates in (\ref{simGD}) can be rewritten as 
\begin{align} \label{simGD_re}
\begin{split}
w^{(k+1)} &= w^{(k)} + \eta \tilde{\nabla} f(w^{(k)})\\
\end{split}
\end{align}
where $\tilde{\nabla}f(w^{(k)}) \triangleq \begin{bmatrix}
-{\nabla}_{\phi}f(w^{(k)}) \\ {\nabla}_{\theta} f(w^{(k)})
\end{bmatrix}$, and thus the Jacobian at $w^{(k)}$ is given by 
\begin{align*}
\small
    A(w^{(k)}) \triangleq {\frac{\partial \tilde{\nabla}f(w^{(k)})}{\partial w^{(k)}}}^T = \begin{bmatrix}
			-\nabla^2_{\phi\phi}f(w^{(k)}) & -\nabla^2_{\phi\theta}f(w^{(k)}) \\
			\nabla^2_{\theta\phi}f(w^{(k)}) & \nabla^2_{\theta\theta}f(w^{(k)})
    \end{bmatrix}
\end{align*}

In the next, we will replace $A(w^{(k)})$ by $A$ for brevity.

\begin{theorem} \label{thm_simgd}
	For any point within $B_{\delta}(w^*)$, the Jacobian $A$ in the simple vanilla GAN example trained via SimGD has the following eigenvalues: $\lambda_{1,2}(A) = \frac{-\sigma^2 \pm {\sqrt{\left(\sigma^2\right)^2-4}}}{4}$ and $
	\lambda_{3,4}(A) = \frac{-\beta^2 \pm {\sqrt{\left(\beta^2\right)^2-4}}}{4}$
	where $\beta^2 \triangleq \sigma^2+\|v\|^2$. 
\end{theorem}

\emph{Proof:} See Appendix \ref{proof_thm_simgd}. \hfill $\square$

The above theorem shows that $\text{Re}\{\lambda_{1,2}(A)\} < 0$ and $\text{Re} \{\lambda_{3,4}(A) \} < 0$, and thus the SimGD updates in this simple GAN example are asymptotically locally convergent, which is consistent with \cite{nagarajan2017gradient}. Next, we discuss lower bounds of the non-asymptotic convergence rate in two cases.

On the one hand, assuming the variance satisfies $0 < \sigma^2 < 2$, $\lambda_{1,2}(A)$ become complex-valued. Denote by $\zeta \triangleq \left| \frac{ \text{Im}\left\{\lambda_{1,2}(A)\right\} }{ \text{Re}\left\{\lambda_{1,2}(A)\right\} } \right|$ the absolute value of the imaginary-to-real ratio of $\lambda_{1,2}(A)$. 
The non-asymptotic convergence property determined by $\lambda_{1,2}(A)$ is given as follows. 
\begin{corollary} \label{coro1}
		To ensure non-asymptotic local convergence, the step size should satisfy $0 <  \eta < \frac{4}{\sqrt{1+\zeta^2}}$.
		The number of iterations to achieve an $\epsilon$-error solution satisfies $N \geq \frac{2 \log {\frac{C_0}{\epsilon }}}{\log (1+\frac{1}{\zeta^2})}$ where $C_0$ is a constant. Specifically, as $\zeta \to \infty$, $N$  will be at least $O\left(\zeta^2 \log \frac{1}{\epsilon}\right)$.
\end{corollary}
\textit{Proof:} See Appendix \ref{proof_coro1}. \hfill $\square$

It means when the absolute value of the imaginary-to-real ratio of $\lambda_{1,2}(A)$ increases, the number of iterations $N$ for a certain convergence performance increases (quadratically in the limit). Since we know $\zeta = \sqrt{(\frac{2}{\sigma^2})^2-1}$ in the simple vanilla GAN example, which is a monotonically decreasing function of $\sigma^2$, if we set $\sigma^2 = 0.01$ for instance, then 
$N \geq O(10^4 \log \frac{1}{\epsilon})$ which shows a quite slow convergence rate.

On the other hand, 
we assume $\beta^2 > 2$, then $\lambda_{3,4}(A)$ are real-valued. Without loss of generality, we assume $| \lambda_{3}(A) | \geq | \lambda_{4}(A) |$ and the absolute value of their ratio is denoted by $\tau \triangleq \left| \frac{  \lambda_{3}(A) }{ \lambda_{4}(A) }  \right|$. Thus, $\tau$ is a lower bound of the condition number of the Jacobian, and the non-asymptotic convergence property determined by $\lambda_{3,4}(A)$ is given as follows.
\begin{corollary} \label{coro2}
	To ensure non-asymptotic local convergence, the step size should also satisfy $0 < \eta < \frac{4}{\sqrt{\tau}}$.
	For $\tau > 2$, the number of iterations  $N$ to achieve an $\epsilon$-error solution satisfies $N > \frac{\log {\frac{\epsilon}{C_1}}}{\log {(1-\frac{2}{\tau})}}$ where $C_1$ is a constant. Specifically, as $\tau \to \infty$, $N$ will be at least $O(\tau \log \frac{1}{\epsilon})$. 
\end{corollary}
\textit{Proof:} See Appendix \ref{proof_coro2}. \hfill $\square$

It means when 
the absolute value of $\frac{\lambda_{3}(A)}{\lambda_{4}(A)}$ 
increases, the number of iterations $N$ for a certain convergence performance also increases (linearly in the limit). 
Since we know $\tau = \frac{1}{4}{( \beta^2 + {\sqrt{(\beta^2)^2-4}} )^2}$ in the simple vanilla GAN example, which is a monotonically increasing function of $\beta^2$, if we set 
$\|v\| = 10$ for instance, then 
$N \geq O(10^4 \log \frac{1}{\epsilon})$, which also implies a very poor convergence rate.


In summary, 
there may exist the following two factors of the Jacobian in the GAN dynamics \textit{simultaneously} (e.g., $0 < \sigma^2 < 2$ and $\beta^2 > 2$ in the simple GAN example) that result in the GAN training issues.
\begin{itemize}
	\item \textit{Phase Factor}: The Jacobian $A$ has complex eigenvalues with a large imaginary-to-real ratio, which has also been reported in \cite{mescheder2017numerics}.
	\item \textit{Conditioning Factor}: The Jacobian $A$ is ill-conditioned, i.e., the largest absolute value of its eigenvalues is much larger than the smallest one.
\end{itemize}

As we can see later in general GANs, it is the special nature of the Jacobian in GANs that makes the GAN training dynamics more unstable than other neural network optimization problems. 
In particular, Theorem \ref{thm_simgd} reveals that in the simple GAN example, both $\sigma^2$ and $\beta^2$ should not be too small or too large, which is a relatively strict requirement for local convergence. Furthermore, simply changing the expressive power of the generator or discriminator may not easily alleviate these two factors simultaneously. Please see Appendix \ref{app_example} for an example of changing the discriminator representations. 
Therefore, how to simultaneously alleviate these two factors we have identified above becomes an important question for the GAN training.


\vspace*{-3pt}
\section{\MakeUppercase{Jacobian Regularization}} \label{sec_gr}
\vspace*{-3pt}

A straightforward method to alleviate these two factors simultaneously is to introduce a \emph{regularization matrix} $\Gamma$ such that the training updates in (\ref{simGD_re}) become
\begin{align} \label{reg_update}
    w^{(k+1)} = w^{(k)} + \eta \Gamma \tilde{\nabla} f(w^{(k)}) 
\end{align}
and thus the (regularized) Jacobian is given by $A = \Gamma {\frac{\partial \tilde{\nabla}f(w^{(k)})}{\partial w^{(k)}}}^T$. The goal is to find a regularization matrix $\Gamma$ such that we can appropriately control the eigenvalues of the Jacobian for points near the equilibrium. 

\vspace*{-3pt}
\subsection{\MakeUppercase{Revisiting previous methods}}
\vspace*{-3pt}

There are several gradient-based regularization methods that have been proposed to deal with the training instabilities of GANs from the perspective of controlling the Jacobian， 
such as \textit{only regularizing generator} \citep{nagarajan2017gradient} and \textit{ConOpt} \citep{mescheder2017numerics}.

\paragraph{Only regularizing generator.}
To overcome the non-convergence issue of training WGAN via SimGD, \cite{nagarajan2017gradient} has proposed to only regularize the generator by using the gradient of the discriminator in a principled way. The regularized updates for the generator become
\begin{align*} 
\begin{split}
\phi^{(k+1)} &= \phi^{(k)} - \eta \nabla_{\phi} f(w^{(k)}) - \frac{1}{2} \eta\gamma \nabla_{\phi} \left\Vert \nabla_{\theta} f(w^{(k)}) \right\Vert ^2
\end{split}
\end{align*}
where the discriminator updates remain the same with SimGD, and thus the corresponding regularization matrix is 
$\Gamma = \begin{bmatrix}
I & -\gamma \nabla_{\phi\theta}^2 f(w^{(k)}) \\
0 & I
\end{bmatrix}$ with $\gamma$ being a tunable hyperparameter.

\paragraph{ConOpt. } 
By directly alleviating the impact of the \textit{Phase Factor}, \cite{mescheder2017numerics} has proposed ConOpt and its regularized updates are
\begin{align*} 
\begin{split}
w^{(k+1)} = w^{(k)} + \eta \tilde{\nabla} f(w^{(k)})  - \frac{1}{2}\eta\gamma \nabla \left\Vert \nabla f(w^{(k)}) \right\Vert ^2
\end{split}
\end{align*}
where the corresponding regularization matrix is 
$\Gamma = \begin{bmatrix}
I+\gamma \nabla_{\phi\phi}^2 f(w^{(k)}) & -\gamma \nabla_{\phi\theta}^2 f(w^{(k)})  \\
\gamma \nabla_{\theta\phi}^2 f(w^{(k)}) & I - \gamma \nabla_{\theta\theta}^2 f(w^{(k)})
\end{bmatrix}$.

\paragraph{Only regularizing discriminator. }
Similar to \textit{only regularizing generator} in \citep{nagarajan2017gradient}, a straightforward idea is to only regularize the discriminator instead by using the gradient of the generator and its regularized updates for the discriminator become
\begin{align*} 
\begin{split}
\theta^{(k+1)} &= \theta^{(k)} + \eta \nabla_{\theta} f(w^{(k)})  - \frac{1}{2} \eta\gamma \nabla_{\theta} \left\Vert \nabla_{\phi} f(w^{(k)}) \right\Vert ^2
\end{split}
\end{align*}
where the generator updates remain the same with SimGD, and thus the corresponding regularization matrix is 
$\Gamma = \begin{bmatrix}
I & 0 \\
\gamma \nabla_{\theta\phi}^2 f(w^{(k)}) & I
\end{bmatrix}$.

Their convergence behaviors in terms of stabilizing the simple vanilla GAN example (\ref{obj_linear}) are given as follows.

\begin{theorem} \label{thm_prev_meth}
    In the simple vanilla GAN example, none of the previous gradient-based regularization methods (i.e., only regularizing generator, ConOpt and only regularizing discriminator) are capable of simultaneously alleviating the \textit{Phase Factor} and \textit{Conditioning Factor}.
\end{theorem}

\textit{Proof:} See Appendix \ref{proof_thm_prev_meth}. \hfill $\square$

From the above theorem, together with the example of changing the representations in Appendix \ref{app_example}, we can see that without carefully taking into account both the \textit{Phase Factor} and \textit{Conditioning Factor}, these GAN variants might still suffer from the poor convergence even in the simple GAN example.

\vspace*{-3pt}
\subsection{\MakeUppercase{JARE}}
\vspace*{-3pt}


Based on the above theoretical analysis, we propose a new but simple Jacobian regularization, called JARE, which also applies the regularization terms based on the gradients of the generator and discriminator. Specifically, the regularized updates are given by
\begin{align} \label{proposed_update}
\begin{split}
\phi^{(k+1)} &= \phi^{(k)} - \eta \nabla_{\phi} f(w^{(k)}) - \frac{1}{2}\eta\gamma \nabla_{\phi} \left\Vert \nabla_{\theta} f(w^{(k)}) \right\Vert ^2\\
\theta^{(k+1)} &= \theta^{(k)} + \eta \nabla_{\theta} f(w^{(k)})  - \frac{1}{2}\eta\gamma \nabla_{\theta} \left\Vert \nabla_{\phi} f(w^{(k)}) \right\Vert ^2
\end{split}
\end{align}
Similarly, the corresponding regularization matrix is
$\Gamma = \begin{bmatrix}
I & -\gamma \nabla_{\phi\theta}^2 f(w^{(k)})  \\
\gamma \nabla_{\theta\phi}^2 f(w^{(k)}) & I
\end{bmatrix}$ with $\gamma > 0$ being a tunable hyperparameter.

Note that the key difference between JARE and ConOpt is that JARE does not introduce the Hessians $\nabla_{\phi\phi}^2 f(w^{(k)})$ and $\nabla_{\theta\theta}^2 f(w^{(k)})$ in the regularization matrix $\Gamma$.
Intuitively, a reason for not doing this is to avoid the risk of reversing the gradient flows, which may diverge the GAN training dynamics (see Appendix \ref{note_conopt} for a detailed explanation).
The following theorem shows the eigenvalues of the Jacobian in the simple vanilla GAN example trained via the proposed method.
\begin{theorem} \label{thm_prop}
    For any point within $B_{\delta}(w^*)$, the Jacobian $A$ in the simple vanilla GAN example trained via JARE has the following eigenvalues: 
    $\lambda_{1,2}({A}) = \frac{-\left(\sigma^2 + \gamma \right) \pm {\sqrt{\left(\sigma^2 + \gamma \right)^2- (\gamma^2 + 4) }}}{4}$ and $\lambda_{3,4}({A}) = \frac{-\left(\beta^2 + \gamma \right) \pm {\sqrt{\left(\beta^2 + \gamma \right)^2- (\gamma^2 + 4) }}}{4}$,
    where $\beta^2 \triangleq \sigma^2+\|v\|^2$.
\end{theorem}

\textit{Proof:} See Appendix \ref{proof_thm_prop}. \hfill $\square$

From the above theorem, given $0 < \sigma^2 < 2$ and $\beta^2 > 2$, we can evaluate both $
\zeta \triangleq \left| \frac{ \text{Im}\left\{\lambda_{1,2}(A)\right\} }{ \text{Re}\left\{\lambda_{1,2}(A)\right\} } \right|$ and $\tau \triangleq \left| \frac{  \lambda_{3}(A) }{ \lambda_{4}(A) } \right|$, two key variables that reflect the impact of the \textit{Phase Factor}
and \textit{Conditioning Factor}, respectively, and see how the tunable parameter $\gamma$ in JARE changes their values. The results are given in the following corollary. 

\begin{corollary} \label{coro_prop}
    In the simple vanilla GAN example trained via JARE, $\zeta$ monotonically decreases as $\gamma$ increases, and if $\gamma \geq 2$,  $\tau$ also monotonically decreases as $\gamma$ increases. In the limit of $\gamma \to \infty$, we get $\zeta \to 0$ (i.e., no complex eigenvalues) and $\tau \to 1$ (i.e., well conditioned). Therefore, we can make $\gamma$ large enough in JARE to alleviate the impact of the \textit{Phase Factor} and \textit{Conditioning Factor} simultaneously.
\end{corollary}

\textit{Proof.} See Appendix \ref{proof_coro_prop}. \hfill $\square$

As we know from Corollary \ref{coro1} and \ref{coro2}, if $\zeta \to 0$ and $\tau \to 1$, the non-asymptotic convergence rate will be increasingly improved. Therefore, the above corollary justifies that the proposed JARE will provide a good local convergence behavior by applying a reasonably large hyperparameter $\gamma$. 
However, we cannot make $\gamma$ arbitrarily large in JARE. According to the non-asymptotic analysis, the step size in JARE should satisfy $0 < \eta < \eta_{\max}$ where $\eta_{\max} \triangleq 8\min \{\frac{\gamma+\sigma^2}{\gamma^2+4}, \frac{1}{(\beta^2+\gamma)+ \sqrt{2\beta^2\gamma+(\beta^2)^2-4}} \}$. As we can see $\eta_{\text{max}}$ decreases with the increment of $\gamma$, and goes to 0 as $\gamma \to \infty$. So when $\gamma$ is sufficiently large, we have to make the step size infinitely small accordingly.

\vspace{-2pt}
\section{\MakeUppercase{Extensions to General GANs}}\label{sec_general}
\vspace{-2pt}

The above analysis is based on the simple GAN example, and here we can extend it to the more general GAN scenario. First, we show that the two factors identified above can also be significant issues in general GANs. Second, we show that JARE can alleviate both of these factors simultaneously in general GANs. 

For ease of analysis, we make the following assumption in terms of equilibrium point in general GANs \citep{nagarajan2017gradient, mescheder2018training}.

\begin{assumption}
 In equilibrium, the optimal generated distribution satisfies $p_{{\phi}^*} = p_r$, and the optimal discriminator satisfies $D_{{\theta}^*}(x) = 0$ for the local neighborhood of any $x \in \mathcal{X}$.
\end{assumption}

Basically, this assumption means that the generator is powerful enough to match the true data distribution in  equilibrium where the discriminator cannot distinguish the real and fake samples. 
In order to avoid trivial solutions, we also make an assumption as follows \citep{nagarajan2017gradient}.
\begin{assumption}
 The two concave functions $g_1$ and $g_2$ satisfy $g_1''(0) + g_2''(0) < 0$ and $g_1'(0) = -g_2'(0)  \neq 0$.
\end{assumption}
For example, the vanilla GAN loss and reverse KL divergence satisfy this assumption but the WGAN loss does not.
Under these two assumptions, the Jacobian of general GANs is given as follows:
\begin{lemma} \label{general_lemma_A}
    For an equilibrium point $(\phi^*, \theta^*)$ satisfying Assumptions 1 and 2, the Jacobian A in general GANs trained via SimGD can be written in the form
    \begin{align} \label{general_A}
        A = \begin{bmatrix}
				0 & -P \\
				P^T  & Q
			\end{bmatrix}
    \end{align}
    where $P \in \mathbb{R}^{m \times n}$ and $Q \in \mathbb{R}^{m \times m}$ are given by
    \begin{align}
    \begin{split}
        P =& g_2'(0) \mathbb{E}_{z \sim P_0}{ [\nabla_{\phi}G_{\phi}(z) \nabla^2_{x \theta}D_{\theta}(x)}] |_{x = G_{\phi}(z)} \\
        Q =& (g''_1(0) + g''_2(0)) \mathbb{E}_{x \sim P_r}[\nabla_{\theta}D_{\theta}(x) \nabla_{\theta}D_{\theta}(x)^T] \\
    \end{split}
    \end{align}
\end{lemma}

\emph{Proof:} See Appendix \ref{proof_general_lemma}.
\hfill $\square$

The off-diagonal matrix $P$ represents how sensitive the discriminator is to the generator's local updates. The diagonal matrix $Q$ represents the local geometry of the discriminator. It is easy to verify that the Jacobian $A$ in the simple GAN example is a special case of Lemma~\ref{general_lemma_A}. Note that for WGAN, since $g''_1(0) = g''_2(0) = 0$, we have $Q = 0$ and therefore it is not even asymptotically convergent \citep{nagarajan2017gradient}, rendering a convergence rate analysis irrelevant. We are now ready to compute the eigenvalues of the Jacobian $A$ in general GANs.

\begin{theorem} \label{thm_general_simgd}
    For the equilibrium point $(\phi^*, \theta^*)$ satisfying Assumptions 1 and 2, the eigenvalues of the Jacobian $A$ in general GANs trained via SimGD can be written in the form 
    \begin{align} \label{lam_A_general}
        \lambda(A) = \frac{a_1 \pm \sqrt{a_1^2-4a_2}}{2}
    \end{align}
    where $a_1$ and $a_2$ are certain convex combinations of the eigenvalues of Q and $P^T P$, respectively. That is,
    \begin{align} \label{a1_a2}
        a_1 = \sum_{i=1}^m \alpha_i \lambda_i(Q), \quad 
        a_2 = \sum_{i=1}^m \tilde{\alpha}_i \lambda_i(P^T P)
    \end{align}
    for some coefficients $\alpha_i \geq 0$ with $\sum_{i=1}^m \alpha_i = 1$  and some coefficients $\tilde{\alpha}_i \geq 0$ with $\sum_{i=1}^m \tilde{\alpha}_i = 1$.
\end{theorem}

\emph{Proof:} See Appendix \ref{proof_thm_general_simgd}.
\hfill $\square$

Let $\lambda_{\text{min}}(\cdot)$ and $\lambda_{\text{max}}(\cdot)$ denote the minimum and maximum eigenvalues of a square matrix, respectively. From Lemma \ref{general_lemma_A} we know $Q \preceq 0$, so $\lambda_{\text{min}}(Q) \leq \lambda_i(Q) \leq \lambda_{\text{max}}(Q) \leq 0$, $\forall i$. Also, by definition we have $P^T P \succeq 0$, so $0 \leq \lambda_{\text{min}}(P^T P) \leq \lambda_i(P^T P) \leq \lambda_{\text{max}}(P^T P)$, $\forall i$. The convex combination in Eq. (\ref{a1_a2}) then implies
\begin{align*}
    \lambda_{\text{min}}(Q) \leq & a_1 \leq  \lambda_{\text{max}}(Q) \leq 0 \\
    0 \leq \lambda_{\text{min}}(P^T P) \leq & a_2 \leq \lambda_{\text{max}}(P^T P)
\end{align*} 
Therefore, how to balance the eigenvalue distributions of $Q$ and $P^T P$ plays an essential role in 
determining the eigenvalues of the Jacobian $A$. To see this, we consider two relatively extreme cases as follows:

First, if the maximum absolute value of the eigenvalues of $Q$ is much smaller than the minimum absolute value of the eigenvalues of $P^T P$, in particular we assume $Q$ and $P^T P$ satisfy $c |\lambda_{\text{min}}(Q)|^2 = 4 |\lambda_{\text{min}}(P^T P)|$ with $c \gg 1$, then for any coefficients $\alpha_i$ and $\tilde{\alpha}_i$ in Theorem \ref{thm_general_simgd}, we have $a_1^2 < 4 a_2$ and thus $\lambda(A)$ is complex-valued with the imaginary-to-real ratio satisfying $\zeta = \sqrt{c - 1}$. We can see that as $c$ becomes larger, the impact of \textit{Phase Factor} will be more severe.

Second, if the minimum absolute value of the eigenvalues of $Q$ is much larger than the maximum absolute value of the eigenvalues of $P^T P$, in particular we assume $Q$ and $P^T P$ in some GAN scenario satisfy $c' |\lambda_{\text{max}}(Q)|^2 = 4 |\lambda_{\text{max}}(P^T P)|$ with $c' \ll 1$, then for any coefficients $\alpha_i$ and $\tilde{\alpha}_i$ in Theorem \ref{thm_general_simgd}, we have $a_1^2 > 4 a_2$ and
thus $\lambda(A)$ is real-valued with the absolute value of eigenvalue ratio satisfying $\tau = (\sqrt{\frac{1}{c'}} + \sqrt{\frac{1}{c'} - 1})^2$. We can observe that when $c'$ is smaller, the impact of \textit{Conditioning Factor} will also be increasingly severe.

\begin{table}[t]
    \small
    \centering
    \begin{tabular}{l|cc}
        \hline
         Requirements & stable SimGD &  stable JARE \\
         \hline
         $Q$ is well-conditioned & \textbf{\textcolor{red}{\checkmark}} &  \\
         $P^TP$ is well-conditioned & \textbf{\textcolor{red}{\checkmark}} & \textbf{\textcolor{red}{\checkmark}} \\
         $Q$ matches $P^TP$ & \textbf{\textcolor{red}{\checkmark}} & \\
         \hline
    \end{tabular}
    \vspace{2mm}
    \caption{ The general requirements of ensuring a good local convergence behavior in both the GAN trained via SimGD (called ``stable SimGD'') and the GAN trained via the proposed JARE (called ``stable JARE''). The more requirements that the GAN training needs, the more difficult to ensure a good local convergence behavior.}
    \label{req}
\end{table}

Therefore, even if $Q$ and $P^T P$ themselves are both well-conditioned with all real eigenvalues, there still exist either the \textit{Phase Factor} or  \textit{Conditioning Factor} in the Jacobian $A$ due to the imbalance between their eigenvalue distributions. More generally in real GANs, it is also likely that either $Q$ or $P^T P$ becomes ill-conditioned, the GAN training dynamics will suffer more from the coexistence of \textit{Phase Factor} and the \textit{Conditioning Factor}. Note that the simple vanilla GAN example in Section \ref{sec_simple_gan} is just an illustrative special case of general GANs.

In summary, we need to make sure that $Q$ and $P^T P$ are both well-conditioned (which requires a well-designed generator and discriminator) \textit{and} have similar eigenvalues (which requires the discriminator to well match the generator) to avoid these two factors in general GANs.
Generally, these requirements are difficult to satisfy, which explains why GANs are hard to train and also why they are so sensitive to network architectures and other hyperparameters. 

Next, we compute the eigenvalues of the Jacobian $A$ in general GANs trained with our proposed JARE.

\begin{theorem} \label{thm_general_jr}
    For the equilibrium point $(\phi^*, \theta^*)$ satisfying Assumptions 1 and 2, the eigenvalues of the Jacobian $A$ in general GANs trained via JARE satisfy that in the limit $\gamma \to \infty$,
    \begin{align} \label{lam_A_JR}
        \lambda(A) = -\gamma \lambda(P^T P)
    \end{align}
\end{theorem}
\emph{Proof:} See Appendix \ref{proof_thm_general_jr}.
\hfill $\square$

We can see that $\lambda(A)$ in (\ref{lam_A_JR}) is real-valued, which means there is no \textit{Phase Factor} any more with a sufficiently large regularization term $\gamma$ in JARE. Also, the eigenvalue distribution of the Jacobian $A$ now \textit{only} depends on $P^T P$, which means in general GANs trained via JARE, the imbalance between the eigenvalue distributions of $Q$ and $P^T P$ will not result in an undesirable properties of the Jacobian $A$. Instead, we only need to make sure $P^T P$ is well-conditioned
to achieve a good convergence behavior. 
The comparison between GANs trained via SimGD and JARE regarding the requirements of good training dynamics is illustrated in Table \ref{req}.
In this sense, JARE will be significantly easier to train, with greater stability and more robustness to different network architectures and hyperparameters.

\vspace*{-3pt}
\section{\MakeUppercase{Experiments}}
\vspace*{-3pt}


\textbf{Isotropic Gaussian.}
First, we empirically verify our theory in the simple vanilla GAN example.
Specifically, we consider a two-dimensional case, i.e. $n=2$ and the mean of true data is $v = {\begin{bmatrix}
	0, \mu
\end{bmatrix}}^T$. To test the local convergence, the parameters of both the discriminator and generator are initialized within $B_{\delta}(w^*)$ where $\delta = 0.05$. For hyperparameters, we set the learning rate to be $\eta = 0.001$, the regularization parameter to be $\gamma = 10$, the optimizer to be stochastic gradient descent (SGD) with a batch size 128, and run 15K iterations. 

\begin{figure} [!t]
	\centering
	\begin{subfigure}[b]{0.36\textwidth}
		\centering
	    \includegraphics[width=\textwidth]{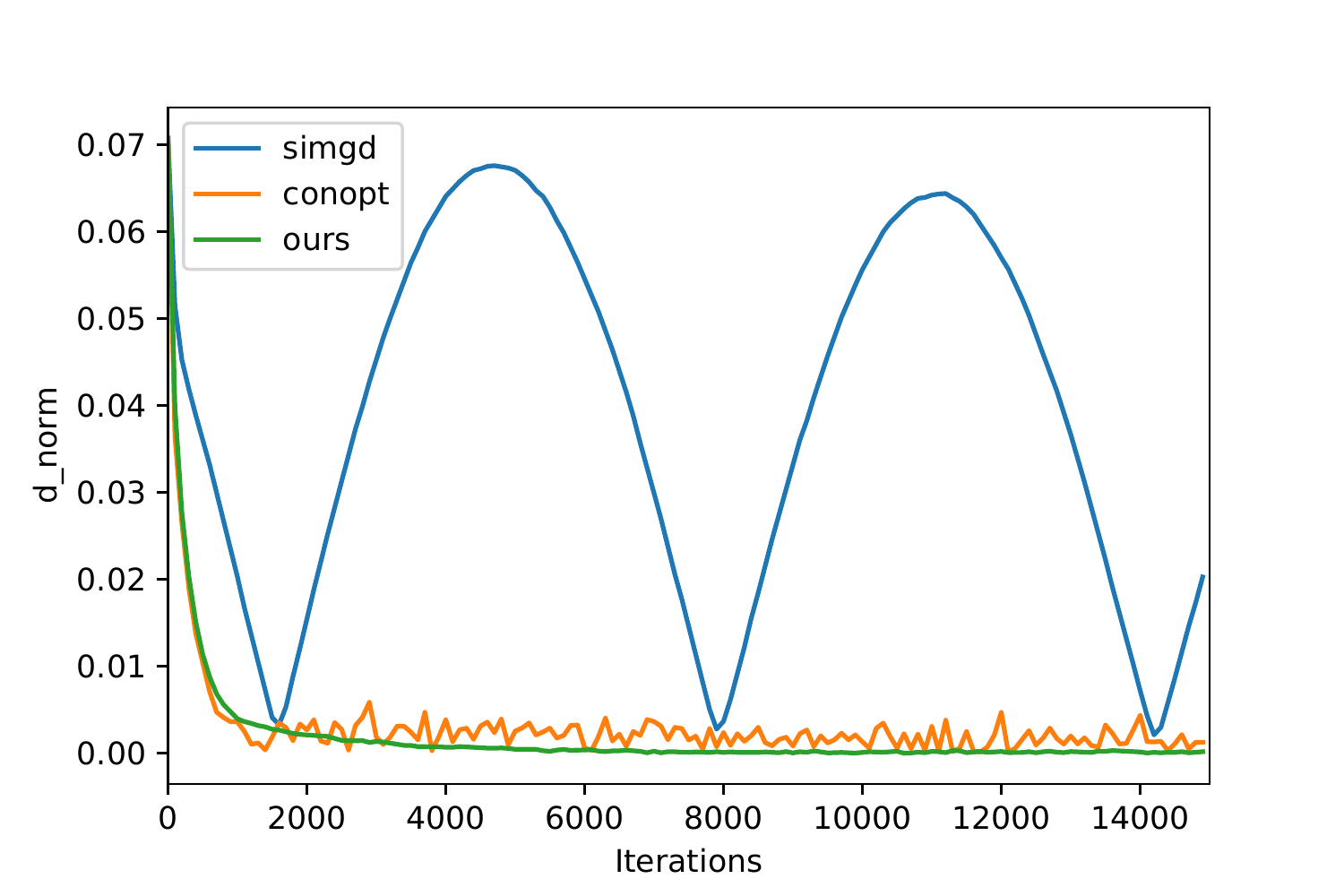}
	    \caption{\small Discriminator training curve}
	\end{subfigure}
	\begin{subfigure}[b]{0.36\textwidth}
		\centering
	    \includegraphics[width=\textwidth]{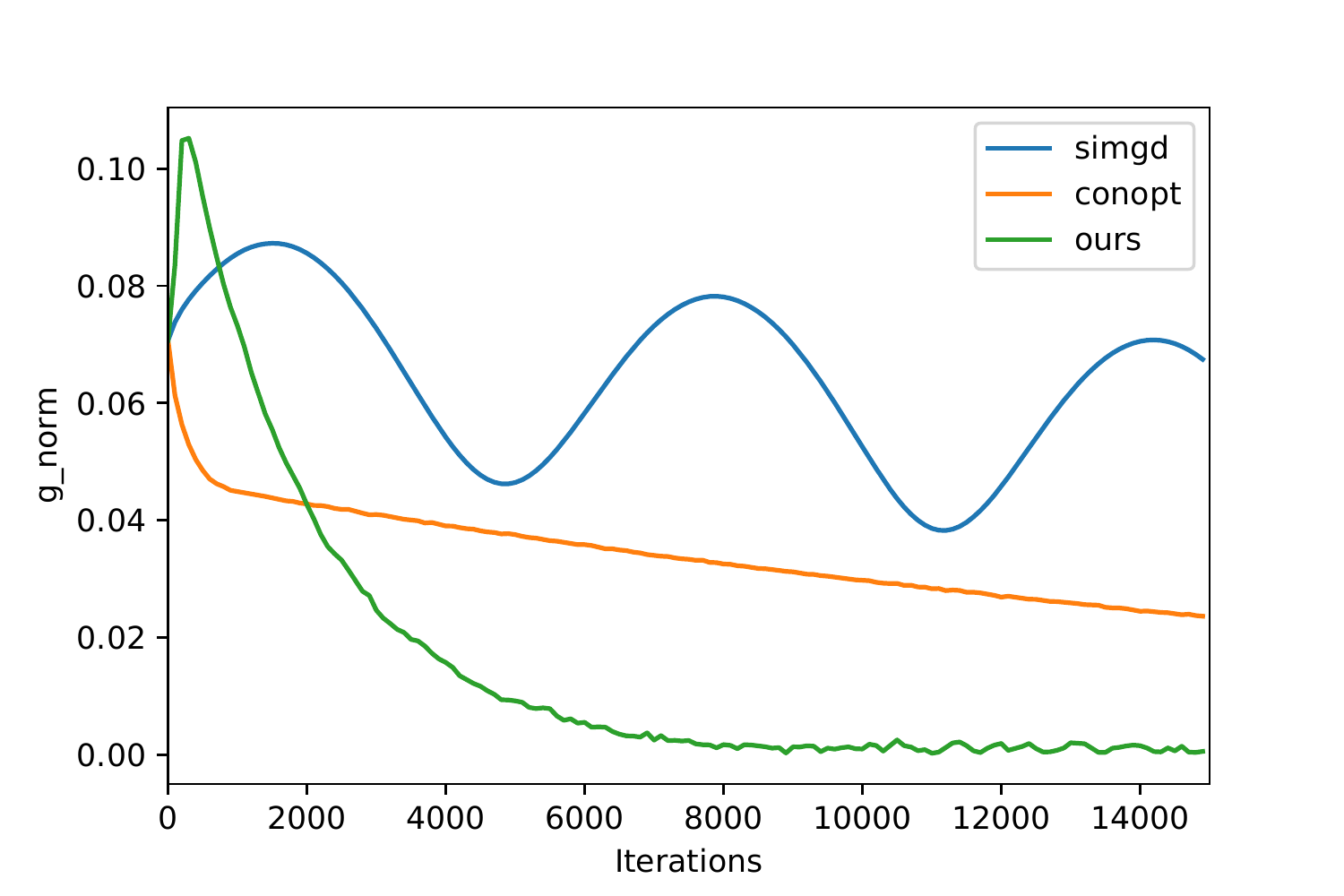}
	    \caption{\small Generator training curve}
	\end{subfigure}

	\caption{\small Training dynamics of SimGD, ConOpt and JARE (Ours) in the simple vanilla GAN example where $\mu=4$ and $\sigma^2 = 0.04$. (a) shows the discriminator convergence where ``d\_norm'' denotes the $l_2$ distance 
	between current and optimal value of the discriminator parameters, and (b) shows the generator convergence where ``g\_norm'' denotes the $l_2$ distance
	between current and optimal value of the generator parameters. } \label{isotLin}
\end{figure}

Figure \ref{isotLin} shows the discriminator and generator training curves, respectively, for three training methods: SimGD, ConOpt and JARE (Ours) by letting $\mu = 4$ and $\sigma^2 = 0.04$. 
We observe that for SimGD, the training curves oscillate with very weak damping which becomes even weaker if we increase $\mu$ or decrease $\sigma^2$ (See Figures \ref{isotLin_mu} and \ref{isotLin_sigma} in Appendix \ref{more_syn_res}). It verifies that SimGD suffers from poor convergence caused by the \textit{Phase Factor} and \textit{Conditioning Factor}. 
Also, ConOpt could alleviate the \textit{Phase Factor} since oscillations caused by complex eigenvalues disappear. However, its generator convergence is heavily slowed down by the \textit{Conditioning Factor} which becomes worse as we increase $\mu$ (See Figure \ref{isotLin_mu} in Appendix \ref{more_syn_res}). In contrast, the JARE enjoys a decent convergence rate for both the generator and discriminator by alleviating the two factors simultaneously.

\begin{figure} [t]
	\centering
	\begin{subfigure}[b]{0.08\textwidth}
		\centering
		\includegraphics[width=\textwidth]{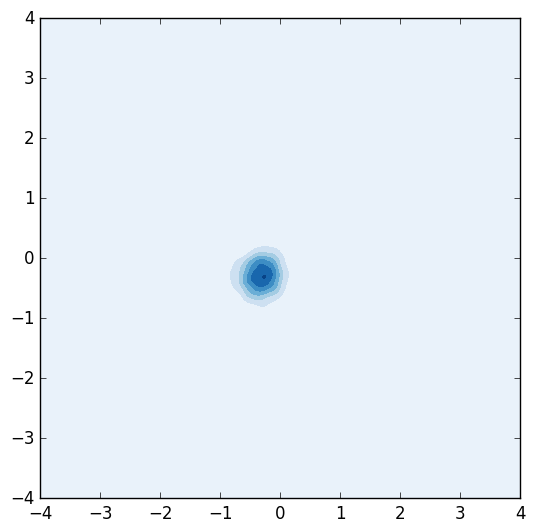}
	\end{subfigure}
	\begin{subfigure}[b]{0.08\textwidth}
		\centering
		\includegraphics[width=\textwidth]{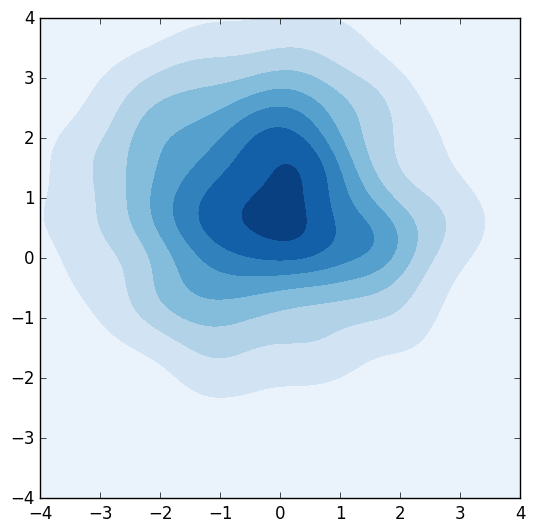}
	\end{subfigure}
	\begin{subfigure}[b]{0.08\textwidth}
		\centering
		\includegraphics[width=\textwidth]{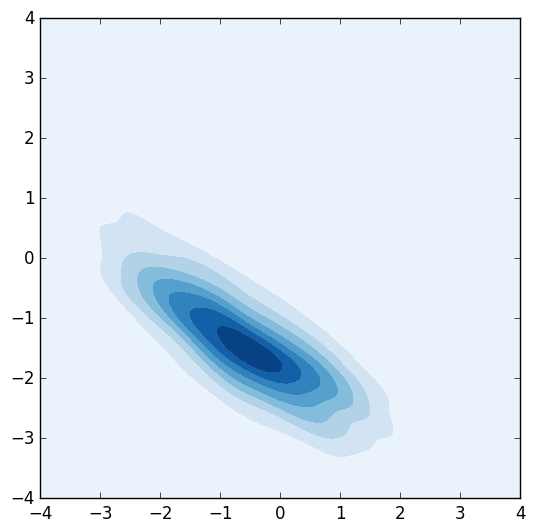}
	\end{subfigure}
	\begin{subfigure}[b]{0.08\textwidth}
		\centering
		\includegraphics[width=\textwidth]{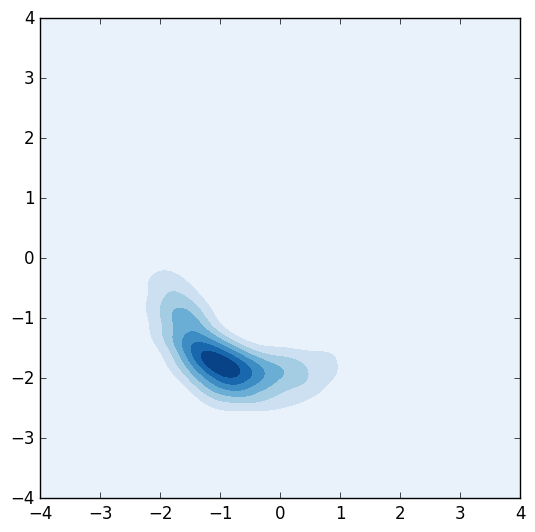}
	\end{subfigure}
	\begin{subfigure}[b]{0.08\textwidth}
		\centering
		\includegraphics[width=\textwidth]{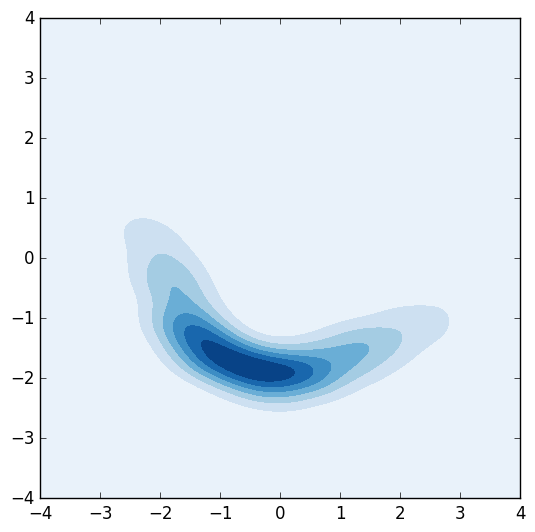}
	\end{subfigure}
	\caption*{\footnotesize (a) SimGD}
	
	\begin{subfigure}[b]{0.08\textwidth}
		\centering
		\includegraphics[width=\textwidth]{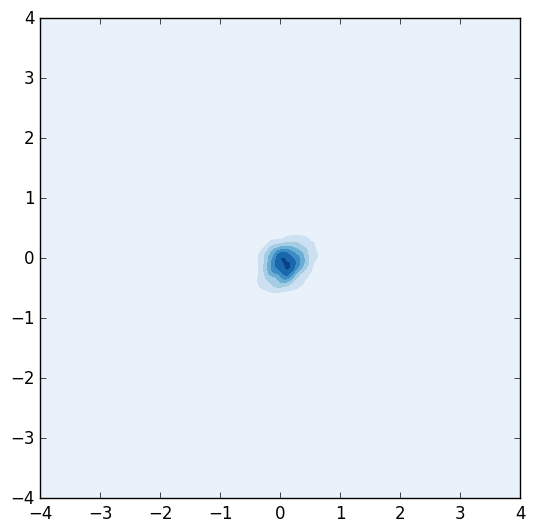}
	\end{subfigure}
	\begin{subfigure}[b]{0.08\textwidth}
		\centering
		\includegraphics[width=\textwidth]{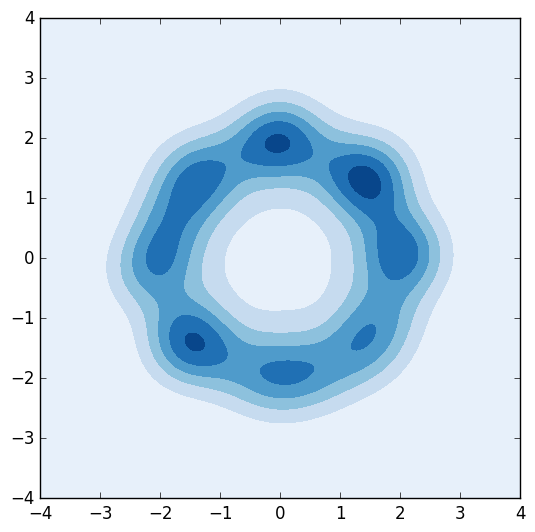}
	\end{subfigure}
	\begin{subfigure}[b]{0.08\textwidth}
		\centering
		\includegraphics[width=\textwidth]{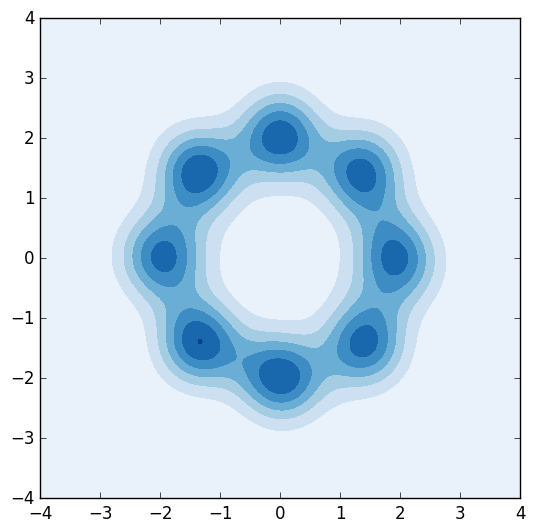}
	\end{subfigure}
	\begin{subfigure}[b]{0.08\textwidth}
		\centering
		\includegraphics[width=\textwidth]{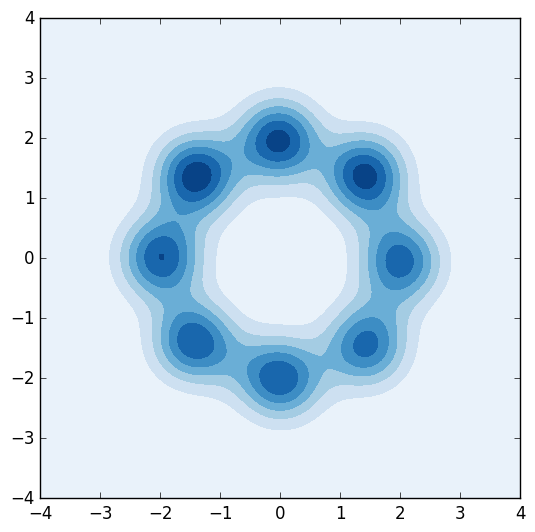}
	\end{subfigure}
	\begin{subfigure}[b]{0.08\textwidth}
		\centering
		\includegraphics[width=\textwidth]{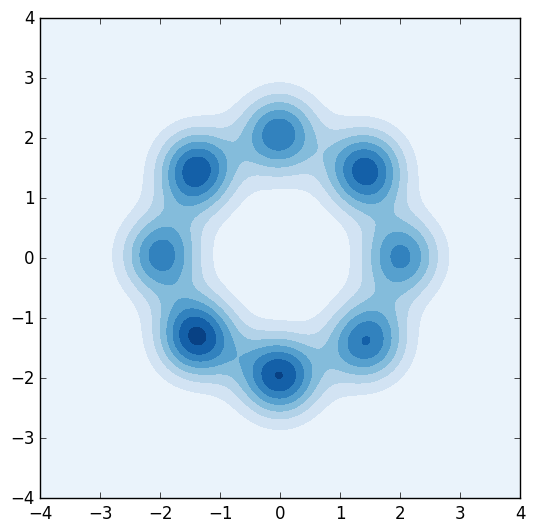}
	\end{subfigure}
	\caption*{\footnotesize (b) ConOpt ($\gamma = 10$)}
	
	\begin{subfigure}[b]{0.08\textwidth}
		\centering
		\includegraphics[width=\textwidth]{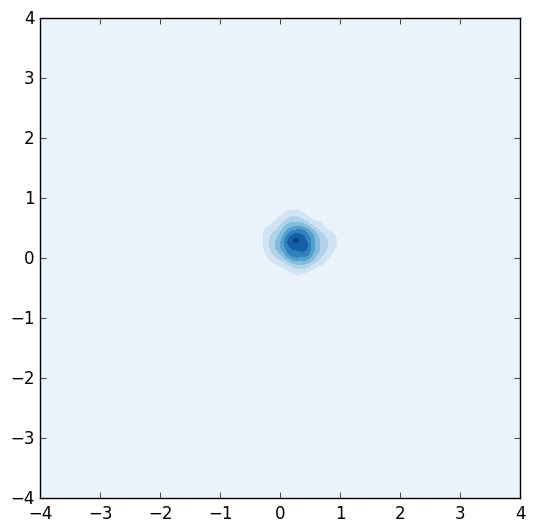}
	\end{subfigure}
	\begin{subfigure}[b]{0.08\textwidth}
		\centering
		\includegraphics[width=\textwidth]{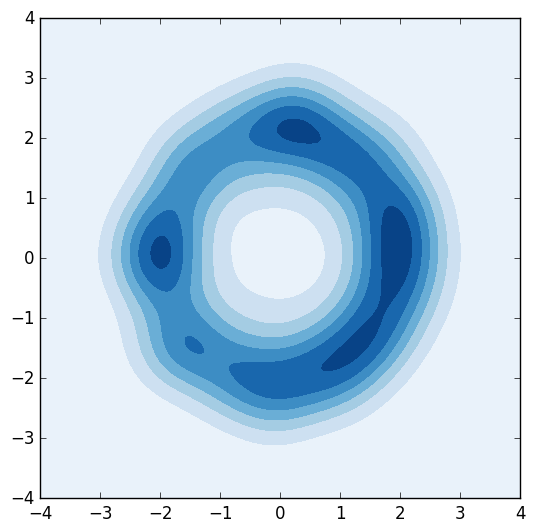}
	\end{subfigure}
	\begin{subfigure}[b]{0.08\textwidth}
		\centering
		\includegraphics[width=\textwidth]{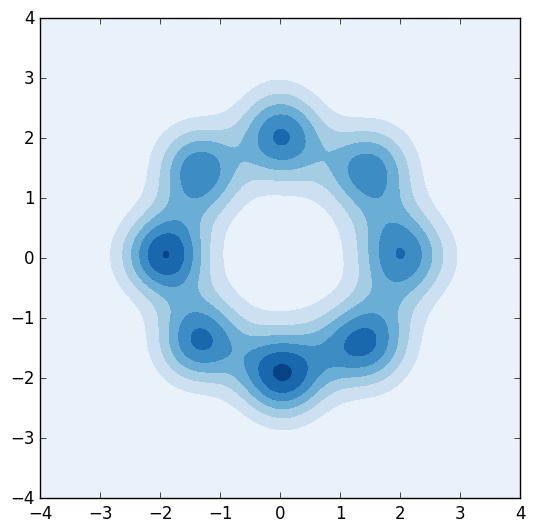}
	\end{subfigure}
	\begin{subfigure}[b]{0.08\textwidth}
		\centering
		\includegraphics[width=\textwidth]{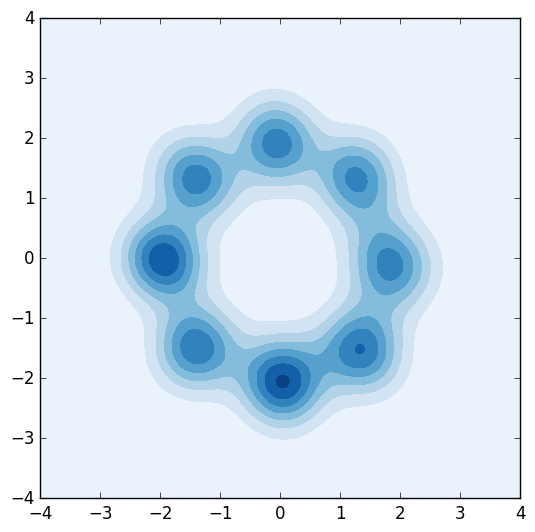}
	\end{subfigure}
	\begin{subfigure}[b]{0.08\textwidth}
		\centering
		\includegraphics[width=\textwidth]{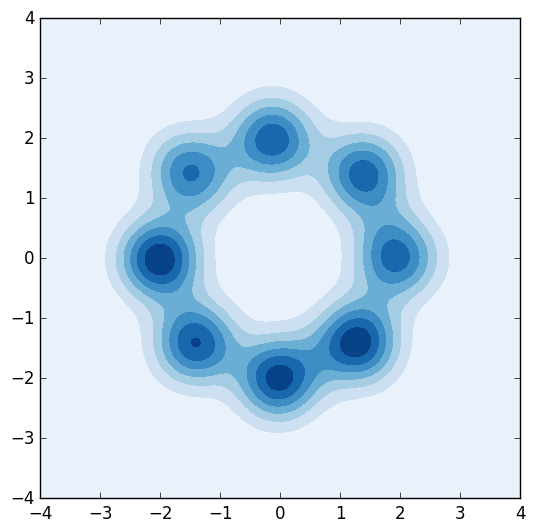}
	\end{subfigure}
	\caption*{\footnotesize (d) Ours ($\gamma = 10$)}

	\caption{\small Comparison of SimGD, ConOpt and JARE (Ours) on the mixture of Gaussians over iterations where $r=2$ and $\gamma = 10$ for both ConOpt and JARE. From left to right, each row consists of the results after 0, 2000, 4000, 6000, 8000 iterations.} \label{gen_gmm_mu20}
\end{figure}

\textbf{Mixture of Gaussians.}
We also test JARE in a commonly used toy example where the goal is to learn a mixture of Gaussians with modes uniformly distributed around a circle with radius $r$. 
Here we set $r=2$ while keeping other settings and network architectures the same with \cite{mescheder2017numerics}. 
We run SimGD, ConOpt and JARE (Ours) with RMSProp \citep{hinton2012neural} and learning rate of $10^{-4}$ for 10K iterations, and the input noise is sampled from a 64-dimensional Gaussian $\mathcal{N}(0, r I_{64})$.
Figure \ref{gen_gmm_mu20} shows their results over different iterations. We can see that SimGD oscillates among different modes and fails to converge, while ConOpt and JARE can both converge to the target data distribution.
Please see Figures \ref{gen_gmm_mu4_full} and \ref{gen_gmm_mu20_full} in Appendix \ref{more_mix_gau} for more detailed comparisons among these methods, where we show JARE with $\gamma = 1000$ tends to behave slightly better than ConOpt in more difficult settings.

\begin{figure} [!t]
    \centering
    \begin{subfigure}[b]{0.37\textwidth}
		\centering
		\includegraphics[width=\textwidth]{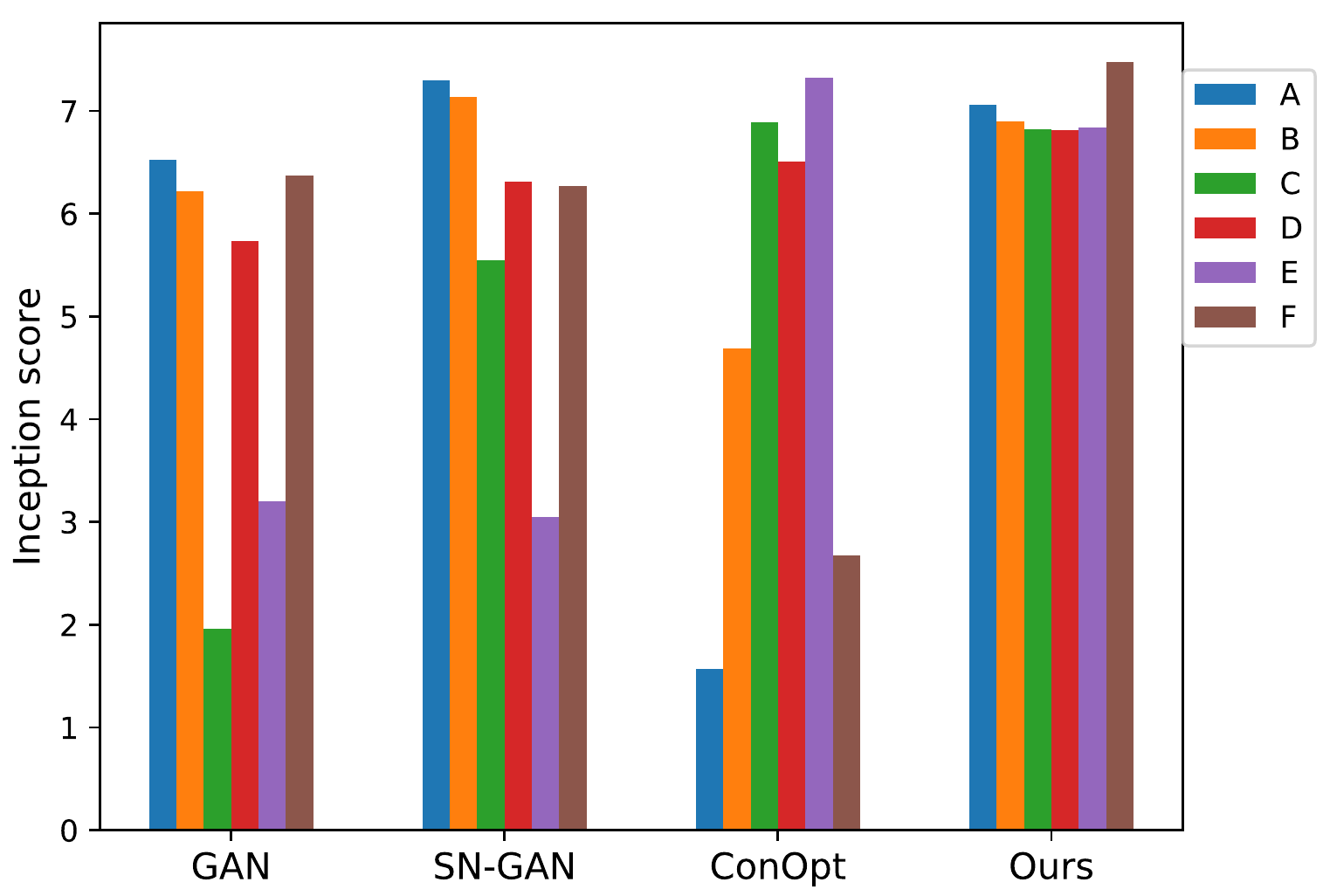}
		\caption{\small Inception score}
	\end{subfigure}
	
	\begin{subfigure}[b]{0.37\textwidth}
		\centering
		\includegraphics[width=\textwidth]{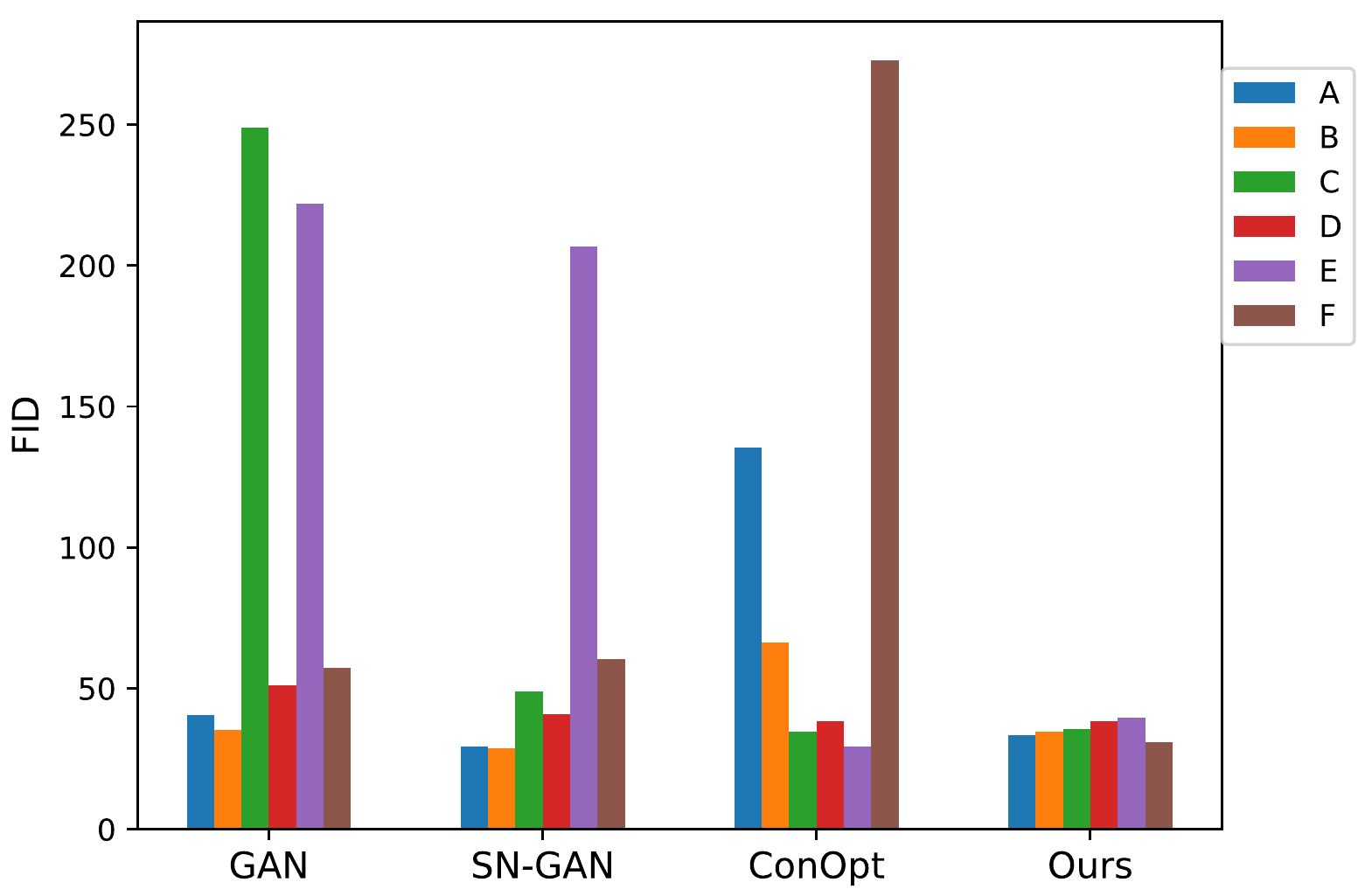}
		\caption{\small FID}
	\end{subfigure}
    \caption{\small Inception scores and FIDs of different methods: GAN, SN-GAN, ConOpt and JARE (Ours) in different GAN settings (A-F) on CIFAR-10. For inception score, the higher is better and for FID, the lower is better. }
    \label{cifar_robust}
\end{figure}


\textbf{CIFAR-10.} 
In this experiment, we quantitatively evaluate the sample quality of JARE on the CIFAR-10 dataset \citep{torralba200880} with the \textit{inception score} \citep{salimans2016improved} and \textit{Frechet inception distance (FID)} \citep{heusel2017gans}. 
We test the dependencies of JARE on different network architectures and hyperparameters. We compare with other GAN training methods, including the standard GAN 
\citep{goodfellow2014generative} (denoted as `GAN'),  ConOpt \citep{mescheder2017numerics}
and SN-GAN \citep{miyato2018spectral}.
For all methods, we use the non-saturating loss as suggested in \cite{goodfellow2014generative}.
For fair comparison, we test 6 settings: the standard CNN model in \cite{miyato2018spectral} with batch normalization \citep{ioffe2015batch} on generator (A) or without batch normalization on generator (B), the DCGAN-like architecture with a constant number of filters in \cite{mescheder2017numerics} via the Adam optimizer \citep{kingma2014adam} (C) or via the RMSProp optimizer (D), and the ResNet \citep{he2016deep} architectures v1 (E) or v2 (F) with a constant number of filters where $M_f=64$. 
Please see Appendix \ref{sec_net_arc} more details. 
Unless otherwise stated,
we use the Adam optimizer with $\beta_1 = 0.5$ and $\beta_2 = 0.999$. Also, we use a batch size of 64 and run all experiments with a learning rate of $10^{-4}$ for 500K iterations. For ConOpt, we set $\gamma = 10$, and for JARE, we set $\gamma = 100$.


\begin{figure} [H]
    \centering
    \includegraphics[width=0.37\textwidth]{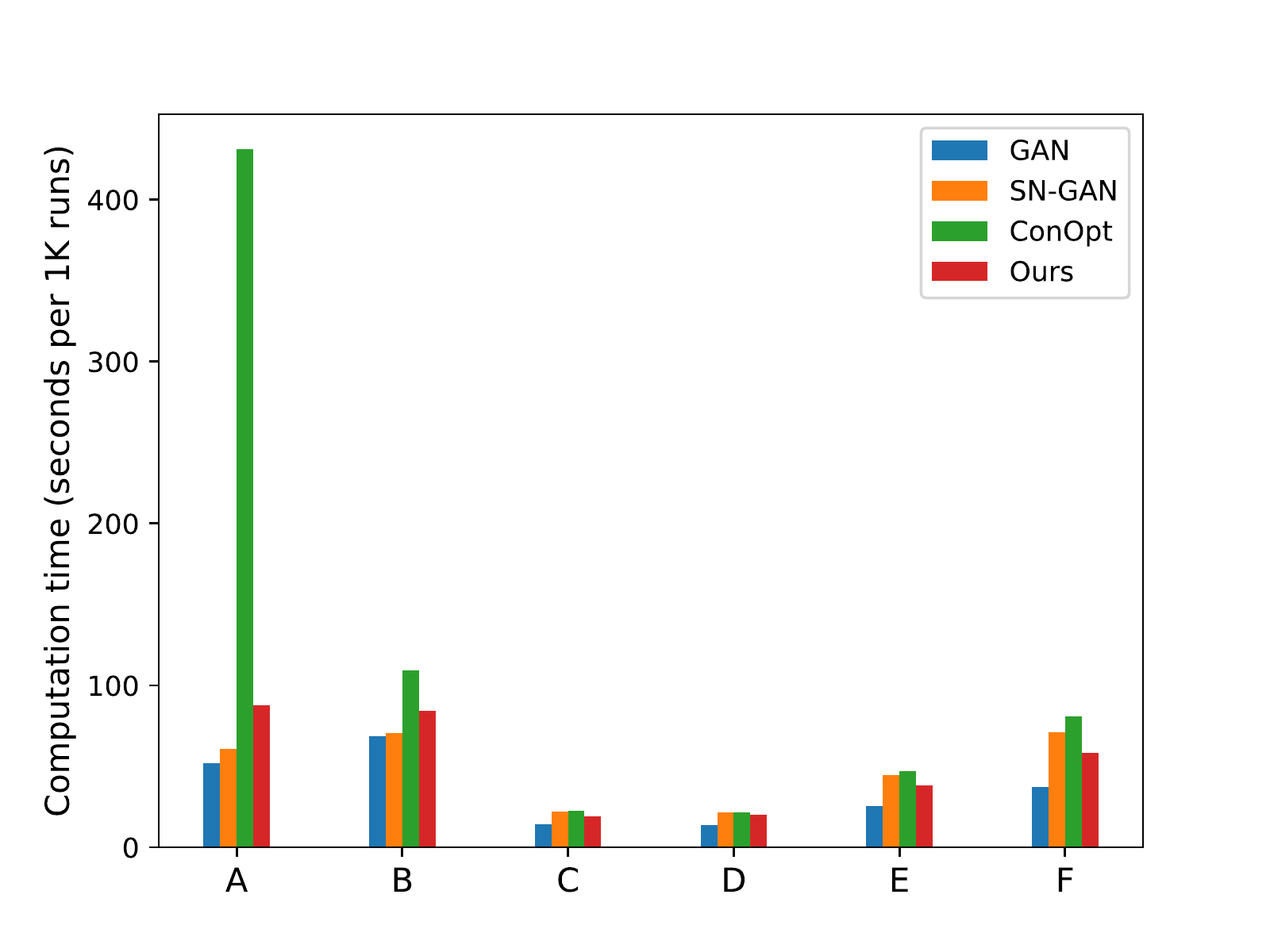}
    \caption{\small Training time on CIFAR-10 with different training methods: GAN, SN-GAN, ConOpt and JARE (Ours) in all the A-F settings. }
    \label{time}
\end{figure}

Figure \ref{cifar_robust} shows the inception scores and FIDs for different training methods with all 6 settings on CIFAR-10 (also see Figure \ref{cifar10_robustness} in Appendix \ref{appendix_cifar10} for the generated samples).
We can see that JARE is more robust than other methods regarding different network architectures and tuning hyperparameters, which shows the potential advantages of JARE in stabilizing the real GAN training.
Both SN-GAN and ConOpt perform almost the best in their own proposed GAN architectures but perform poorly in other cases. 
Besides, the training time (in seconds per 1K runs) on CIFAR-10 with these different methods is given in Table \ref{time}. We can see that the training time of JARE is always lower than ConOpt and on par with SN-GAN, which is not much higher than the standard GAN, which means the extra computational cost introduced by the regularization terms in JARE is relatively low at least in there settings. 


\section{\MakeUppercase{Discussion and Conclusions}}

In this paper, we first analyzed the non-asymptotic local convergence behavior of GAN training dynamics in a simple GAN example and later extended the analysis to the general GAN scenario. We found out that in order to ensure a good convergence behavior in GANs, both the \textit{Phase Factor} and \textit{Conditioning Factor} need to be addressed simultaneously. However, we showed that previous gradient-based regularizations can only avoid one factor while making the other more severe. Therefore, we proposed a new Jacobian regularization for GANs, called JARE, and showed theoretically it can alleviate the two factors simultaneously. Finally, we did experiments on isotropic Gaussian, mixture of Gaussians and CIFAR-10 to show the training stability of JARE. 

However, the proposed JARE also has its limitation:
Although it is constructed in a principled way and the preliminary experimental results showed its potential benefits at stabilizing GANs, in order to scale to the large-scale GAN training \citep{brock2018large}, we will need to get rid of the computationally expensive second-order derivatives in the regularization term of JARE. 
A potential direction of alleviating this limitation could be interpreting JARE as a form of \textit{adversarial extrapolation} where two agents playing the game anticipate each other's learning updates, which we think is game-theoretic, intuitive and thought-provoking. 

For example, 
different from current extrapolation methods applied in GANs \citep{gidel2018variational,daskalakis2017training, yadav2017stabilizing}, we can introduce the \textit{adversarial extrapolation} as follows:
\begin{align} \label{advext}
    \begin{split}
        \phi^{(k+1)} &= \phi^{(k)} - \eta \nabla_{\phi}f(\phi^{(k)}, \theta^{(k+\frac{1}{2})}) \\
        \theta^{(k+1)} &= \theta^{(k)} + \eta \nabla_{\theta}f(\phi^{(k+\frac{1}{2})}, \theta^{(k)})
    \end{split}
\end{align}
where the intermediate prediction terms $\phi^{(k+\frac{1}{2})}$ and $\theta^{(k+\frac{1}{2})}$ are given by
\begin{align}
    \begin{split}
        \phi^{(k+\frac{1}{2})} &= \phi^{(k)} - \frac{\gamma}{2} \nabla_{\phi}f(\phi^{(k)}, \theta^{(k)}) \\
        \theta^{(k+\frac{1}{2})} &= \theta^{(k)} + \frac{\gamma}{2} \nabla_{\theta}f(\phi^{(k)}, \theta^{(k)})
    \end{split}
\end{align}
As we can see, by applying the first-order Taylor approximation with respect to $\theta^{(k+\frac{1}{2})}$ and $\phi^{(k+\frac{1}{2})}$, respectively, we have
\begin{align*} 
    \begin{split}
        f(\phi^{(k)}, \theta^{(k+\frac{1}{2})}) & \approx f(\phi^{(k)}, \theta^{(k)}) + \frac{\gamma}{2} \left\Vert \nabla_{\theta}f(\phi^{(k)}, \theta^{(k)}) \right\Vert^2\\
        f(\phi^{(k + \frac{1}{2})}, \theta^{(k)}) & \approx f(\phi^{(k)}, \theta^{(k)}) - \frac{\gamma}{2} \left\Vert \nabla_{\phi}f(\phi^{(k)}, \theta^{(k)}) \right\Vert^2
    \end{split}
\end{align*}
Thus, by substituting the above two approximations into (\ref{advext}), we can get the proposed JARE in (\ref{proposed_update}).

From the above analysis, the new adversarial extrapolation method derived directly from JARE may enjoy both our theoretical grounding and the computational benefits of the first-order optimization methods. We leave the further investigation of the adversarial extrapolation method, and the relationship between JARE and other extrapolation methods in GANs, as the future work.



\vspace*{-3pt}
\section*{Acknowledgement}
\vspace*{-3pt}

We would like to thank all the reviewers for their helpful comments. WN and ABP were supported by IARPA via DoI/IBC contract D16PC00003 and NSF NeuroNex grant DBI-1707400.

\newpage 

\bibliography{main}



\newpage
\text{ }
\newpage

\appendix

\twocolumn[ {\begin{center}
    \textbf{\Large Appendix \\ \text{ }} 
\end{center}}]

\section{Proofs in Section \ref{sec_simple_gan}}

\subsection{Proof of Lemma \ref{lem_concave_concave}}
\label{proof_lemma1}

By taking the second derivative of $f(\phi, \theta)$ in (\ref{obj_linear}) w.r.t. $\phi$, we have 
\begin{align*}
\begin{split}
\nabla^2_{\phi\phi} f(\phi,\theta) = \theta \theta^T \mathbb{E}_{z \sim \mathcal{N}(0, \sigma^2 I)}[g''_2({\theta}^T(\phi + z))]
\end{split}
\end{align*}
By the concavity of $g_2$, we know the scalar term $\mathbb{E}_{z \sim \mathcal{N}(0, \sigma^2 I)}[g''_2({\theta}^T(\phi + z))] \leq 0$. Thus, we have 
\begin{align*}
\begin{split}
\nabla^2_{\phi\phi} f(\phi,\theta) \preceq 0
\end{split}
\end{align*}
Similarly, by taking the second derivative of $f(\phi, \theta)$ in (\ref{obj_linear}) w.r.t. $\theta$, we have 
\begin{align*}
\begin{split}
\nabla^2_{\phi\phi} &f(\phi,\theta) = \mathbb{E}_{z \sim \mathcal{N}(v, \sigma^2 I)}[g''_1({\theta}^T x) xx^T] \\
 &+ \mathbb{E}_{z \sim \mathcal{N}(0, \sigma^2 I)}[g''_2({\theta}^T(\phi + z)) (\phi + z) (\phi + z)^T]
\end{split}
\end{align*}
By the concavity of $g_1$ and $g_2$, we know the scalar terms $g''_1({\theta}^T x) \leq 0$ and $g''_2({\theta}^T(\phi + z)) \leq 0$. Since  $xx^T \succeq 0$ and $(\phi + z) (\phi + z)^T \succeq 0$, we have
\begin{align*}
\begin{split}
\nabla^2_{\theta\theta} f(\phi,\theta) \preceq 0
\end{split}
\end{align*}
as required. \hfill $\square$

\subsection{Proof of Lemma \ref{lemma2}}
\label{proof_lemma2}

\emph{Proof.} 
First, we have 
\begin{align*}
\begin{split}
\nabla_{\phi} f(\phi,\theta) = \mathbb{E}_{z \sim \mathcal{N}(0, \sigma^2 I)}[g'_2({\theta}^T(\phi + z))\theta]
\end{split}
\end{align*}
Since the equilibrium point $(\phi^*, \theta^*)$ satisfies ${\theta^*}^T(\phi^* + z) = 0$, for points $(\phi,\theta)$ near the equilibrium, we know $g'_2(\theta^T(\phi+z)) = g'_2(0) + g''_2(0)\theta^T(\phi+z) + o(\|\theta\|)$ by Taylor expansion. That is, by ignoring the small term with norm $o(\|\theta\|)$, we have
\begin{align*}
\begin{split}
\nabla_{\phi} f(\phi,\theta) &\approx \mathbb{E}_{z \sim \mathcal{N}(0, \sigma^2 I)}[g'_2(0)\theta + g''_2(0)\theta{\theta}^T(\phi + z)] \\
&= g'_2(0)\theta +  g''_2(0)\theta\theta^T\phi \\
&\mathop \approx \limits^{(a)} g'_2(0)\theta 
\end{split}
\end{align*}
where $(a)$ is also from ignoring the small term with norm $o(\|\theta\|)$. Similarly, 
\begin{align*}
\begin{split}
 \nabla_{\theta} & f(\phi,\theta) = \mathbb{E}_{x \sim \mathcal{N}(v, \sigma^2 I)}\left[g'_1({\theta}^Tx)x \right]  \\
& \qquad\qquad + \mathbb{E}_{\tilde{x} \sim \mathcal{N}(\phi, \sigma^2 I)}\left[g'_2({\theta}^T\tilde{x})\tilde{x}\right] \\
\mathop \approx \limits^{(a)} & \mathbb{E}_{x \sim \mathcal{N}(v, \sigma^2 I)} \left[\left(g'_1(0) + g''_1(0)\theta^Tx\right)x \right]  \\
& +\mathbb{E}_{\tilde{x} \sim \mathcal{N}(\phi, \sigma^2 I)} \left[\left(g'_2(0) + g''_2(0){\theta}^T\tilde{x}\right)\tilde{x} \right] \\
=& g'_1(0)v + g''_1(0)\left(\sigma^2I + vv^T\right)\theta  + g'_2(0)\phi \\
& + g''_2(0)\left(\sigma^2I+\phi\phi^T\right)\theta \\
\mathop \approx \limits^{(b)} & g'_1(0)v + g'_2(0)\phi + \left( g''_1(0)+g''_2(0)\right)\left(\sigma^2 I + vv^T\right) \theta
\end{split}
\end{align*}
where $(a)$ is from $g'_1(\theta^Tx) = g'_1(0) + g''_1(0)\theta^Tx + o(\|\theta\|)$ and $g'_2(\theta^T \tilde{x}) = g'_2(0) + g''_2(0)\theta^T \tilde{x} + o(\|\theta\|)$ by Taylor expansion, and $(b)$ is from $\|\phi - v\| = o(1)$.

For second-order derivatives, we have
\begin{align*}
\begin{split}
\nabla^2_{\phi\phi} f(\phi,\theta) =& \mathbb{E}_{z \sim \mathcal{N}(0, \sigma^2 I)} \left[g''_2({\theta}^T(\phi + z))\theta\theta^T \right] \\
\mathop \approx \limits^{(a)} & g''_2(0)\theta\theta^T
\end{split}
\end{align*}
where $(a)$ also follows from $g''_2(\theta^T(\phi+z)) = g''_2(0) + o(1)$ by Taylor expansion. Also,
\begin{align*}
\begin{split}
& \nabla^2_{\theta\phi} f(\phi,\theta) = \mathbb{E}_{\tilde{x} \sim \mathcal{N}(\phi, \sigma^2 I)}[g'_2({\theta}^T\tilde{x})I + g''_2({\theta}^T \tilde{x})\tilde{x}\theta^T] \\
&\mathop \approx \limits^{(a)} \mathbb{E}_{\tilde{x} \sim \mathcal{N}(\phi, \sigma^2 I)}[\left(g'_2(0) + g''_2(0){\theta}^T\tilde{x}\right)I + g''_2(0)\tilde{x}\theta^T]\\
& = g'_2(0)I + g''_2(0)\theta^T\phi I + g''_2(0)\phi\theta^T \\
& \mathop \approx \limits^{(b)} g'_2(0) I
\end{split}
\end{align*}
where $(a)$ is from $g'_2(\theta^T \tilde{x}) = g'_2(0) + g''_2(0)\theta^T \tilde{x} + o(\|\theta\|)$ and $g''_2(\theta^T \tilde{x}) = g''_2(0) + o(1)$ by Taylor expansion, and $(b)$ is from  $\|\theta\| =  o(1)$,
and 
\begin{align*}
\begin{split}
& \nabla^2_{\theta\theta} f(\phi,\theta) = \mathbb{E}_{x \sim \mathcal{N}(v, \sigma^2 I)}\left[g''_1({\theta}^Tx)x x^T \right]  \\ 
& \qquad \qquad \qquad  + \mathbb{E}_{\tilde{x} \sim \mathcal{N}(\phi, \sigma^2 I)}\left[g''_2({\theta}^T\tilde{x})\tilde{x}\tilde{x}^T\right] \\
&\mathop \approx \limits^{(a)} \mathbb{E}_{x \sim \mathcal{N}(v, \sigma^2 I)}\left[g''_1(0)x x^T \right]   + \mathbb{E}_{\tilde{x} \sim \mathcal{N}(\phi, \sigma^2 I)}\left[g''_2(0)\tilde{x}\tilde{x}^T\right] \\
&\mathop \approx \limits^{(b)} \left(g''_1(0) + g''_2(0)\right)\left(\sigma^2 I + vv^T\right)
\end{split}
\end{align*}
where $(a)$ is from $g''_1(\theta^T x) = g''_1(0) + o(1)$ and $g''_2(\theta^T \tilde{x}) = g''_2(0) + o(1)$ by Taylor expansion, and $(b)$ is from $\|\phi - v\| = o(1)$.  
\hfill $\square$

\subsection{Proof of Theorem \ref{thm_simgd}}
\label{proof_thm_simgd}

\textit{Proof.}
For the vanilla GAN, we know $g_1(t) = g_2(-t) = -\log(1+e^{-t})$. Then we have $g'_1(0) = \frac{1}{2}$, $g'_2(0) = -\frac{1}{2}$ and $g''_1(0) = g''_2(0) = -\frac{1}{4}$. 
From the proof of Lemma \ref{lemma2}, the updates (\ref{simGD_re}) of SimGD for points near the equilibrium $w^*$ become
\begin{align} \label{A_simgd}
\begin{split}
    w^{(k+1)} &= w^{(k)} + \eta \begin{bmatrix}
\frac{1}{2} \theta^{(k)} \\ \frac{1}{2}(\phi^{(k)}-v) + \frac{1}{2} \left(\sigma^2 I + vv^T\right) \theta^{(k)}
\end{bmatrix} \\
&= w^{(k)} + \eta 
\underbrace{\begin{bmatrix}
	0 & \frac{1}{2} I \\
	-\frac{1}{2} I  & -\frac{1}{2} \left(\sigma^2 I + vv^T\right) 
\end{bmatrix} }_{\triangleq A}
w^{(k)}
\end{split}
\end{align}
where $w^{(k)} \triangleq \begin{bmatrix}
	\phi^{(k)} - v \\ \theta^{(k)}
\end{bmatrix}$. Next, we need to compute the eigenvalues of the Jacobian $A$. By definition, let $A y=\lambda y$ where the eigenvector satisfies $y = \begin{bmatrix}
 y_1 \\ y_2
\end{bmatrix} \neq 0$, then we have 
\begin{align}
	\frac{1}{2} y_2 = \lambda y_1 \label{simGD_eig_1}\\
	-\frac{1}{2} y_1 - \frac{1}{2} \left(\sigma^2 I + vv^T\right) y_2 = \lambda y_2 \label{simGD_eig_2}
\end{align}
First, we know $\lambda \neq 0$, otherwise, we get $y=0$ which violates the definition of eigenvectors. Thus from (\ref{simGD_eig_1}) we have $y_1 = \frac{1}{2\lambda}y_2$. Plugging it into (\ref{simGD_eig_2}) yields
\begin{align} \label{eq_eig1}
	- \lambda vv^Ty_2 = (2 \lambda^2 + \sigma^2 \lambda + \frac{1}{2}) y_2
\end{align}
Then we can evaluate $\lambda$ in two cases: 

1) $v^Ty_2 = 0$. From (\ref{eq_eig1}) we have $(4 \lambda^2 + 2 \sigma^2 \lambda + 1) y_2 = 0$. Similarly we know $y_2 \neq 0$, otherwise, we get $y_1=0$ as well from (\ref{simGD_eig_2}) which again violates the definition of eigenvectors. Thus, the coefficient satisfies $4 \lambda^2 + 2 \sigma^2 \lambda + 1 = 0$, and solving this equation yields $\lambda_{1,2}(A)$ in the theorem.

2) $v^Ty_2 \neq 0$. By left multiplying $v^T$ on  both sides of Eq. (\ref{eq_eig1}) we get 
$- \lambda \|v\|^2 v^Ty_2 = (2 \lambda^2 + \sigma^2 \lambda + \frac{1}{2}) v^Ty_2$. 
Since $v^Ty_2 \neq 0$,
then
$4 \lambda^2 + 2 (\sigma^2+\|v\|^2) \lambda + 1 = 0$, and solving this equation yields $\lambda_{3,4}(A)$ in the theorem.  \hfill $\square$


\subsection{Proof of Corollary \ref{coro1}} 
\label{proof_coro1}

\textit{Proof.} In the first part of the proof, we try to find the range of the step size $\eta$. Given $\sigma^2 < 2$, we know $\lambda_{1,2}(A)$ are complex eigenvalues and thus $|1+\eta \lambda_{1,2}(A)| = \frac{1}{4}\eta^2-\frac{\sigma^2}{2}\eta + 1$. Since it requires $|1+\eta \lambda_{1,2}(A)| < 1$ to ensure the non-asymptotic convergence, by setting $\frac{1}{4}\eta^2-\frac{\sigma^2}{2}\eta + 1 < 1$ we get $0 <  \eta < 2\sigma^2$. As we know $\zeta = \sqrt{(\frac{2}{\sigma^2})^2-1}$ in the simple vanilla GAN example, then $\sigma^2 = \frac{2}{\sqrt{1+\zeta^2}}$, which means $0 <  \eta < \frac{4}{\sqrt{1+\zeta^2}}$.

In the second part of the proof, we try to find the lower bound of the number of iterations $N$ given the step size constraint. We know $\frac{1}{4}\eta^2-\frac{\sigma^2}{2}\eta + 1 \geq \sqrt{1 - \left(\frac{\sigma^2}{2}\right)^2} $ with the equality holds at $\eta = \sigma^2$. 
Therefore, for the step size $\eta$ satisfying $0 <  \eta < \frac{4}{\sqrt{1+\zeta^2}}$, we have $ \frac{1}{\sqrt{1+\frac{1}{\zeta^2}}} \leq |1+\eta \lambda_{1,2}(A)| < 1$.
Thus, for the updates $w^{(k)} = (I + \eta A) w^{(k-1)}$, it is easy to get $\tilde{w}^{(k)} = (I + \eta \Lambda) \tilde{w}^{(k-1)}$ where the eigen-matrix $\Lambda$ satisfying $\Lambda = P A P^{-1}$ with $P$ invertible and $\tilde{w}^{(k)} = P w^{(k)}$. Apparently, $|\tilde{w}_j^{(k)}| = | I + \eta \lambda_{1,2}(A) |^k | \tilde{w}_j^{(0)} |$ where the index $j$ refers to the entry in $\tilde{w}^{(k)}$ related to the eigenvalues $\lambda_{1,2}(A)$. Also, we know $  \| \tilde{w}^{(k)} \|  \geq |\tilde{w}_j^{(k)}|$ and $ \| \tilde{w}^{(k)} \| = \| P w^{(k)} \| \leq \| P \| \| w^{(k)} \| $, so we have $\|{w}^{(k)}\| \geq | I + \eta \lambda_{1,2}(A) |^k \| P \|^{-1} | \tilde{w}_j^{(0)} |$. Therefore, for the $\epsilon$-error solution $\|w^{(N)}\| \leq \epsilon$ after $N$ iterations,  we have $( 1 + \frac{1}{\zeta^2} )^{-\frac{N}{2}} \| P \|^{-1} | \tilde{w}_j^{(0)} | \leq \epsilon$. By letting $C_0 = \| P \|^{-1} | \tilde{w}_j^{(0)} |$, we can easily get the lower bound of $N$. 
\hfill $\square$

\vspace{3mm}

\subsection{Proof of Corollary \ref{coro2}}
\label{proof_coro2}

\textit{Proof.} In the first part of the proof, we try to find the range of the step size $\eta$. Given $\beta^2 > 2$, $\lambda_{3,4}(A)$ are both real eigenvalues. Similarly, to ensure the non-asymptotic convergence, the step size $\eta$ also satisfies $|1+\eta \lambda_{3,4}(A)| < 1$. From Theorem \ref{thm_simgd} we have $1+\eta \lambda_{3}(A) = 1 -  \frac{\beta^2 + {\sqrt{\left(\beta^2\right)^2-4}}}{4} \eta$ and $1+\eta \lambda_{4}(A) = 1 -  \frac{\beta^2 - {\sqrt{\left(\beta^2\right)^2-4}}}{4} \eta$. 
Next, we analyze $\lambda_3(A)$ and $\lambda_4(A)$ separately. 
To ensure $|1+\eta \lambda_{3}(A)| < 1$, then $0 < \eta < \frac{8}{\beta^2 + {\sqrt{\left(\beta^2\right)^2-4}}}$. 
As we know $\tau = \frac{1}{4}{( \beta^2 + {\sqrt{(\beta^2)^2-4}} )^2}$ in the simple vanilla GAN example, then $\frac{8}{\beta^2 + {\sqrt{\left(\beta^2\right)^2-4}}} = \frac{4}{\sqrt{\tau}}$, which means $0 < \eta <  \frac{4}{\sqrt{\tau}}$. Also, to satisfy $|1+\eta \lambda_{4}(A)| < 1$, then $0 < \eta < 2(\beta^2 + {\sqrt{\left(\beta^2\right)^2-4}}) = 4 \sqrt{\tau}$. As we know $\tau > 1$ by definition, the step size $\eta$ satisfies $0 < \eta < \min\{ \frac{4}{\sqrt{\tau}}, 4\sqrt{\tau} \} = \frac{4}{\sqrt{\tau}}$.

In the second part of the proof, we try to find the lower bound of the number of iterations $N$ given the step size constraint. We know $|1+\eta \lambda_{4}(A)| = | 1 -  \frac{1}{2\sqrt{\tau}} \eta |$ and for $0 < \eta < \frac{4}{\sqrt{\tau}}$ we get $1-\frac{2}{\tau} < 1 - \frac{1}{2\sqrt{\tau}}\eta < 1$, Therefore, if $1 < \tau < 2$, then $-1 < 1-\frac{2}{\tau} < 0$, and thus $0 < |1+\eta \lambda_{4}(A)| < 1$. If $\tau \geq 2$, then $1 - \frac{2}{\tau} > 0$, and thus $1-\frac{2}{\tau} < |1+\eta \lambda_{4}(A)| < 1$. Putting them together, we get $\max \{1-\frac{2}{\tau}, 0 \} < |1+\eta \lambda_{4}(A)| < 1$. 
Similar to the proof of Corollary \ref{coro1}, we rewrite the updates as $\tilde{w}^{(k)} = (I + \eta \Lambda) \tilde{w}^{(k-1)}$ where the eigen-matrix $\Lambda$ satisfying $\Lambda = P A P^{-1}$ with $P$ invertible and $\tilde{w}^{(k)} = P w^{(k)}$. Here we focus on $|\tilde{w}_{j'}^{(k)}| = | I + \eta \lambda_{1,2}(A) |^k | \tilde{w}_{j'}^{(0)} |$ where the index $j'$ refers to the entry in $\tilde{w}^{(k)}$ related to the eigenvalues $\lambda_{4}(A)$. Also, we know $\|{w}^{(k)}\| \geq | I + \eta \lambda_{1,2}(A) |^k C_1$ where $ C_1 = \| P \|^{-1} | \tilde{w}_{j'}^{(0)} |$. Therefore,
for $\tau > 2$, we get $\|w^{(k)}\| \geq ( 1 - \frac{2}{\tau} )^{\frac{k}{2}} C_1$. For the $\epsilon$-error solution $\|w^{(N)}\| \leq \epsilon$,  we have $( 1 - \frac{2}{\tau} )^{\frac{N}{2}} C_1 \leq \epsilon$ which yields the  lower bound of $N$. 
\hfill $\square$

\section{An Example of Full Rank Representations}
\label{app_example}

In the simple vanilla GAN example, if we consider the zero noise-limit case, i.e. $\sigma^2 = 0$, and assume $n=1$, from Theorem \ref{thm_simgd} we know the eigenvalues of the Jacobian $A$ are 
\begin{align} \label{rank_1}
    \lambda_{1,2}(A) = \frac{- v^2 \pm {\sqrt{\left(v^2\right)^2-4}}}{4}
\end{align}
When $v \to 0$, $\lambda_{1,2}(A) \to \pm \frac{1}{2}i$ with an infinitely large imaginary-to-real ratio $\zeta$, which obviously suffers from the impact of the \textit{Phase Factor}. 

To alleviate this issue, one solution could be to increase the expressive power of discriminator. For instance, it is suggested by \cite{mescheder2018training} that we can replace the linear discriminator $D_{\theta}(x) = \theta x$ by the discriminator with the so-called full-rank representations $D_{\theta}(x) = \theta e^{x}$. Similarly, in the zero noise-limit case with $n=1$, we first rewrite the objective (\ref{obj_linear}) as $f(\theta, \phi) = g_1(\theta e^x) + g_2(-\theta e^x)$. For the vanilla GAN, we have $g_1(t) = g_2(-t) = -\log(1+e^{-t})$. Then the Jacobian $A$ of all points within $B_{\delta}(w^*)$ is evaluated as $A = \begin{bmatrix} 
0 & \frac{1}{2}e^v \\
-\frac{1}{2}e^v & -\frac{1}{2}e^{2v}
\end{bmatrix}
$
and its eigenvalues are 
\begin{align} \label{rank_2}
    \lambda_{1,2}(A) = \frac{- e^{2v} \pm {\sqrt{e^{4v}-4e^{2v}}}}{4}
\end{align}
Now when $v \to 0$, $\lambda_{1,2}(A) \to \frac{- 1 \pm \sqrt{3}i}{4}$ with the imaginary-to-real ratio $\zeta = \sqrt{3}$. By Corollary \ref{coro1}, the impact of the \textit{Phase Factor} has been effectively alleviated when $v$ is very small. 

However, the impact of the \textit{Conditioning Factor}, if it exists, becomes much more severe. Asymptotically when $v$ is sufficiently large, from (\ref{rank_1}) we know that $\tau$ increases in the order of $v^4$, but (\ref{rank_2}) shows that $\tau$ increases in the order of $e^{2v}$. For example, if we assume $v = 5$, the eigenvalues of the original Jacobian (\ref{rank_1}) is evaluated as $\lambda_{1,2}(A) =  \frac{-25 \pm \sqrt{621}}{2}$ with $\tau = \Omega(10^2)$. However, after using the discriminator with full-rank representations, the eigenvalues of the new Jacobian (\ref{rank_2}) is evaluated as $\lambda_{1,2}(A) = \frac{- e^{10} \pm {\sqrt{e^{20}-4e^{10}}}}{4}$ with $\tau = \Omega(10^5)$.

\section{A Condition of Choosing the Regularization Matrix }
\label{note_conopt}

First, we note that the regularization matrix $\Gamma$ introduced by a good Jacobian regularization method cannot be arbitrary and a particular condition is given as follows. 
\begin{claim} [\textbf{Non-Reversing-Flow Condition}] \label{claim_gflow}
	By applying the regularization matrix $\Gamma$, it should not reverse the overall gradient flow for the original minimax problem (\ref{gan_form}). 
\end{claim}
A counterexample of the Non-Reversing-Flow Condition is to choose $\Gamma = -M^T$ where $M \triangleq {\frac{\partial \tilde{\nabla} f(w^{(k)})}{\partial w^{(k)}}}^T$ such that the new Jacobian becomes ${A} = - M^T M$. Now the Jacobian $A$ is a Hessian which has no complex eigenvalues and thus it could avoid the \textit{Phase Factor}. From (\ref{reg_update}), the updates become
\begin{align*}
\begin{split}
    w^{(k+1)} &= w^{(k)} - \eta M^T \tilde{\nabla} f(w^{(k)}) \\
    &= w^{(k)} - \eta \nabla^2 f(w^{(k)}) \nabla f(w^{(k)})
\end{split}
\end{align*}
As we know, in general, the objective $f(\phi, \theta)$ is not convex-concave in $\phi$ and $\theta$. 
For example, $f(\phi, \theta)$ becomes concave-concave in $\phi$ and $\theta$ near the equilibrium in the simple vanilla GAN example (\ref{obj_linear}).
Therefore, for any $w^{(k)}$ satisfying $\nabla^2_{\phi\phi} f(w^{(k)}) \prec 0$, particularly if assuming $\nabla^2_{\phi\phi} f(w^{(k)}) = -t^2 I$ where $t$ is a non-zero scalar, the update for $\phi$ becomes
$$
\phi^{(k+1)} = \phi^{(k)} + \eta t^2 \nabla_{\phi} f(w^{(k)}) - \eta \nabla^2_{\theta\phi} f(w^{(k)}) \nabla_{\theta} f(w^{(k)})
$$
According to the first two terms on the right-hand side of the above equation, it is actually a gradient flow of the generator $G_{\phi}$ maximizing the objective $f(\phi,\theta)$ instead. This partly explains why directly minimizing a surrogate loss $l(w) = \frac{1}{2} \|\nabla f(w^{(k)})\|^2$ does not work well in practice as has been observed by \cite{mescheder2017numerics}.

Next, we point out that ConOpt may also violate the Non-Reversing-Flow Condition in some cases. Similarly, for any point $w^{(k)}$ satisfying $\nabla^2_{\phi\phi} f(w^{(k)}) \prec 0$, particularly if we assume $\nabla^2_{\phi\phi} f(w^{(k)}) = -t^2 I$, the update for $\phi$ in (\ref{regularize_both}) for ConOpt becomes
\begin{align*}
    \begin{split}
        \phi^{(k+1)} =& \phi^{(k)} + \eta (\gamma t^2-1) \nabla_{\phi} f(w^{(k)}) \\
        &- \eta \gamma \nabla^2_{\theta\phi} f(w^{(k)}) \nabla_{\theta} f(w^{(k)})
    \end{split}
\end{align*}
If $\gamma t^2>1$, it is also a gradient flow of the generator $G_{\phi}$ maximizing the objective $f(\phi, \theta)$ instead. 
Note that the Hessian $\nabla^2_{\phi\phi} f(w^{(k)})$, introduced by ConOpt to the parameter updates, serves as the root cause of violating Condition \ref{claim_gflow}.
This might also partly explains why ConOpt is less robust than our proposed method in some experiments. Even worse, as $\gamma$ increases, it is more likely for ConOpt to reverse the gradient flow. It intuitively explains why $\gamma$ should be kept relatively small for ConOpt.

\section{Proofs in Section \ref{sec_gr}}

\subsection{Proof of Theorem \ref{thm_prev_meth}}
\label{proof_thm_prev_meth}

\textit{Proof}. we revisit each of these three regularization methods by evaluating and analyzing the eigenvalues of their Jacobians in the simple vanilla GAN example separately.

\textbf{Only regularizing generator.} 
The regularized updates for generator become
\begin{align} 
\label{regularize_G}
\begin{split}
\phi^{(k+1)} &= \phi^{(k)} - \eta \nabla_{\phi} f(w^{(k)}) - \frac{1}{2} \eta\gamma \nabla_{\phi} \left\Vert \nabla_{\theta} f(w^{(k)}) \right\Vert ^2
\end{split}
\end{align}
In the simple vanilla GAN example, from (\ref{hessian}) in Lemma \ref{lemma2}, $\frac{\partial \tilde{\nabla} f(w^{(k)})}{\partial w^{(k)}} = \begin{bmatrix}
	0 & -\frac{1}{2} I \\
	\frac{1}{2} I  & -\frac{1}{2} \left(\sigma^2 I + vv^T\right) 
\end{bmatrix}$. Also the regularization matrix becomes $\Gamma = \begin{bmatrix}
I & \frac{\gamma}{2} I \\
0 & I 
\end{bmatrix}$.
Thus, for all points in $B_{\delta}(w^*)$, the Jacobian is 
\begin{align*}
\begin{split}
        A &= \Gamma {\frac{\partial \tilde{\nabla} f(w^{(k)})}{\partial w^{(k)}}}^T \\
    & = \begin{bmatrix}
I & \frac{\gamma}{2} I \\
0 & I 
\end{bmatrix} \begin{bmatrix}
0 & \frac{1}{2} I \\
-\frac{1}{2} I  & -\frac{1}{2} \left(\sigma^2 I + vv^T\right) 
\end{bmatrix} \\
& = \begin{bmatrix}
-\frac{\gamma}{4}I & \frac{1}{2}I - \frac{\gamma}{4} \left(\sigma^2 I + vv^T \right) \\
-\frac{1}{2} I & -\frac{1}{2} \left(\sigma^2 I + vv^T\right)
\end{bmatrix}
\end{split}
\end{align*}
By definition of eigenvalues, let $Ay=\lambda y$ where $y=\begin{bmatrix}
y_1 \\ y_2
\end{bmatrix} \neq 0$, then
\begin{align}
-\frac{\gamma}{4}y_1 + \left(\frac{1}{2}I - \frac{\gamma}{4}(\sigma^2 I + vv^T) \right) y_2 = \lambda y_1 \label{regg_eig_1}\\
-\frac{1}{2} y_1 - \frac{1}{2} \left(\sigma^2 I + vv^T\right) y_2 = \lambda y_2 \label{regg_eig_2}
\end{align}
From (\ref{regg_eig_1}) we have $y_1 = \frac{1}{\lambda + \frac{\gamma}{4}} \left(\frac{1}{2}I - \frac{\gamma}{4}(\sigma^2 I + vv^T) \right) y_2$ (note that $\lambda \neq -\frac{\gamma}{4}$; otherwise, we get $y=0$). Plugging it into (\ref{regg_eig_2}) yields
\begin{align} \label{regg_eq}
- \lambda vv^Ty_2 = \left(2 \lambda^2 + (\frac{1}{2}\gamma+  \sigma^2 ) \lambda + \frac{1}{2} \right) y_2
\end{align}
Similarly, we can also solve (\ref{regg_eq}) in two cases yielding the eigenvalues of the Jacobian as follows,
\begin{align} \label{eigenval1}
	\begin{split}
	&\lambda_{1,2}({A}) = \frac{-\left(\sigma^2 + \frac{\gamma}{2}\right) \pm {\sqrt{\left(\sigma^2 + \frac{\gamma}{2}\right)^2-4  }}}{4}, \\
	&\lambda_{3,4}({A}) = \frac{-\left(\beta^2+ \frac{\gamma}{2} \right) \pm {\sqrt{\left(\beta^2 + \frac{\gamma}{2} \right)^2-4}}}{4}
	\end{split}
\end{align}

As we can see, the resulting ${\zeta}= \sqrt{(\frac{2}{\sigma^2+\frac{\gamma}{2}})^2-1}$ for $\sigma^2+\frac{\gamma}{2} < 2$, which means increasing $\gamma$ will decrease ${\zeta}$ and thus could alleviate the impact of the \textit{Phase Factor} by Corollary \ref{coro1}. However, the resulting ${\tau} = \frac{\left( (\beta^2+ \frac{\gamma}{2} ) + {\sqrt{(\beta^2 + \frac{\gamma}{2})^2-4}} \right)^2}{4}$ for $\beta^2+\frac{\gamma}{2} > 2$, which means increasing $\gamma$ will also increase ${\tau}$ and thus the impact of \textit{Conditioning Factor} will not be alleviated but become much severer by Corollary \ref{coro2}. 
Therefore, if the \textit{Conditioning Factor} is the main obstacle for the GAN convergence (for example, $\|v\|$ is sufficiently large in the simple vanilla GAN example), only regularizing generator as in (\ref{regularize_G}) will make the convergence performance of the GAN training worse.

\textbf{Only regularizing discriminator.}
The regularized updates for the discriminator become
\begin{align} 
\label{regularize_D}
\begin{split}
\theta^{(k+1)} &= \theta^{(k)} + \eta \nabla_{\theta} f(w^{(k)})  - \frac{1}{2} \eta\gamma \nabla_{\theta} \left\Vert \nabla_{\phi} f(w^{(k)}) \right\Vert ^2
\end{split}
\end{align}
Similarly in the simple vanilla GAN example, the regularziation matrix becomes $\Gamma = \begin{bmatrix}
I & 0 \\
-\frac{\gamma}{2} I & I 
\end{bmatrix}$. For any point in $B_{\delta}(w^*)$, the Jacobian is 
\begin{align*}
\begin{split}
    A &= \Gamma {\frac{\partial \tilde{\nabla} f(w^{(k)})}{\partial w^{(k)}}}^T \\
    &= \begin{bmatrix}
I & 0 \\
-\frac{\gamma}{2} I & I 
\end{bmatrix} \begin{bmatrix}
0 & \frac{1}{2} I \\
-\frac{1}{2} I  & -\frac{1}{2} \left(\sigma^2 I + vv^T\right) 
\end{bmatrix} \\
&= \begin{bmatrix}
	0 & \frac{1}{2} I \\
	-\frac{1}{2} I & -\frac{1}{2} \left(\left(\sigma^2 + \frac{\gamma}{2} \right) I + vv^T\right)
	\end{bmatrix}
\end{split}
\end{align*}
Then by following from the exact proof of Theorem \ref{thm_simgd} after replacing $\sigma^2$ in the Jacobian of (\ref{A_simgd}) by $\sigma^2+ \frac{\gamma}{2}$, we can get the eigenvalues of the Jacobian as follows, 
\begin{align} \label{eigenval_d}
	\begin{split}
	& \lambda_{1,2}({A}) = \frac{-\left(\sigma^2 + \frac{\gamma}{2}\right) \pm {\sqrt{\left(\sigma^2 + \frac{\gamma}{2}\right)^2-4  }}}{4}, \\
	& \lambda_{3,4}({A}) = \frac{-\left(\beta^2+ \frac{\gamma}{2} \right) \pm {\sqrt{\left(\beta^2 + \frac{\gamma}{2} \right)^2-4}}}{4}
	\end{split}
\end{align}

As the eigenvalues here are exactly the same with (\ref{eigenval1}), the local convergence properties of only regularizing the discriminator are identical to those of only regularizing the generator. Similarly, if \textit{Conditioning Factor} becomes the main obstacle for GAN convergence, only regularizing discriminator as in (\ref{regularize_D}) will make the convergence performance of the GAN training worse.

\textbf{Consensus optimization (ConOpt).} 
The regularized updates for the generator and discriminator are
\begin{align} 
\label{regularize_both}
\begin{split}
w^{(k+1)} = w^{(k)} + \eta \tilde{\nabla} f(w^{(k)})  - \frac{1}{2}\eta\gamma \nabla \left\Vert \nabla f(w^{(k)}) \right\Vert ^2
\end{split}
\end{align}
Since for ConOpt, it is a little bit tricky to obtain the eigenvalues of its Jacobian directly, we turn to comparing the eigenvalues of it Jacobian with those of the Jacobian for SimGD.

First, we define $M \triangleq {\frac{\partial \tilde{\nabla} f(w^{(k)})}{\partial w^{(k)}}}^T$. For SimGD, we know its Jacobian is $M$. For ConOpt, since the regularization matrix $\Gamma = I - \gamma M^T$, its Jacobian is
\begin{align} \label{A_conopt}
    A = \Gamma M = M - \gamma M^T M
\end{align}
Then, we define $\overline{\lambda}(M)$ and $\underline{\lambda}(M)$ as the two eigenvalues of $M$ with the largest and smallest absolute values, respectively, and the similar definitions of $\overline{\lambda}(A)$ and $\underline{\lambda}(A)$ apply to $A$.
Thus, the condition numbers of $A$ and $M$ are $\tau(A) \triangleq \frac{|\overline{\lambda}(A)|}{|\underline{\lambda}(A)|}$ and $\tau(M) \triangleq \frac{|\overline{\lambda}(M)|}{|\underline{\lambda}(M)|}$, respectively.

If $\sigma^2 < 2$ and $\beta^2>2$, from Theorem \ref{thm_simgd} we know for any point in $B_{\delta}(w^{*})$, the Jacobian for SimGD satisfies $|\lambda_{1,2}(M)| = \frac{1}{2}$, $|\lambda_{3}(M)| =  \frac{\beta^2+\sqrt{(\beta^2)^2-4}}{4} > \frac{1}{2}$ and $|\lambda_{4}(M)| =  \frac{\beta^2-\sqrt{(\beta^2)^2-4}}{4} < \frac{1}{2}$. 
Thus, $\overline{\lambda}(M) = \lambda_{3}(M)$ and $\underline{\lambda}(M) = \lambda_{4}(M)$, which are both negative values. 

By definition of eigenvalues, we have $M y_1 = \overline{\lambda}(M) y_1$ and $M y_2 = \underline{\lambda}(M) y_2$ where $y_1$ and $y_2$ are two normalized eigenvectors of $M$ with unit length. Thus, $y_1^T M y_1 = \overline{\lambda} (M)$ and $y_2^T M y_2 = \underline{\lambda} (M)$. From (\ref{A_conopt}), we have $y_1^T M y_1 = \overline{\lambda}(M) -\gamma {\overline{\lambda}(M)}^2$ and $y_2^T M y_2 = \underline{\lambda}(M)-\gamma {\underline{\lambda}(M)}^2$.  
From the definition of $\overline{\lambda}(A)$ and $\underline{\lambda}(A)$, we know $| y_1^T A y_1 | \leq |\overline{\lambda}(A)|$ and $| y_2^T A y_2 | \geq |\underline{\lambda} (A)|$, then $| \overline{\lambda}(M) -\gamma {\overline{\lambda}(M)}^2 | \leq |\overline{\lambda}(A)|$ and $ | \underline{\lambda}(M)-\gamma {\underline{\lambda}(M)}^2 | \geq |\underline{\lambda}(A)|$. Combining the two inequalities yields
\begin{align}
    \tau(A) \geq \tau(M) \cdot \frac{1 + \gamma |\overline{\lambda}(M)|}{1 + \gamma |\underline{\lambda}(M)|}
\end{align}

Define by $\Delta(\gamma) \triangleq \frac{1 + \gamma |\overline{\lambda}(M)|}{1 + \gamma |\underline{\lambda}(M)|}$. As $|\overline{\lambda}(M)| > |\underline{\lambda}(M)| > 0$ and $\gamma > 0$, we have $\Delta(\gamma) > 1$, which means $\tau(A) > \tau(M)$ for any $\gamma>0$. Even worse, since the derivative $\Delta'(\gamma) = \frac{\overline{\lambda}(M) - \underline{\lambda}(M)}{(1-\gamma \underline{\lambda}(M))^2} > 0$, when $\gamma$ increases, $\Delta(\gamma)$ also increases.
Thus, by using ConOpt, the impact of \textit{Conditioning Factor} is not alleviated but becomes more severe by Corollary \ref{coro2}. Furthermore, the Jacobian will be worse-conditioned as $\gamma$ increases.
Therefore, although ConOpt could alleviate the impact of the \textit{Phase Factor} as shown in \cite{mescheder2017numerics}, it will make the GAN convergence performance worse if the \textit{Conditioning Factor} becomes the main obstacle for the GAN convergence.  
 
From the above analysis, all these three gradient-based regularization methods cannot alleviate the \textit{Phase Factor} and \textit{Conditioning Factor} simultaneously. 
\hfill  $\square$

\vspace{-5pt}

\subsection{Proof of Theorem \ref{thm_prop}}
\label{proof_thm_prop}

\textit{Proof.} 
When applying the proposed Jacobian regularization in the simple vanilla GAN example (\ref{obj_linear}), the regularization matrix becomes $\Gamma = \begin{bmatrix}
	I & \frac{\gamma}{2}I \\
	-\frac{\gamma}{2} I & I
\end{bmatrix}$. Therefore, for any point in $B_{\delta}(w^*)$, 
\begin{align*}
\begin{split}
    A &= \Gamma {\frac{\partial \tilde{\nabla} f(w^{(k)})}{\partial w^{(k)}}}^T \\
    &=  \begin{bmatrix}
I & \frac{\gamma}{2} I \\
-\frac{\gamma}{2} I  & I 
\end{bmatrix} \begin{bmatrix}
0 & \frac{1}{2} I \\
-\frac{1}{2} I  & -\frac{1}{2} \left(\sigma^2 I + vv^T\right) 
\end{bmatrix} \\
&= \begin{bmatrix}
-\frac{\gamma}{4}I & \frac{1}{2}I - \frac{\gamma}{4} \left(\sigma^2 I + vv^T \right) \\
-\frac{1}{2} I & -\frac{\gamma}{4}I -\frac{1}{2} \left(\sigma^2 I + vv^T\right)
\end{bmatrix}
\end{split}
\end{align*}
By definition of eigenvalues, let ${A} y=\lambda y$ where $y=\begin{bmatrix}
y_1 \\ y_2
\end{bmatrix} \neq 0$, then
\begin{align}
-\frac{\gamma}{4}y_1 + \left(\frac{1}{2}I - \frac{\gamma}{4}(\sigma^2 I + vv^T) \right) y_2 = \lambda y_1 \label{prop_eig_1}\\
-\frac{1}{2} y_1 - \frac{\gamma}{4}y_2 - \frac{1}{2} \left(\sigma^2 I + vv^T\right) y_2 = \lambda y_2 \label{prop_eig_2}
\end{align}
Similarly, $\lambda \neq 0$, otherwise, we get $y=0$ which violates the definition of eigenvectors. By applying $(\ref{prop_eig_1}) - (\ref{prop_eig_2})*\frac{\gamma}{2}$, we have $y_1 = \frac{1}{\lambda} \left(\frac{\gamma}{2}\lambda+\frac{\gamma^2}{8}+\frac{1}{2}\right) y_2$. Plugging it into (\ref{prop_eig_2}) yields 
\begin{align} \label{prop_eq}
- \lambda vv^Ty_2 = \left(2 \lambda^2 + (\gamma+  \sigma^2 ) \lambda + \frac{\gamma^2}{8}+\frac{1}{2} \right) y_2
\end{align}
Similarly, we can solve (\ref{prop_eq}) in two cases yielding the desired results by following the same process in the proof of Theorem \ref{thm_simgd}. 
\hfill $\square$

\subsection{Proof of Corollary \ref{coro_prop}}
\label{proof_coro_prop}

\textit{Proof.} From Theorem \ref{thm_prop} we know for $\sigma^2 < 2$, $\lambda_{1,2}(A)$ are complex eigenvalues only if $\gamma < \frac{2}{\sigma^2}-\frac{\sigma^2}{2}$. According to the above definition of $\zeta$, we get
\begin{align} \label{zeta_prop}
    \zeta = \begin{cases}
    \sqrt{h_1(\gamma)-1}, & \gamma < \frac{2}{\sigma^2}-\frac{\sigma^2}{2}\\
    0, & \gamma \geq \frac{2}{\sigma^2}-\frac{\sigma^2}{2}
    \end{cases} 
\end{align}
where $h_1(\gamma) = \frac{\gamma^2+4}{(\sigma^2+{\gamma})^2} > 1$. 
Since the derivative of $h_1(\gamma)$ satisfies $h'_1(\gamma) = \frac{2(\gamma+\sigma^2)(\sigma^2 \gamma - 4)}{(\sigma^2+{\gamma})^4} < 0$ and $\zeta$ is a monotonically increasing function of $h_1(\gamma)$ for $\gamma < \frac{2}{\sigma^2}-\frac{\sigma^2}{2}$, $\zeta$ is a monotonically decreasing function of $\gamma$ for $\gamma < \frac{2}{\sigma^2}-\frac{\sigma^2}{2}$.
As $\zeta = 0$ if $\gamma \geq \frac{2}{\sigma^2}-\frac{\sigma^2}{2}$, by the continuity of the function in (\ref{zeta_prop}), we have $\zeta$ is a monotonically decreasing function of $\gamma$ where $\zeta \to 0$ as $\gamma \to \infty$. It means that we can increase $\gamma$ to alleviate the impact of the \textit{Phase Factor}.

Furthermore, from Theorem \ref{thm_prop} we know for $\beta^2 > 2$, 
\begin{align} \label{tau_prop}
    {\tau} = \left(\sqrt{h_2(\gamma)}+\sqrt{h_2(\gamma)^2-1}\right)^2
\end{align}
where $h_2(\gamma) = \frac{(\beta^2+\gamma)^2}{\gamma^2+4} > 1$.
Since the derivative of $h_2(\gamma)$ satisfies $h'_2(\gamma) = \frac{2(\gamma+\beta^2)(4 - \beta^2 \gamma)}{(\gamma+4)^4} < 0$ for $\gamma > \frac{4}{\beta^2}$ and $\tau$ is a monotonically increasing function of $h_2(\gamma)$, $\tau$ is a monotonically decreasing function of $\gamma$ for $\gamma > \frac{4}{\beta^2}$. As $\beta^2 > 2$, then $\frac{4}{\beta^2} < 2$ and we thus can safely replace the above condition $\gamma > \frac{4}{\beta^2}$ by $\gamma \geq 2$. In the limit of $\gamma \to \infty$, we have $h_2(\gamma) \to 1$ and thus from (\ref{tau_prop}) $\tau \to 1$. It means that we can increase $\gamma$ to alleviate the impact of the \textit{Conditioning Factor} for all $\gamma > \frac{4}{\beta^2}$.

Therefore, it is reasonable to keep increasing the tunable parameter $\gamma$ so as to alleviate or even eliminate both the \textit{Phase Factor} and \textit{Conditioning Factor} simultaneously, which demonstrates the advantages of JARE. 
\hfill $\square$

\section{Proof in Section \ref{sec_general}}

\subsection{Proof of Lemma \ref{general_lemma_A}}
\label{proof_general_lemma}

Although the proof is very similar to \cite{mescheder2018training}, we provide the proof details for completeness. 

Since we know the objective is
\begin{align*}
\begin{split}
f(\phi, \theta) \triangleq \mathbb{E}_{x \sim P_r}   [g_1(D_{\theta}(x))]  + \mathbb{E}_{z \sim P_{0}}[g_2(D_{\theta}(G_{\phi}(z)))]
\end{split}
\end{align*}
By taking its derivative w.r.t. $\phi$ and $\theta$ at the equilibrium $(\phi^*, \theta^*)$, respectively, we have 
\begin{align} \label{der_phi_general}
    \begin{split}
        \nabla_{\phi} f(\phi^*, \theta^*) = \mathbb{E}_{z \sim P_0} [ & g'_2(D_{\theta^*}(x))  \nabla_{\phi} G_{\phi^*}(z) \\
        & \cdot \nabla_x D_{\theta^*}(x)] |_{x = G_{\phi^*}(z)}
    \end{split}
    \end{align}
    \begin{align} \label{der_theta_general}
    \begin{split}
        \nabla_{\theta} f(\phi^*, \theta^*) = & \mathbb{E}_{x \sim P_r}[  g'_1(D_{\theta^*}(x)) \nabla_{\theta} D_{\theta^*}(x)] \\
        & + \mathbb{E}_{x \sim P_{\phi^*}}[g'_2(D_{\theta^*}(x)) \nabla_{\theta}D_{\theta^*}(x)]
    \end{split}
\end{align}
Since the Jacobian $A$ at $(\phi^*, \theta^*)$ in general GANs trained via SimGD are given by 
$$
A = \begin{bmatrix}
			-\nabla^2_{\phi\phi}f(\phi^*, \theta^*) & -\nabla^2_{\phi\theta}f(\phi^*, \theta^*) \\
			\nabla^2_{\theta\phi}f(\phi^*, \theta^*) & \nabla^2_{\theta\theta}f(\phi^*, \theta^*)
    \end{bmatrix}
$$
First, from Assumption 1 we know that $D_{\theta^*}(x) = 0$ for some local neighborhood of any $x \in \mathcal{X}$, which means we also have $\nabla_x D_{\theta^*}(x) = 0$ and $\nabla^2_{xx} D_{\theta^*}(x) = 0$ for any $x \in \mathcal{X}$. By taking the derivative of (\ref{der_phi_general}) w.r.t. $\phi$ at the equilibrium $(\phi^*, \theta^*)$ and using $\nabla_x D_{\theta^*}(x) = 0$ and $\nabla^2_{xx} D_{\theta^*}(x) = 0$ for any $x \in \mathcal{X}$, we have
\begin{align*}
    \nabla^2_{\phi\phi}f(\phi^*, \theta^*) = 0
\end{align*}

By taking the derivative of (\ref{der_theta_general}) w.r.t. $\phi$ at the equilibrium $(\phi^*, \theta^*)$, we have
\begin{align*}
    \begin{split}
        \nabla^2_{\phi\theta}f(\phi^*, \theta^*) = \mathbb{E}_{z \sim P_{0}}[ & g''_2(D_{\theta^*}(x)) \nabla_{\phi} G_{\phi^*}(z) \\
        & \cdot \nabla^2_{x \theta}D_{\theta^*}(x)] |_{x = G_{\phi^*}(z)} \\
        \mathop = \limits^{(a)} g''_2(0) \mathbb{E}_{z \sim P_{0}}[ & \nabla_{\phi} G_{\phi^*}(z) \nabla^2_{x \theta}D_{\theta^*}(x)] |_{x = G_{\phi^*}(z)}
    \end{split}
\end{align*}
where $(a)$ is from the assumption that $D_{\theta^*} = 0$.

By taking the derivative of (\ref{der_theta_general}) w.r.t. $\theta$, respectively, at the equilibrium $(\phi^*, \theta^*)$, we have
\begin{align*}
    \begin{split}
        \nabla^2_{\theta \theta}f(\phi^*, \theta^*) 
        \mathop = \limits^{(a)} & \mathbb{E}_{x \sim P_r}[ (g'_1(0) + g'_2(0)) \nabla^2_{\theta \theta} D_{\theta^*}(x) \\
        & + (g''_1(0) + g''_2(0))\nabla_{\theta} D_{\theta^*}(x) D_{\theta^*}(x)^T] \\
        \mathop = \limits^{(b)}  (g''_1(0) & + g''_2(0))  \mathbb{E}_{x \sim P_r}[\nabla_{\theta} D_{\theta^*}(x) D_{\theta^*}(x)^T]
    \end{split}
\end{align*}
where $(a)$ is from Assumption 1 that $P_r = P_{\phi^*}$ and $D_{\theta^*} = 0$, $(b)$ is from Assumption 2 that $g'_1(0) =  -g'_2(0) $.

Finally, by setting $P = \nabla^2_{\phi\theta}f(\phi^*, \theta^*)$ and $Q = \nabla^2_{\theta \theta}f(\phi^*, \theta^*)$, we get the results. \hfill $\square$

\subsection{Proof of Theorem \ref{thm_general_simgd}}
\label{proof_thm_general_simgd}

Since the Jacobian $A = \begin{bmatrix}
				0 & -P \\
				P^T  & Q
			\end{bmatrix}$,
by the definition of eigenvector equations we have 
\begin{align*}
    \begin{bmatrix}
				0 & -P \\
				P^T  & Q
			\end{bmatrix} \begin{bmatrix}
				y_1 \\
				y_2
			\end{bmatrix} = \lambda \begin{bmatrix}
				y_1 \\
				y_2
			\end{bmatrix}
\end{align*}
where $y_1$, $y_2$ and $\lambda$ may be complex-valued. We can rewrite the above equations as follows:
\begin{align}
-P y_2 &= \lambda y_1 \label{thm_general_eq1}\\
P^T y_1 + Q y_2 &= \lambda y_2 \label{thm_general_eq2}
\end{align}
Plugging Eq. (\ref{thm_general_eq1}) into Eq. (\ref{thm_general_eq2}) yields
\begin{align} \label{thm_general_eq3}
    \lambda^2 y_2 - \lambda Q y_2 + P^T P y_2 = 0
\end{align}
\textbf{Case 1.} Consider $y_2 = 0$, then 1) if $P$ has the full column rank, we have $y_1 = 0$ as well which violates the definition of eigenvectors; 2) if $P$ does not have the full column rank, we have $\lambda = 0$. 

\textbf{Case 2.} Consider $y_2 \neq 0$, we can multiply Eq. (\ref{thm_general_eq3}) by $y_2^H$ (conjugate transpose of $y_2$) and then divide by $\|y_2\|^2$ in both sides, yielding
\begin{align} \label{case_2}
    \lambda^2 - \frac{y_2^H Q y_2}{\|y_2\|^2} \lambda + \frac{y_2^H P^T P y_2}{\|y_2\|^2} = 0
\end{align}

Let $a_1 = \frac{y_2^H Q y_2}{\|y_2\|^2}$ and $a_2 = \frac{y_2^H P^T P y_2}{\|y_2\|^2}$, by solving the equation $\lambda^2 - a_1 \lambda + a_2 = 0$, we can get the results of (\ref{lam_A_general}). Next, we need to evaluate $a_1$ and $a_2$, respectively.

First note that $a_1 = \frac{y_2^H Q y_2}{\|y_2\|^2}$ is actually the \textit{Rayleigh Quotient} of $Q$. Therefore, we consider a set of $m$ orthonormal eigenvectors $\{x_{Q,i}\}_{i=1}^m$ corresponding to its $m$ eigenvalues $\{\lambda_i(Q)\}_{i=1}^m$, and then there exists some set of $m$ coefficients $\{b_i\}_{i=1}^n$, such that 
\begin{align*}
    y_2 = \sum_{i=1}^m b_i x_{Q,i}
\end{align*}
where $b_i$ may be complex-valued. Thus, we have 
\begin{align*}
    Qy_2 = \sum_{i=1}^m b_i \lambda_i(Q) x_{Q,i}
\end{align*}
and 
\begin{align*}
    a_1 = \frac{\sum_{i=1}^m |b_i|^2 \lambda_i(Q) }{\sum_{i=1}^m | b_i |^2 } = \sum_{i=1}^m \alpha_i \lambda_i(Q)
\end{align*}
where we let $\alpha_i = \frac{|b_i|^2}{\sum_{i=1}^m | b_i |^2 }$ for $i = 1, \cdots, m$, which satisfies $\alpha_i \geq 0$ and $\sum_{i=1}^m \alpha_i = 1$.  

Similarly, as  $a_2 = \frac{y_2^H P^T P y_2}{\|y_2\|^2}$ is a \textit{Rayleigh Quotient} of $P^T P$, we have 
\begin{align*}
    a_2 = \sum_{i=1}^m \tilde{\alpha}_i \lambda_i(P^T P)
\end{align*}
with $\tilde{\alpha}_i$ satisfying $\tilde{\alpha}_i \geq 0$ and $\sum_{i=1}^m \tilde{\alpha}_i = 1$.

Finally, if $P$ does not have the full column rank, we can choose $y_2 \in \text{Null} (P)$ and $y_1 = 0$ such that $a_2 = 0$  and thus $\lambda = 0$ becomes a solution of Eq. (\ref{case_2}). Therefore, the analysis of Case 1 is a special case of Case 2.
\hfill $\square$

\subsection{Proof of Theorem \ref{thm_general_jr}} \label{proof_thm_general_jr}

From Lemma \ref{general_lemma_A}, we know that for JARE, the corresponding regularization matrix is $$\Gamma = \begin{bmatrix}
        I & -\gamma P \\
        \gamma P^T & I
    \end{bmatrix}
$$
Thus, the Jacobian becomes
\begin{align*}
\begin{split}
    A &= \Gamma \begin{bmatrix}
				0 & -P \\
				P^T  & Q
			\end{bmatrix} \\
	  &= -\gamma \begin{bmatrix}
				 P P^T & P Q \\
				0  & P^T P
			\end{bmatrix} + \begin{bmatrix}
				0 & -P \\
				P^T  & Q
			\end{bmatrix}
\end{split}
\end{align*}
In the limit of $\gamma \to \infty$, we have 
\begin{align*}
    A = -\gamma \begin{bmatrix}
				 P P^T & P Q \\
				0  & P^T P
			\end{bmatrix}
\end{align*}
Its eigenvalues $\lambda(A)$ are solutions of $\text{det}(\lambda I - A) = 0$. As a block upper triangular matrix, we have 
\begin{align*}
    \text{det}(\lambda I - A) = \text{det}(\lambda I + \gamma P^T P) \text{det}(\lambda I + \gamma P P^T)
\end{align*}
which means the eigenvalues of $A$ satisfy
\begin{align*}
    \lambda(A) = -\gamma \lambda(P^T P) \;\; \text{and} \;\;  \lambda(A) = -\gamma \lambda(P P^T)
\end{align*}
Also, since $P^T P$ and $P P^T$ have the same set of eigenvalues, we have 
\begin{align*}
    \lambda(A) = -\gamma \lambda(P^T P)
\end{align*}
as required. \hfill $\square$

\onecolumn

\section{More experimental results}

\subsection{More results on Isotropic Gaussian}
\label{more_syn_res}


\begin{figure} [H]
	\centering
	\begin{subfigure}[b]{0.32\textwidth}
		\centering
		\includegraphics[width=\textwidth]{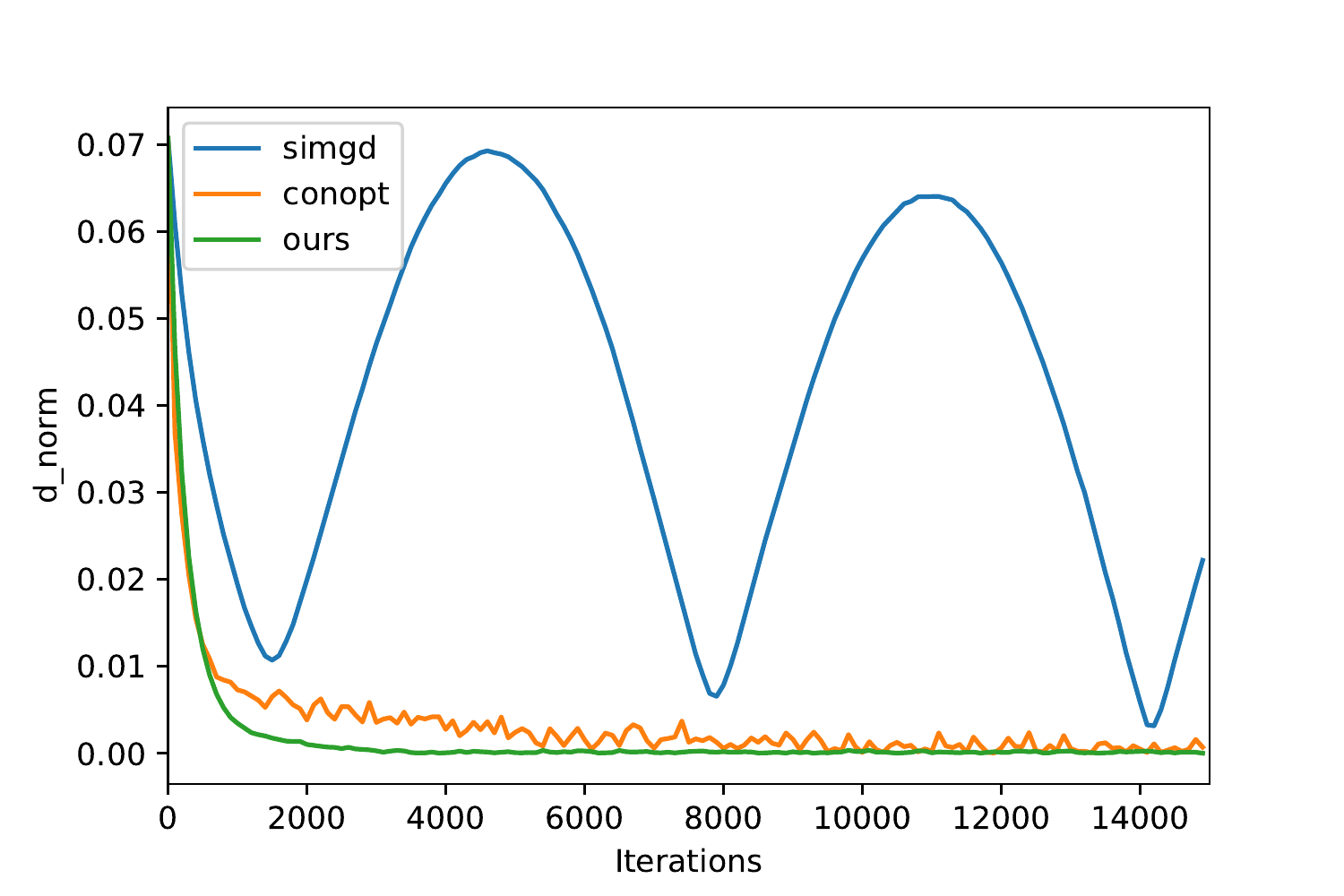}
	\end{subfigure}
	\begin{subfigure}[b]{0.32\textwidth}
		\centering
		\includegraphics[width=\textwidth]{IsotLin/d_norm_mu4.pdf}
	\end{subfigure}
	\begin{subfigure}[b]{0.32\textwidth}
		\centering
		\includegraphics[width=\textwidth]{IsotLin/d_norm_mu4.pdf}
	\end{subfigure}

	\begin{subfigure}[b]{0.32\textwidth}
		\centering
		\includegraphics[width=\textwidth]{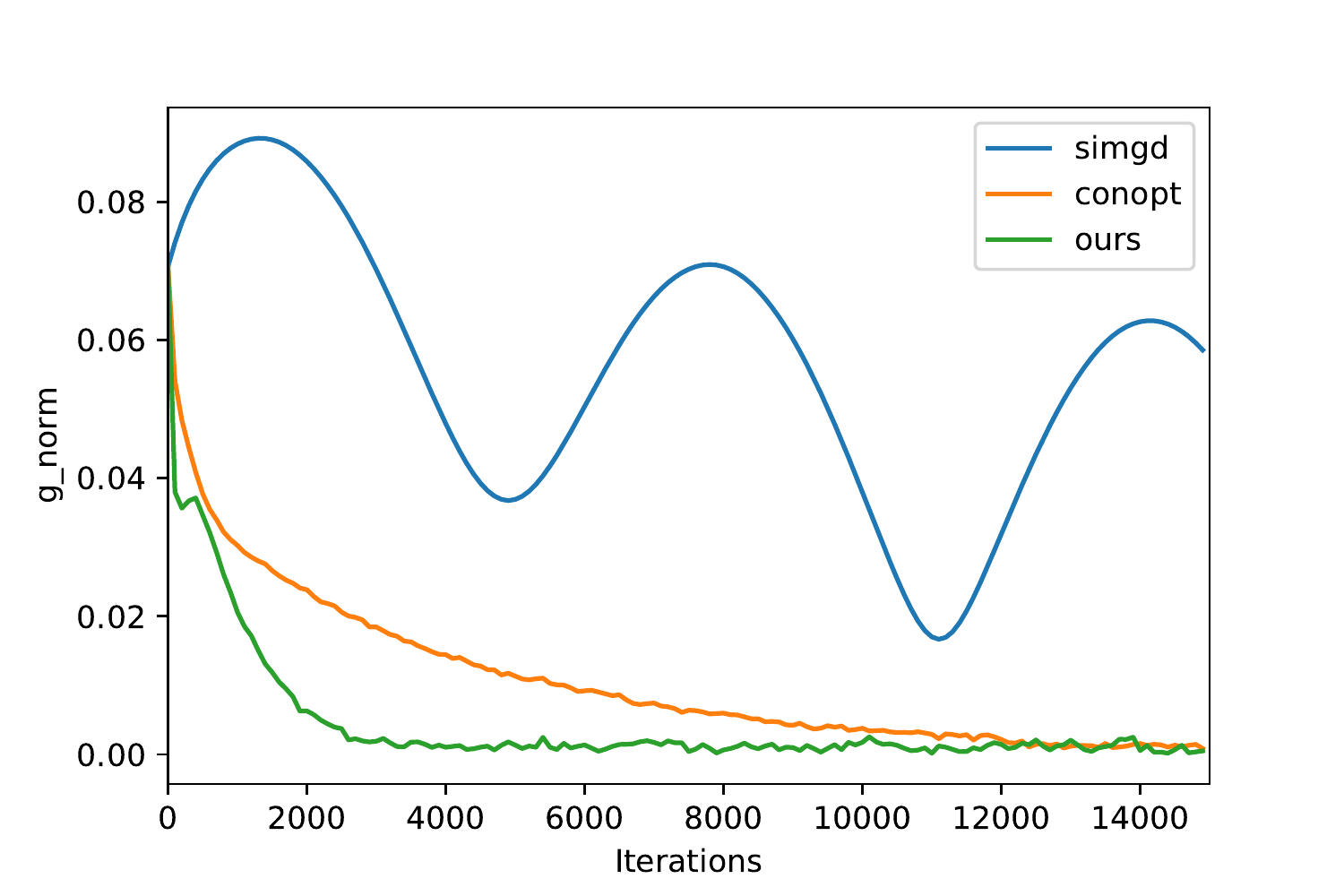}
		\caption{$\mu=2$}
	\end{subfigure}
	\begin{subfigure}[b]{0.32\textwidth}
		\centering
		\includegraphics[width=\textwidth]{IsotLin/g_norm_mu4.pdf}
		\caption{$\mu=4$}
	\end{subfigure}
	\begin{subfigure}[b]{0.32\textwidth}
		\centering
		\includegraphics[width=\textwidth]{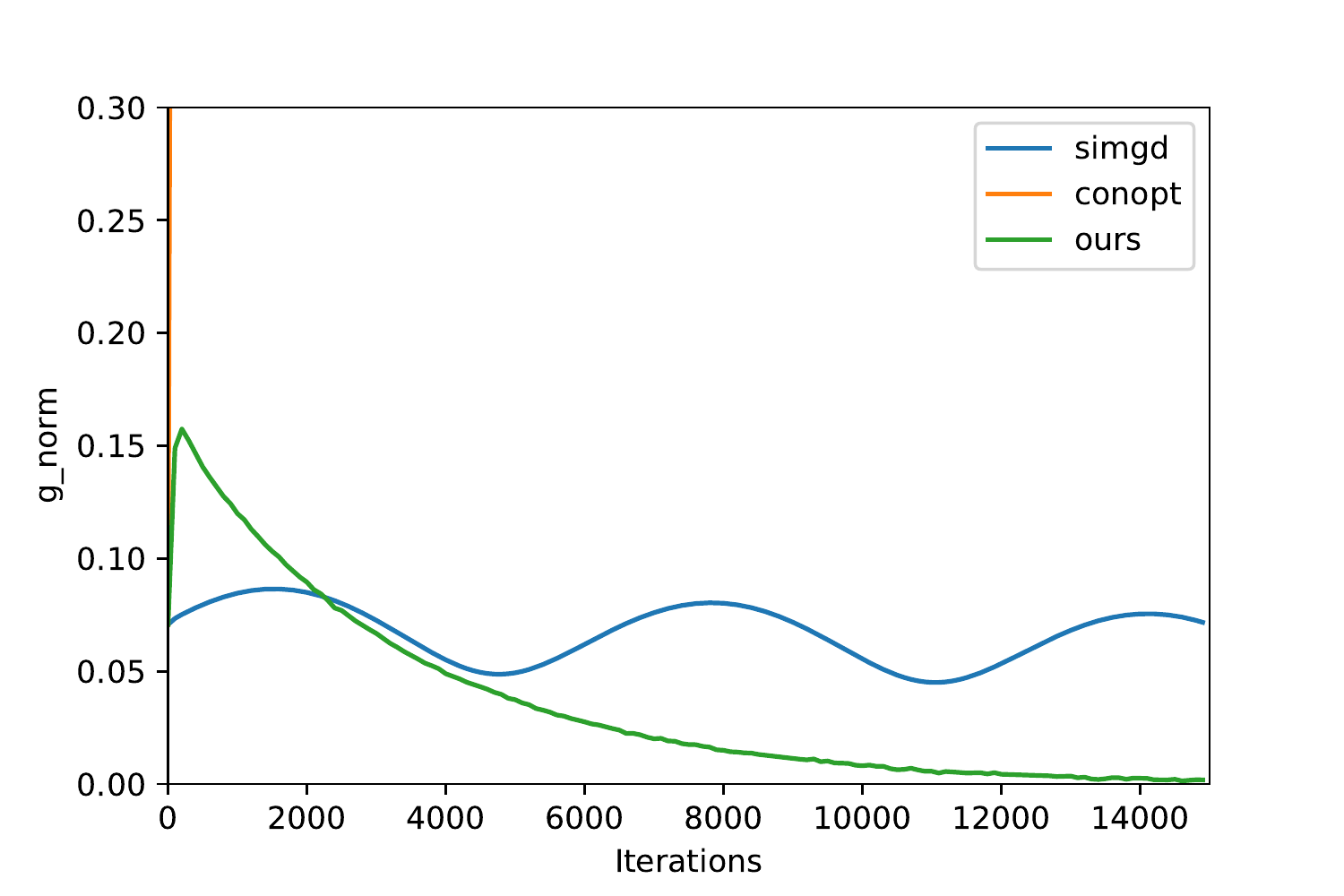}
		\caption{$\mu=6$}
	\end{subfigure}

	\caption{Training dynamics of SimGD, ConOpt and JARE for the discriminator (top row) and the generator (bottom row) with varying mean value $\mu$ where $\sigma = 0.2$. 
	Note that as $\mu$ increases, the convergence rate for either SimGD or ConOpt becomes slower. When $\mu = 6$, the generator training curve for the ConOpt directly blow up. } \label{isotLin_mu}
\end{figure}

\begin{figure} [H]
	\centering
	\begin{subfigure}[b]{0.32\textwidth}
		\centering
		\includegraphics[width=\textwidth]{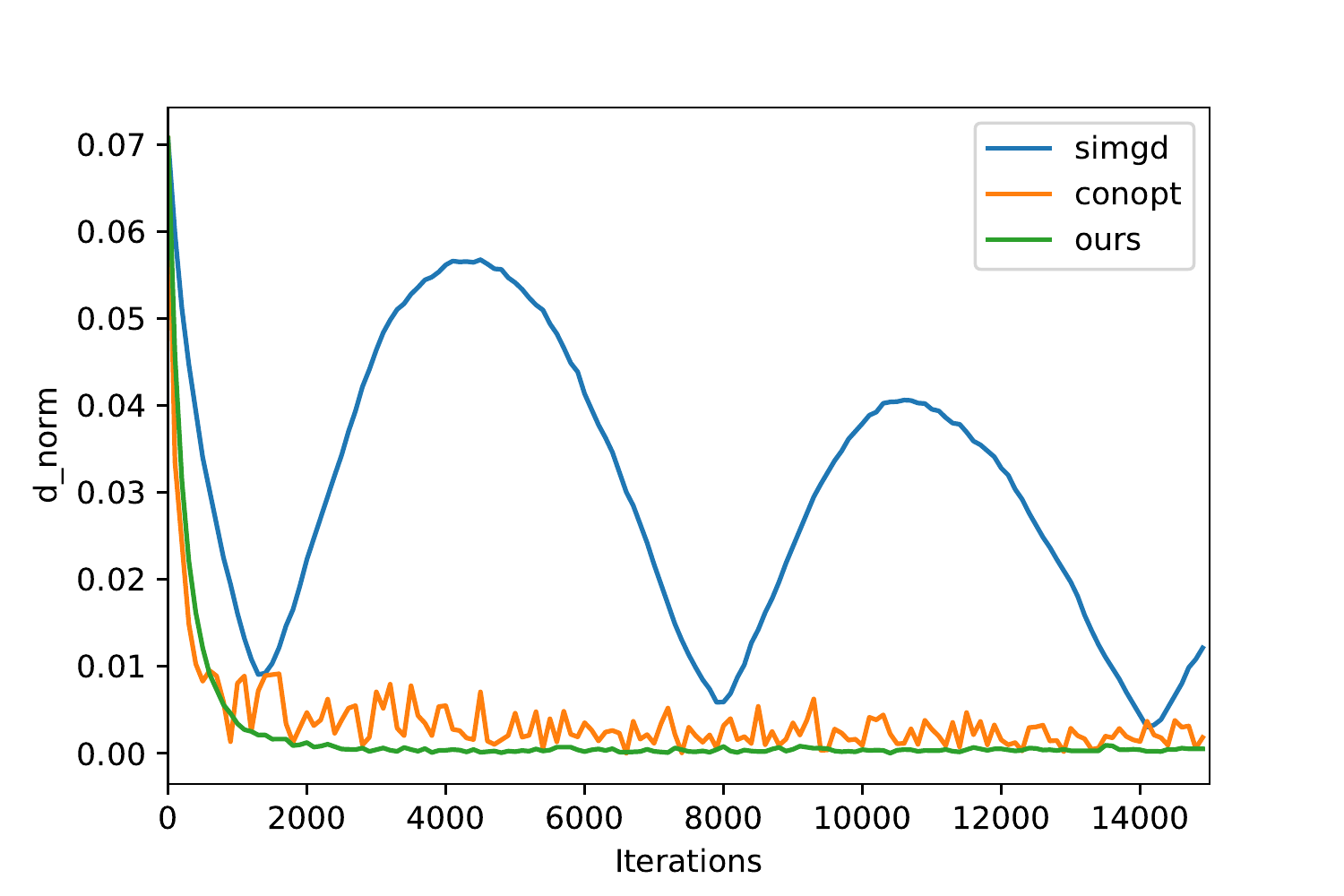}
	\end{subfigure}
	\begin{subfigure}[b]{0.32\textwidth}
		\centering
		\includegraphics[width=\textwidth]{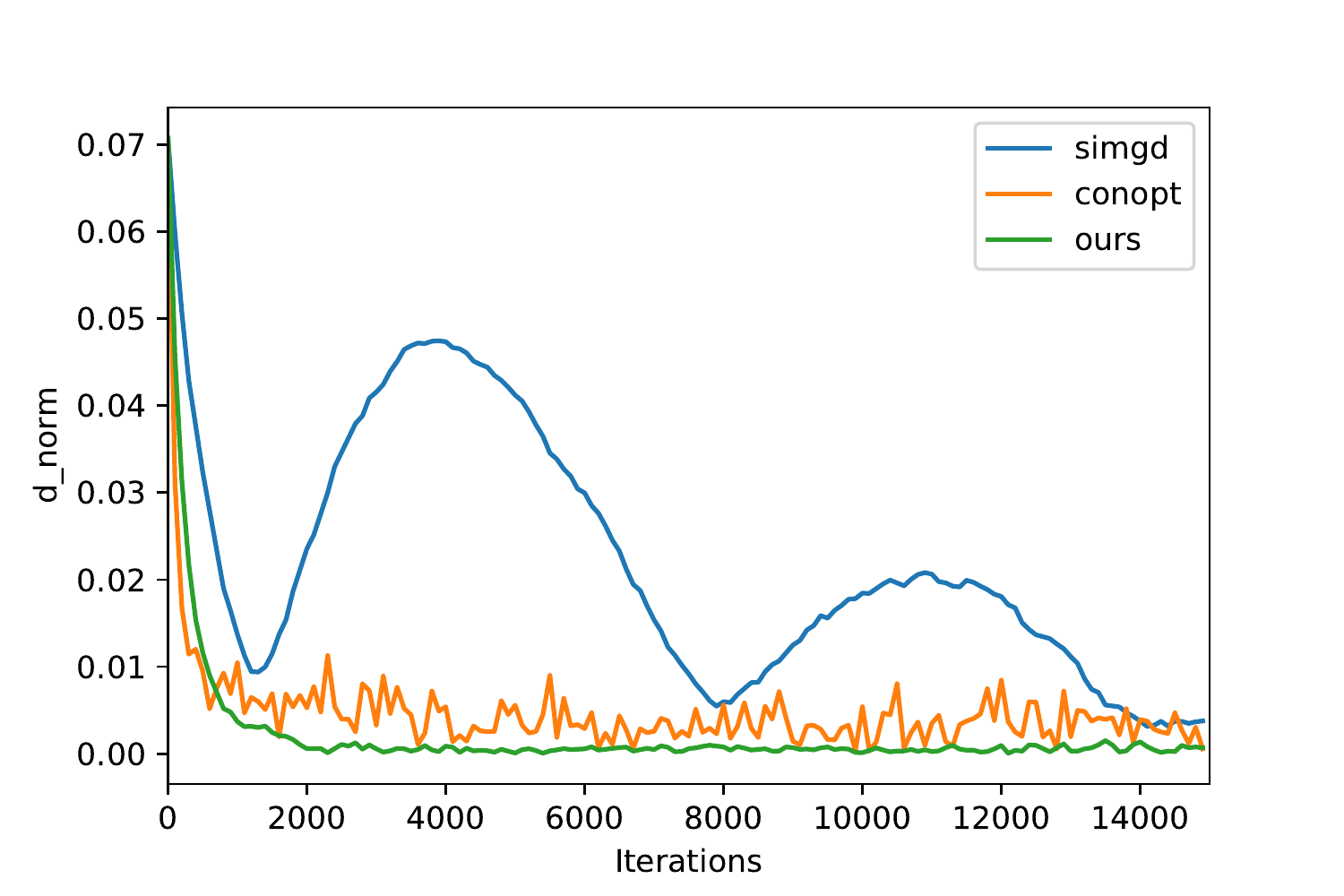}
	\end{subfigure}
	\begin{subfigure}[b]{0.32\textwidth}
		\centering
		\includegraphics[width=\textwidth]{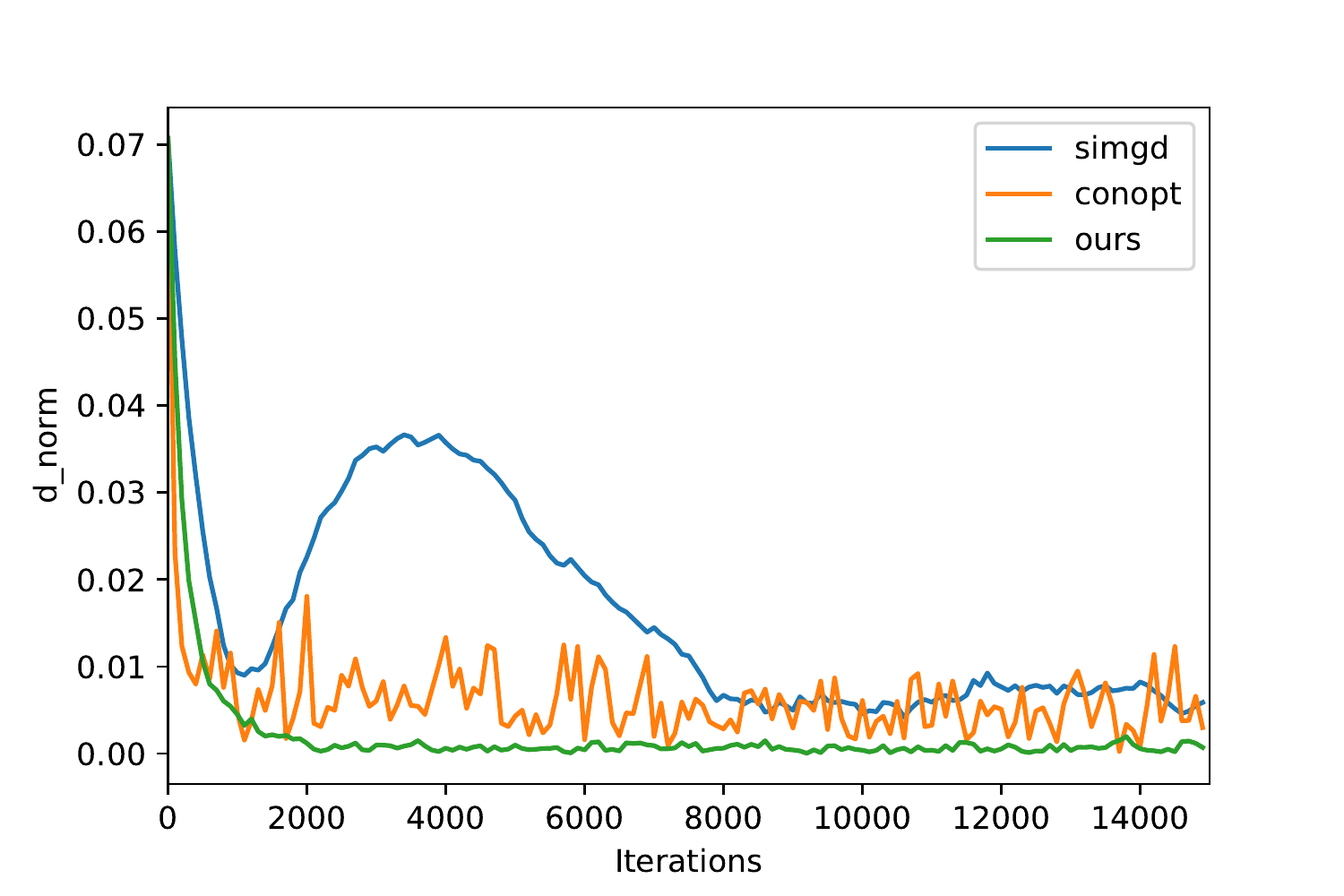}
	\end{subfigure}
	
	\begin{subfigure}[b]{0.32\textwidth}
		\centering
		\includegraphics[width=\textwidth]{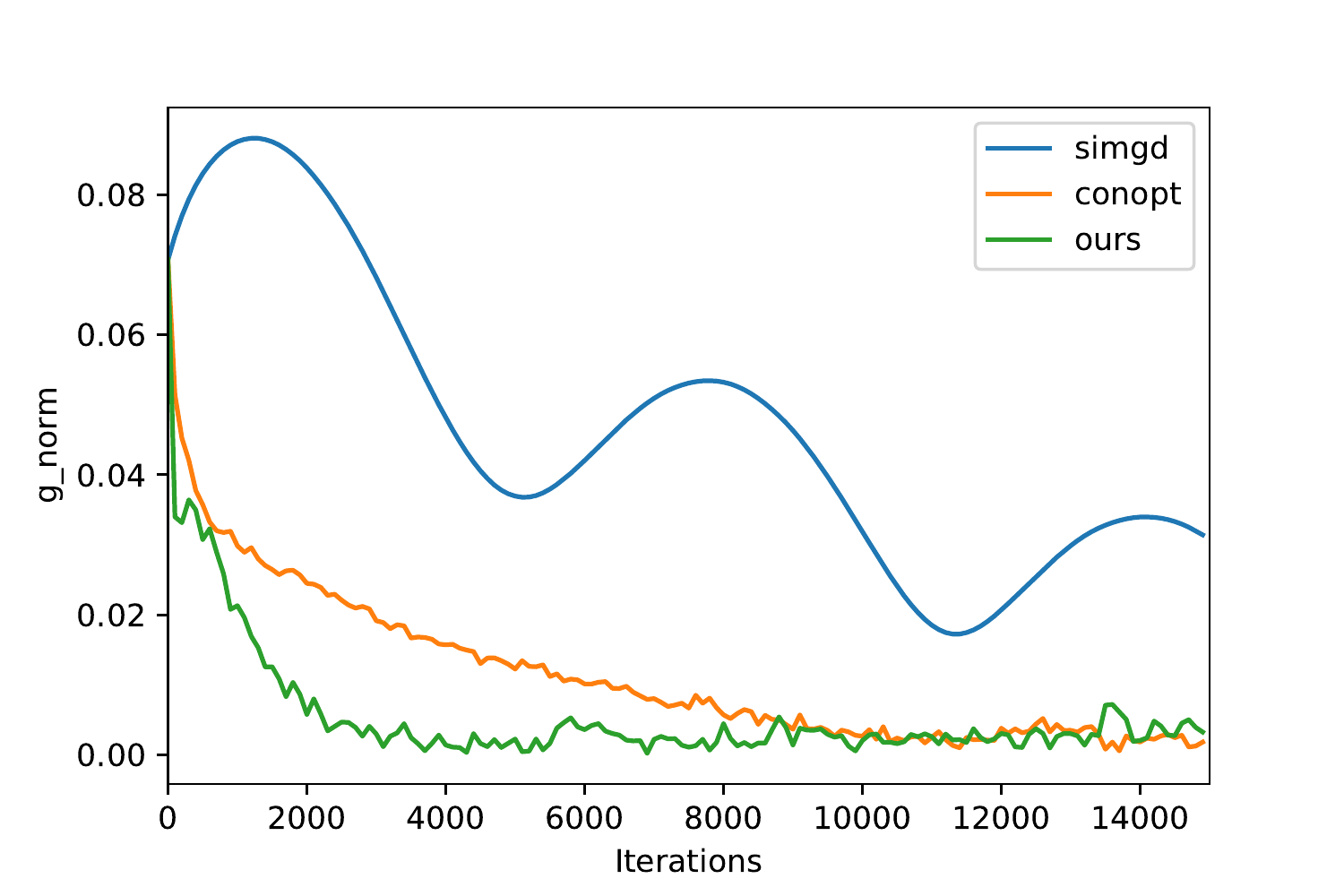}
		\caption{$\sigma=0.5$}
	\end{subfigure}
	\begin{subfigure}[b]{0.32\textwidth}
		\centering
		\includegraphics[width=\textwidth]{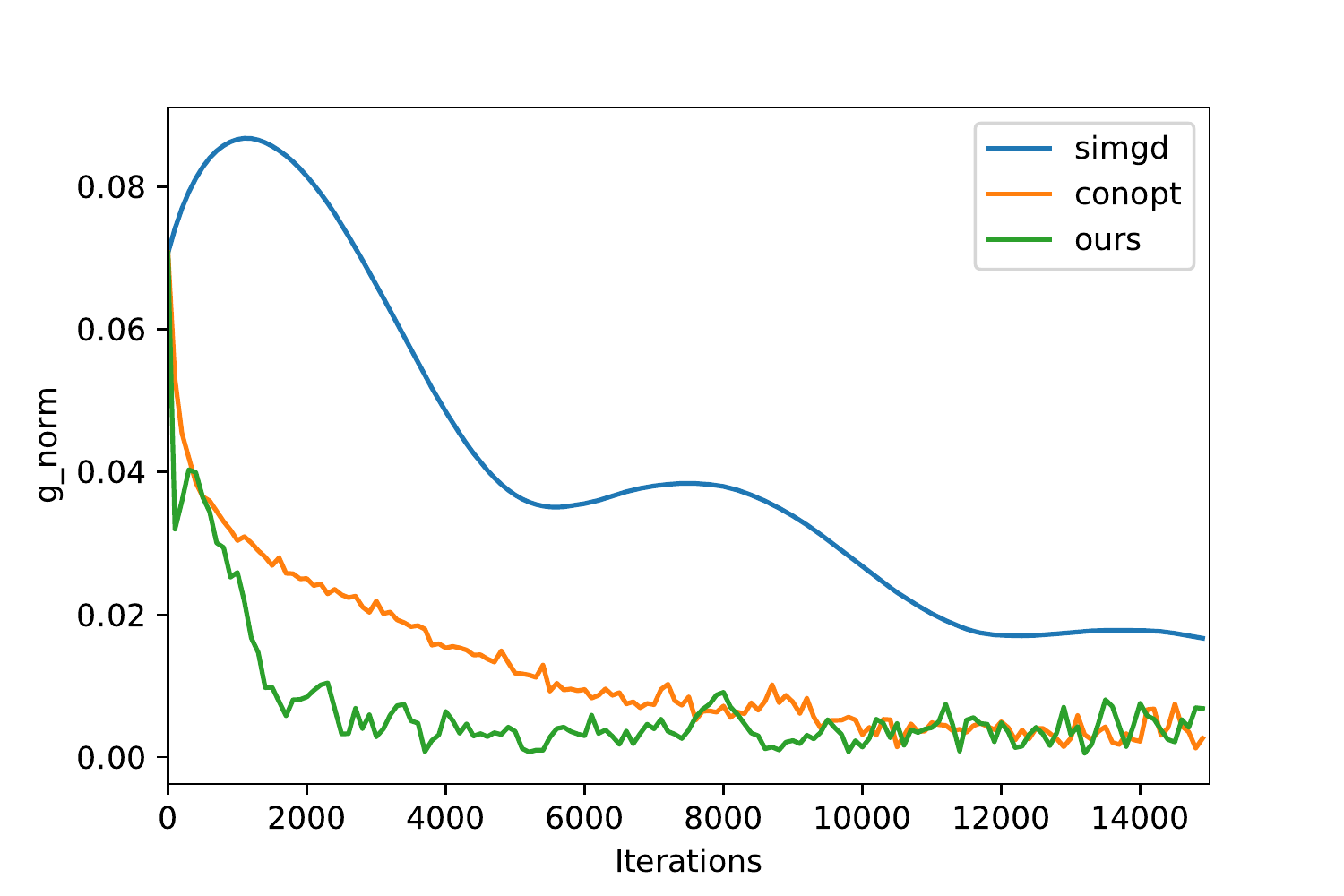}
		\caption{$\sigma=0.7$}
	\end{subfigure}
	\begin{subfigure}[b]{0.32\textwidth}
		\centering
		\includegraphics[width=\textwidth]{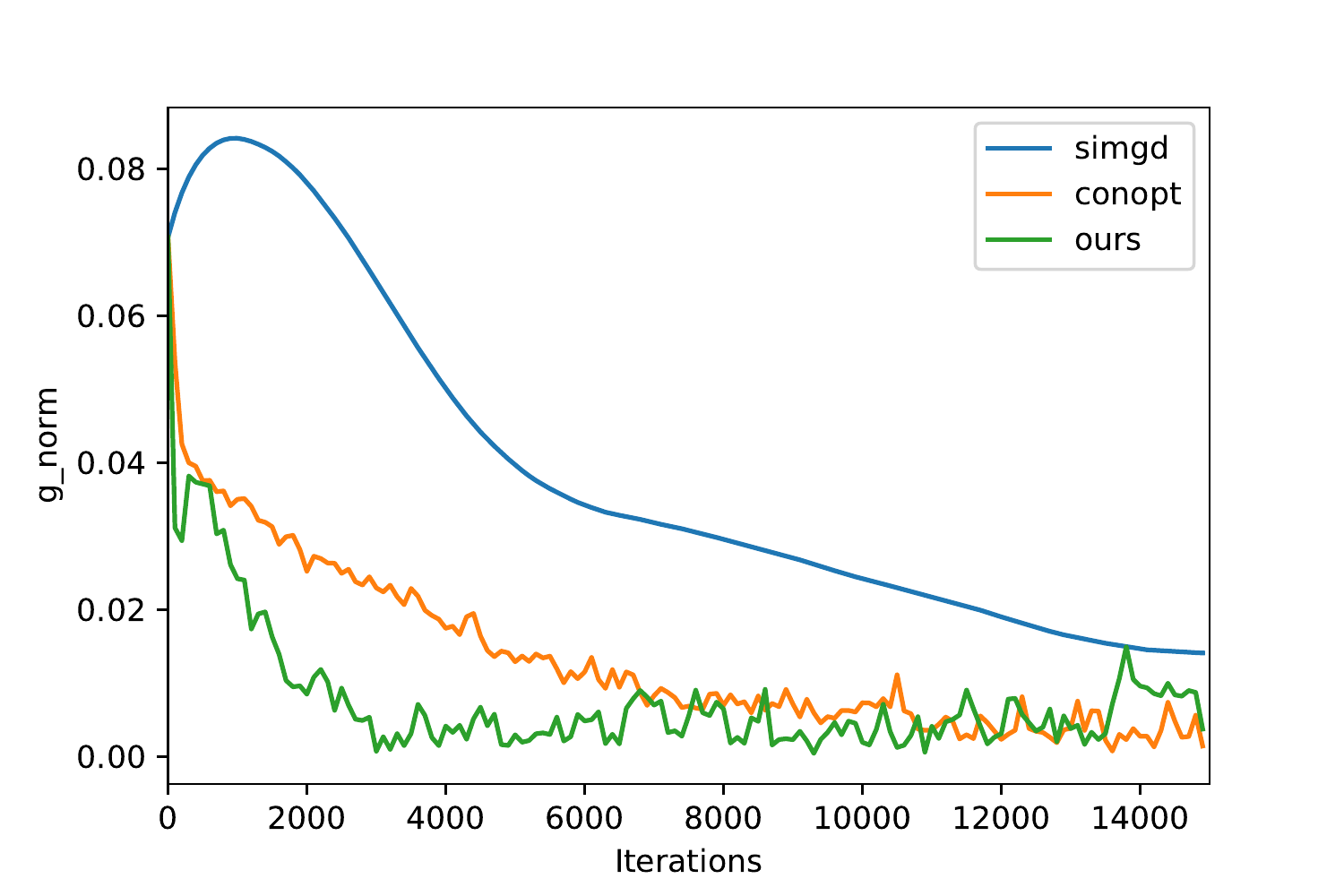}
		\caption{$\sigma=0.9$}
	\end{subfigure}
	
	\caption{Training dynamics of SimGD, ConOpt and JARE for the discriminator (top row) and the generator (bottom row) with varying standard deviation $\sigma$ where $\mu = 2$. 
	Note that the damping effect in SimGD becomes stronger as the standard derivation $\sigma$ increases.} \label{isotLin_sigma}
\end{figure}

\vspace{-7pt}
\subsection{More results on Mixture of Gaussians}
\label{more_mix_gau}
\vspace{-20pt}

\newpage


\begin{figure} [h]
	\centering
	\begin{subfigure}[b]{0.08\textwidth}
		\centering
		\includegraphics[width=\textwidth]{GMM/simgd_JS_bs128_zdim64_reg10_lr1e-4_rmsprop_model1_mu4_std6e-2/0.png}
	\end{subfigure}
	\begin{subfigure}[b]{0.08\textwidth}
		\centering
		\includegraphics[width=\textwidth]{GMM/simgd_JS_bs128_zdim64_reg10_lr1e-4_rmsprop_model1_mu4_std6e-2/2000.png}
	\end{subfigure}
	\begin{subfigure}[b]{0.08\textwidth}
		\centering
		\includegraphics[width=\textwidth]{GMM/simgd_JS_bs128_zdim64_reg10_lr1e-4_rmsprop_model1_mu4_std6e-2/4000.png}
	\end{subfigure}
	\begin{subfigure}[b]{0.08\textwidth}
		\centering
		\includegraphics[width=\textwidth]{GMM/simgd_JS_bs128_zdim64_reg10_lr1e-4_rmsprop_model1_mu4_std6e-2/6000.png}
	\end{subfigure}
	\begin{subfigure}[b]{0.08\textwidth}
		\centering
		\includegraphics[width=\textwidth]{GMM/simgd_JS_bs128_zdim64_reg10_lr1e-4_rmsprop_model1_mu4_std6e-2/8000.png}
	\end{subfigure}
	\begin{subfigure}[b]{0.08\textwidth}
		\centering
		\includegraphics[width=\textwidth]{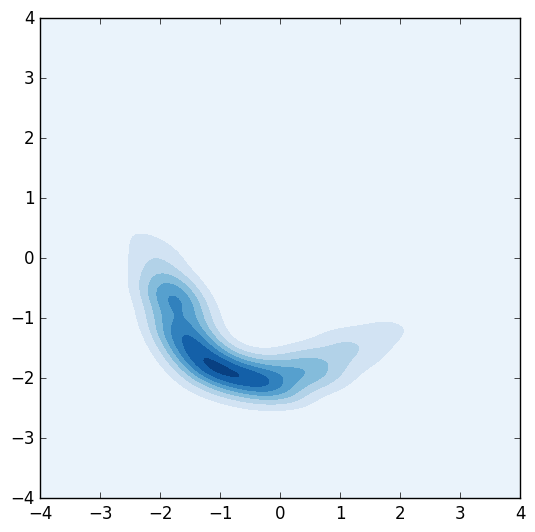}
	\end{subfigure}
	\caption*{\footnotesize (a) SimGD}
	
	\begin{subfigure}[b]{0.08\textwidth}
		\centering
		\includegraphics[width=\textwidth]{GMM/conopt_JS_bs128_zdim64_reg10_lr1e-4_rmsprop_model1_mu4_std6e-2/0.png}
	\end{subfigure}
	\begin{subfigure}[b]{0.08\textwidth}
		\centering
		\includegraphics[width=\textwidth]{GMM/conopt_JS_bs128_zdim64_reg10_lr1e-4_rmsprop_model1_mu4_std6e-2/2000.png}
	\end{subfigure}
	\begin{subfigure}[b]{0.08\textwidth}
		\centering
		\includegraphics[width=\textwidth]{GMM/conopt_JS_bs128_zdim64_reg10_lr1e-4_rmsprop_model1_mu4_std6e-2/4000.png}
	\end{subfigure}
	\begin{subfigure}[b]{0.08\textwidth}
		\centering
		\includegraphics[width=\textwidth]{GMM/conopt_JS_bs128_zdim64_reg10_lr1e-4_rmsprop_model1_mu4_std6e-2/6000.png}
	\end{subfigure}
	\begin{subfigure}[b]{0.08\textwidth}
		\centering
		\includegraphics[width=\textwidth]{GMM/conopt_JS_bs128_zdim64_reg10_lr1e-4_rmsprop_model1_mu4_std6e-2/8000.png}
	\end{subfigure}
	\begin{subfigure}[b]{0.08\textwidth}
		\centering
		\includegraphics[width=\textwidth]{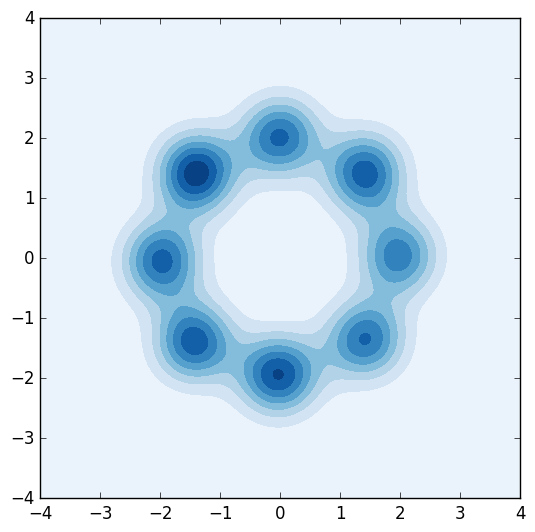}
	\end{subfigure}
	\caption*{\footnotesize (b) ConOpt ($\gamma = 10$)}
	
	\begin{subfigure}[b]{0.08\textwidth}
		\centering
		\includegraphics[width=\textwidth]{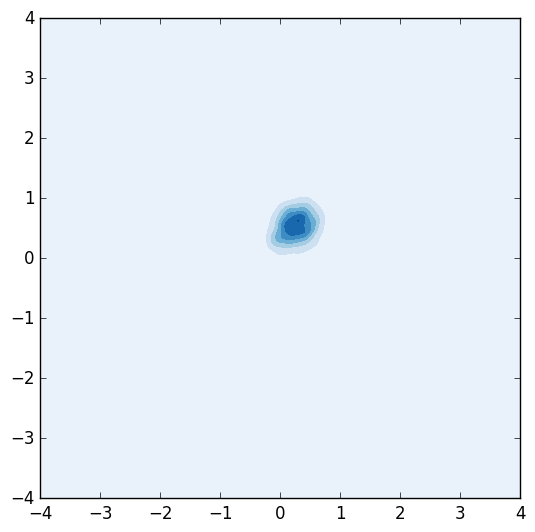}
	\end{subfigure}
	\begin{subfigure}[b]{0.08\textwidth}
		\centering
		\includegraphics[width=\textwidth]{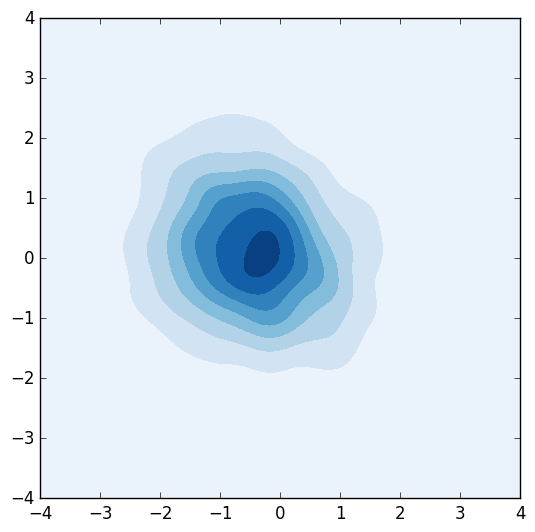}
	\end{subfigure}
	\begin{subfigure}[b]{0.08\textwidth}
		\centering
		\includegraphics[width=\textwidth]{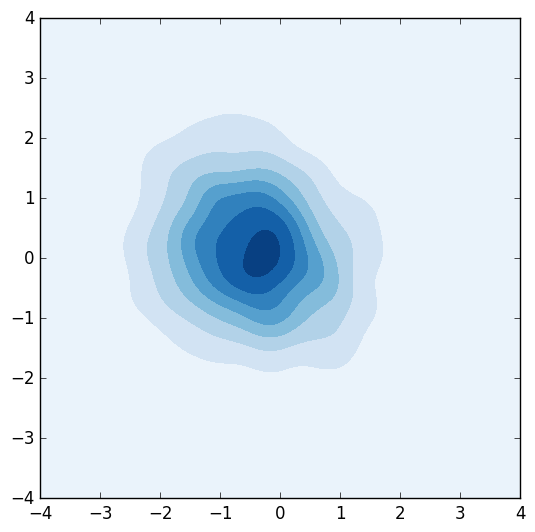}
	\end{subfigure}
	\begin{subfigure}[b]{0.08\textwidth}
		\centering
		\includegraphics[width=\textwidth]{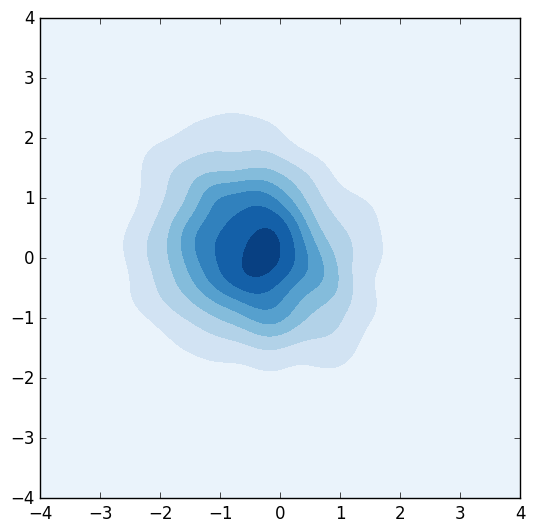}
	\end{subfigure}
	\begin{subfigure}[b]{0.08\textwidth}
		\centering
		\includegraphics[width=\textwidth]{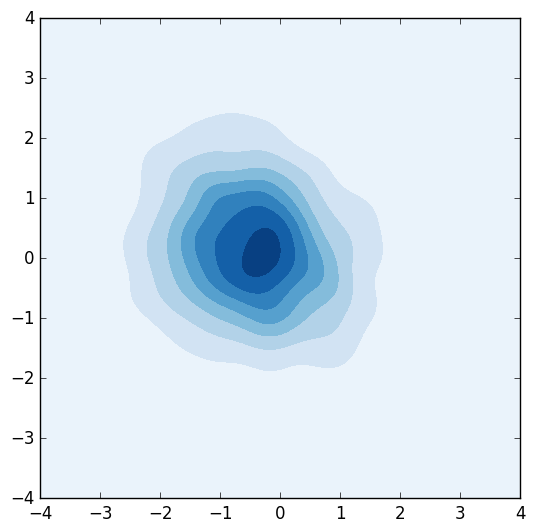}
	\end{subfigure}
	\begin{subfigure}[b]{0.08\textwidth}
		\centering
		\includegraphics[width=\textwidth]{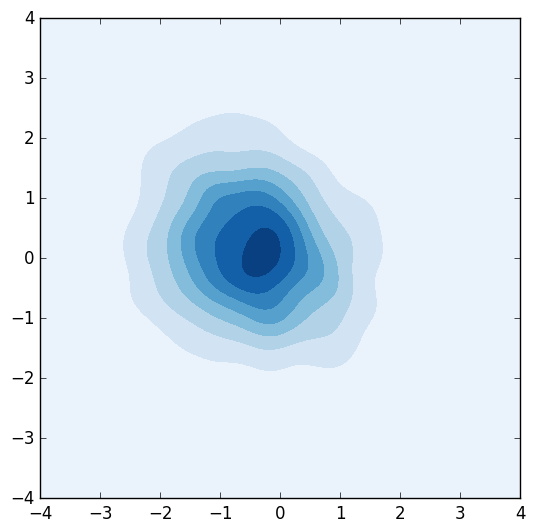}
	\end{subfigure}
	\caption*{\footnotesize (c) ConOpt ($\gamma = 1000$)}
	
	\begin{subfigure}[b]{0.08\textwidth}
		\centering
		\includegraphics[width=\textwidth]{GMM/ours_JS_bs128_zdim64_reg10_lr1e-4_rmsprop_model1_mu4_std6e-2/0.png}
	\end{subfigure}
	\begin{subfigure}[b]{0.08\textwidth}
		\centering
		\includegraphics[width=\textwidth]{GMM/ours_JS_bs128_zdim64_reg10_lr1e-4_rmsprop_model1_mu4_std6e-2/2000.png}
	\end{subfigure}
	\begin{subfigure}[b]{0.08\textwidth}
		\centering
		\includegraphics[width=\textwidth]{GMM/ours_JS_bs128_zdim64_reg10_lr1e-4_rmsprop_model1_mu4_std6e-2/4000.png}
	\end{subfigure}
	\begin{subfigure}[b]{0.08\textwidth}
		\centering
		\includegraphics[width=\textwidth]{GMM/ours_JS_bs128_zdim64_reg10_lr1e-4_rmsprop_model1_mu4_std6e-2/6000.png}
	\end{subfigure}
	\begin{subfigure}[b]{0.08\textwidth}
		\centering
		\includegraphics[width=\textwidth]{GMM/ours_JS_bs128_zdim64_reg10_lr1e-4_rmsprop_model1_mu4_std6e-2/8000.png}
	\end{subfigure}
	\begin{subfigure}[b]{0.08\textwidth}
		\centering
		\includegraphics[width=\textwidth]{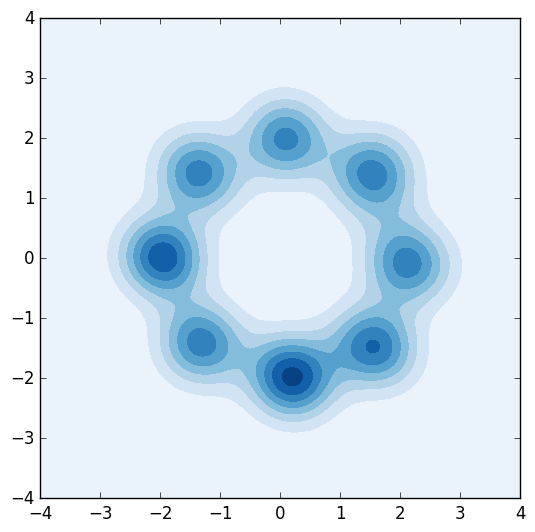}
	\end{subfigure}
	\caption*{\footnotesize (d) Ours ($\gamma = 10$)}
	
	\begin{subfigure}[b]{0.08\textwidth}
		\centering
		\includegraphics[width=\textwidth]{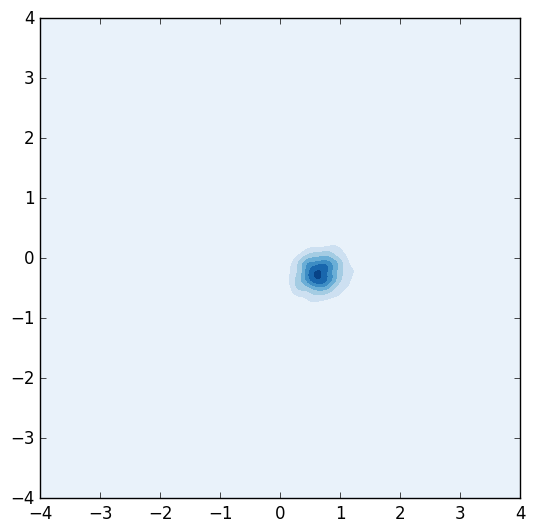}
	\end{subfigure}
	\begin{subfigure}[b]{0.08\textwidth}
		\centering
		\includegraphics[width=\textwidth]{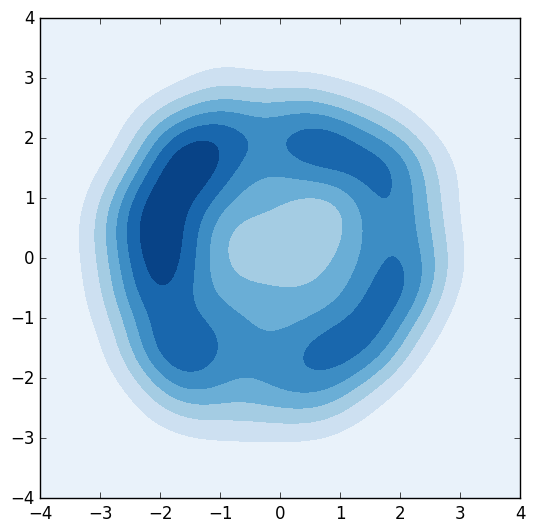}
	\end{subfigure}
	\begin{subfigure}[b]{0.08\textwidth}
		\centering
		\includegraphics[width=\textwidth]{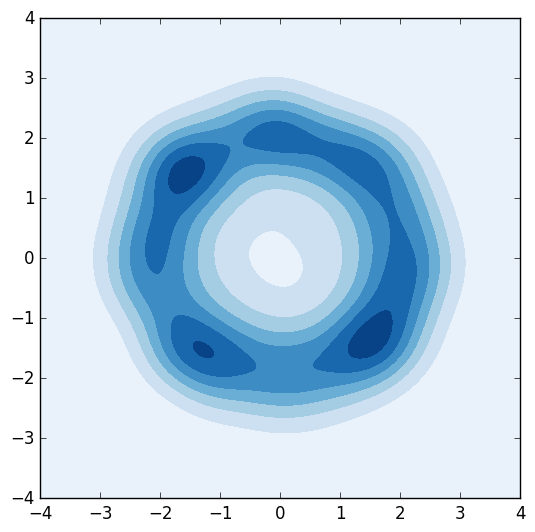}
	\end{subfigure}
	\begin{subfigure}[b]{0.08\textwidth}
		\centering
		\includegraphics[width=\textwidth]{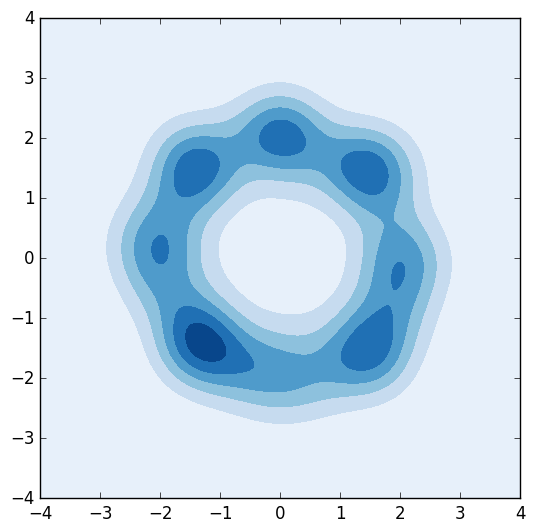}
	\end{subfigure}
	\begin{subfigure}[b]{0.08\textwidth}
		\centering
		\includegraphics[width=\textwidth]{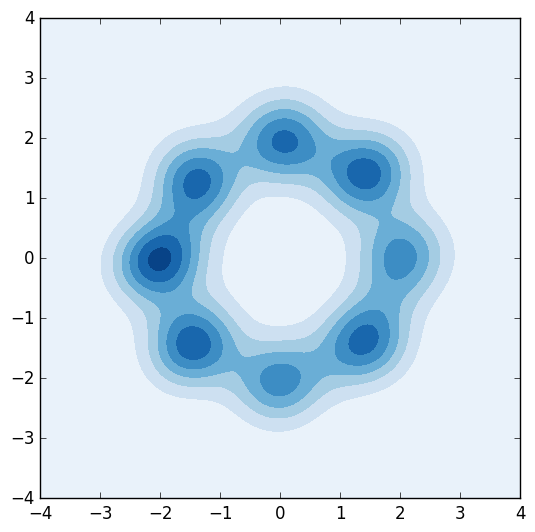}
	\end{subfigure}
	\begin{subfigure}[b]{0.08\textwidth}
		\centering
		\includegraphics[width=\textwidth]{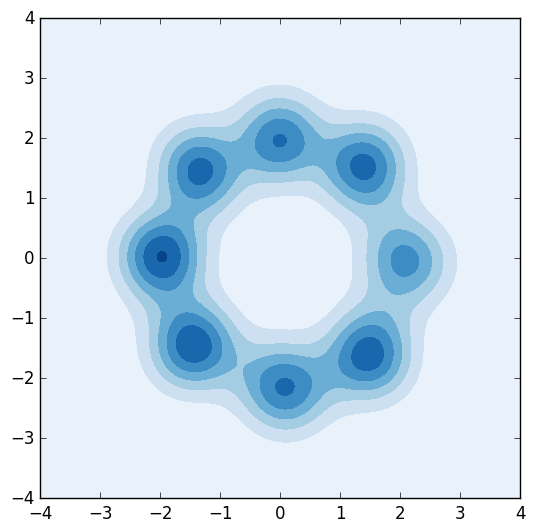}
	\end{subfigure}
	\caption*{\footnotesize (e) Ours ($\gamma = 1000$)}
	\caption{\small Comparison of SimGD (a), ConOpt (b,c) and Ours (d,e) on the mixture of Gaussians over iterations where $r=2$. From left to right, each row consists of the results after 0, 2000, 4000, 6000, 8000 and 10000 iterations. } \label{gen_gmm_mu4_full}
\end{figure}

\begin{figure} [t]
	\centering
	\begin{subfigure}[b]{0.08\textwidth}
		\centering
		\includegraphics[width=\textwidth]{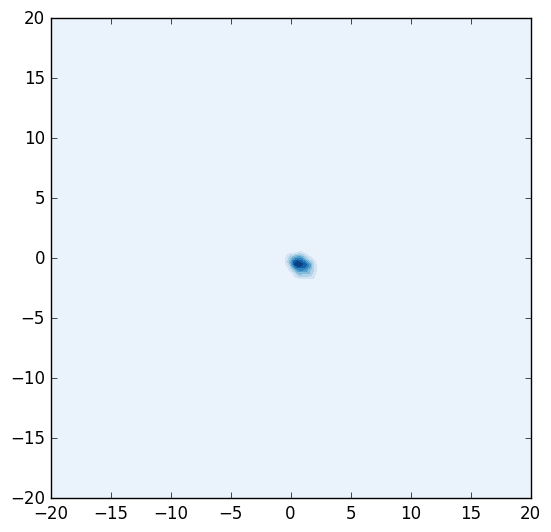}
	\end{subfigure}
	\begin{subfigure}[b]{0.08\textwidth}
		\centering
		\includegraphics[width=\textwidth]{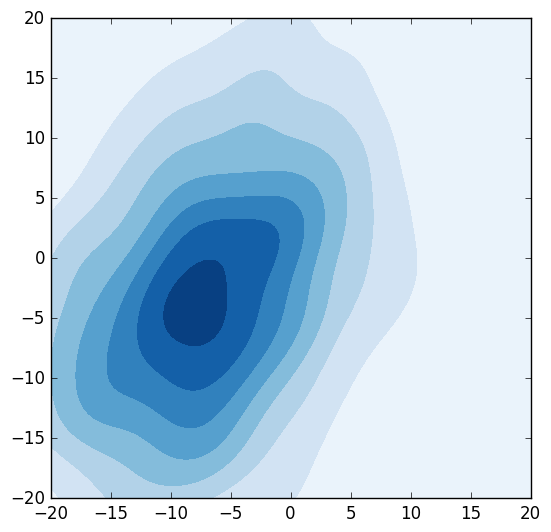}
	\end{subfigure}
	\begin{subfigure}[b]{0.08\textwidth}
		\centering
		\includegraphics[width=\textwidth]{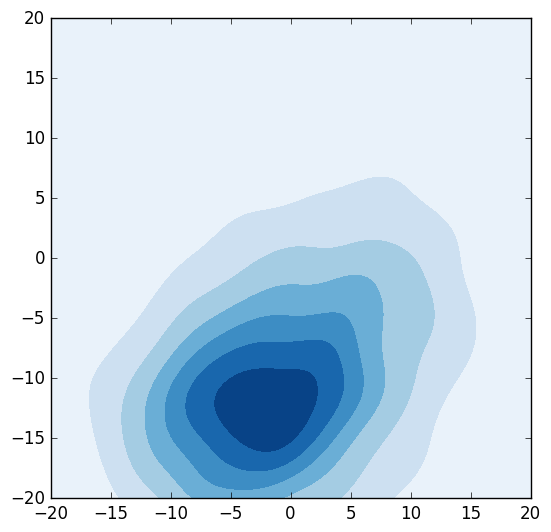}
	\end{subfigure}
	\begin{subfigure}[b]{0.08\textwidth}
		\centering
		\includegraphics[width=\textwidth]{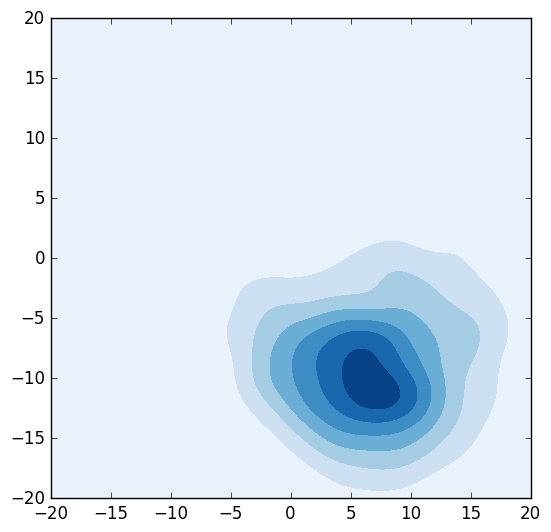}
	\end{subfigure}
	\begin{subfigure}[b]{0.08\textwidth}
		\centering
		\includegraphics[width=\textwidth]{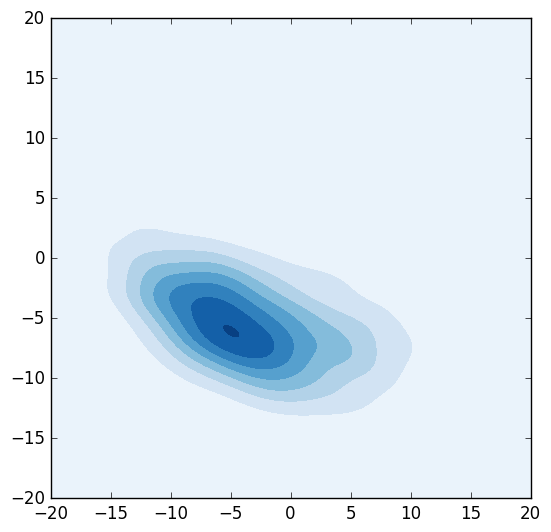}
	\end{subfigure}
	\begin{subfigure}[b]{0.08\textwidth}
		\centering
		\includegraphics[width=\textwidth]{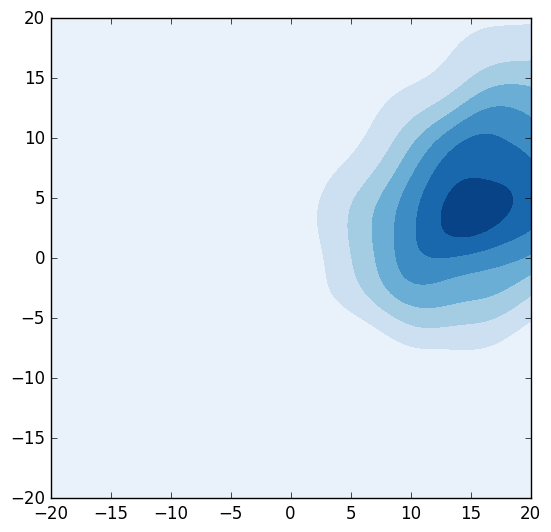}
	\end{subfigure}
	\caption*{\footnotesize (a) SimGD}
	
	\begin{subfigure}[b]{0.08\textwidth}
		\centering
		\includegraphics[width=\textwidth]{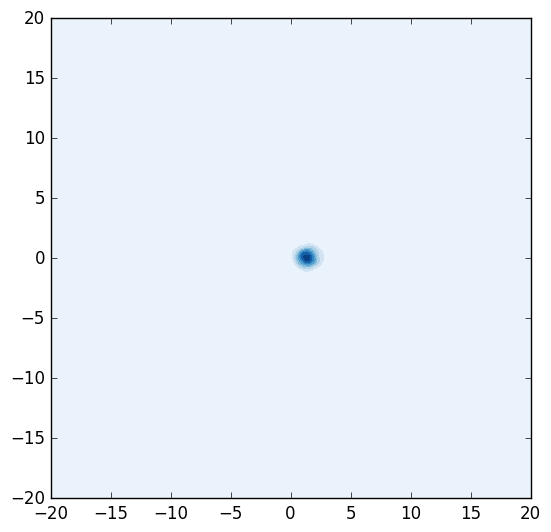}
	\end{subfigure}
	\begin{subfigure}[b]{0.08\textwidth}
		\centering
		\includegraphics[width=\textwidth]{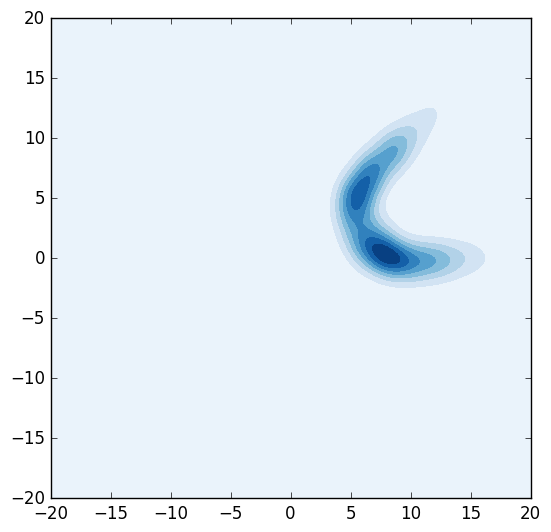}
	\end{subfigure}
	\begin{subfigure}[b]{0.08\textwidth}
		\centering
		\includegraphics[width=\textwidth]{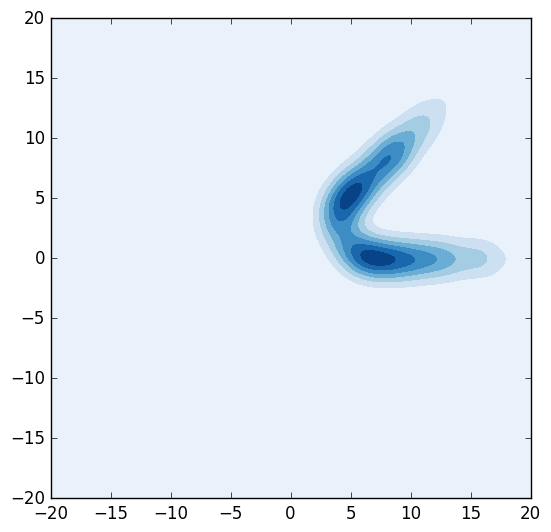}
	\end{subfigure}
	\begin{subfigure}[b]{0.08\textwidth}
		\centering
		\includegraphics[width=\textwidth]{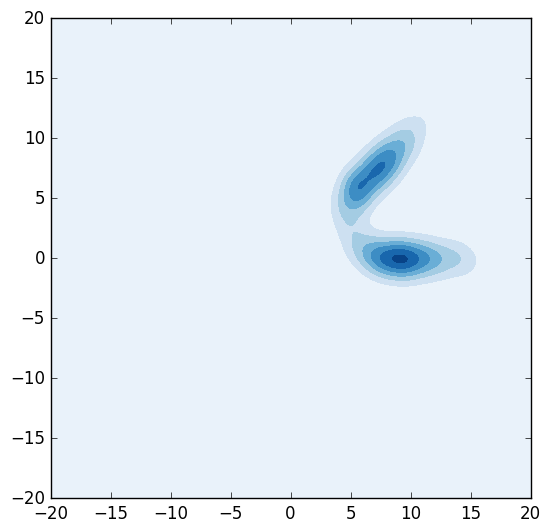}
	\end{subfigure}
	\begin{subfigure}[b]{0.08\textwidth}
		\centering
		\includegraphics[width=\textwidth]{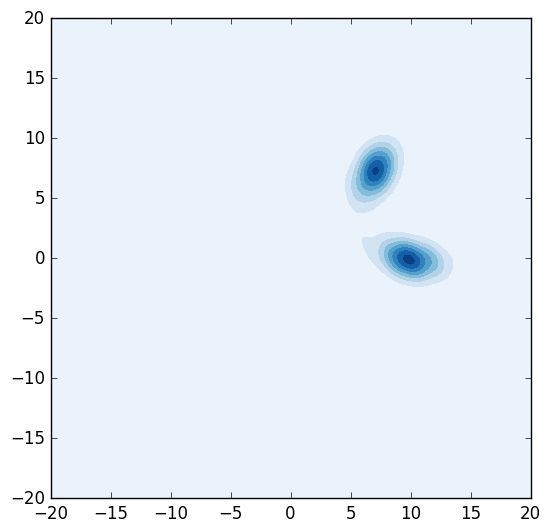}
	\end{subfigure}
	\begin{subfigure}[b]{0.08\textwidth}
		\centering
		\includegraphics[width=\textwidth]{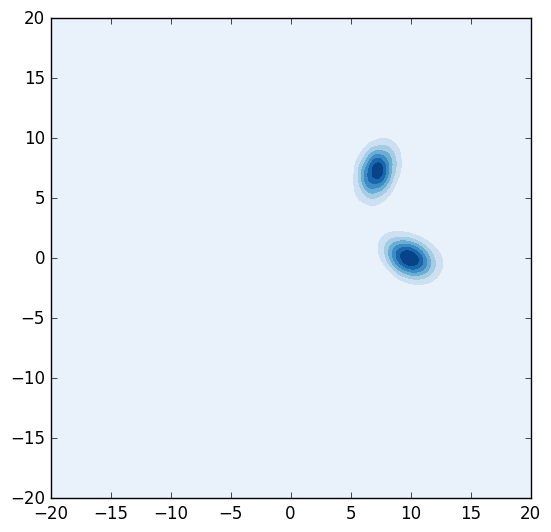}
	\end{subfigure}
	\caption*{\footnotesize (b) ConOpt ($\gamma = 10$)}
	
	\begin{subfigure}[b]{0.08\textwidth}
		\centering
		\includegraphics[width=\textwidth]{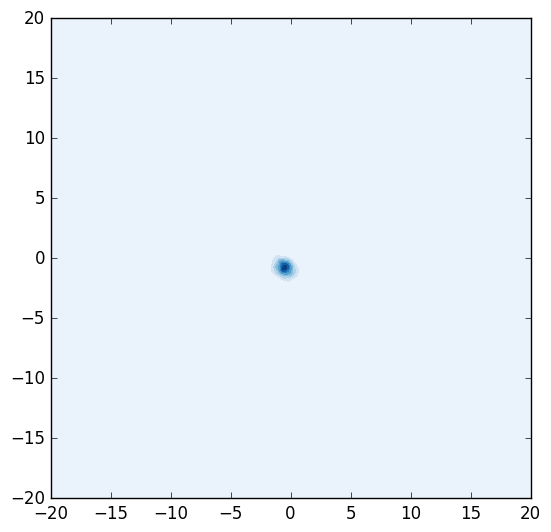}
	\end{subfigure}
	\begin{subfigure}[b]{0.08\textwidth}
		\centering
		\includegraphics[width=\textwidth]{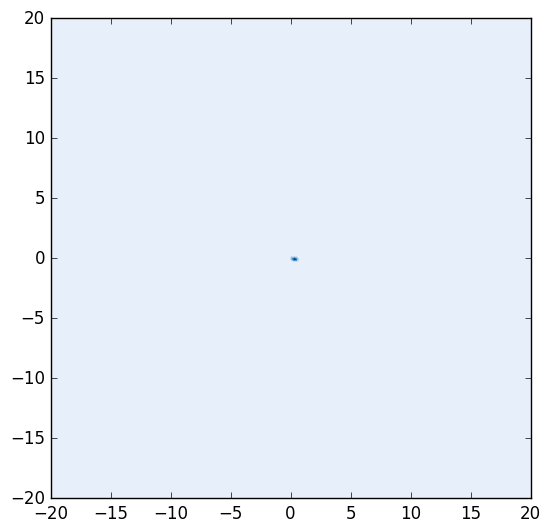}
	\end{subfigure}
	\begin{subfigure}[b]{0.08\textwidth}
		\centering
		\includegraphics[width=\textwidth]{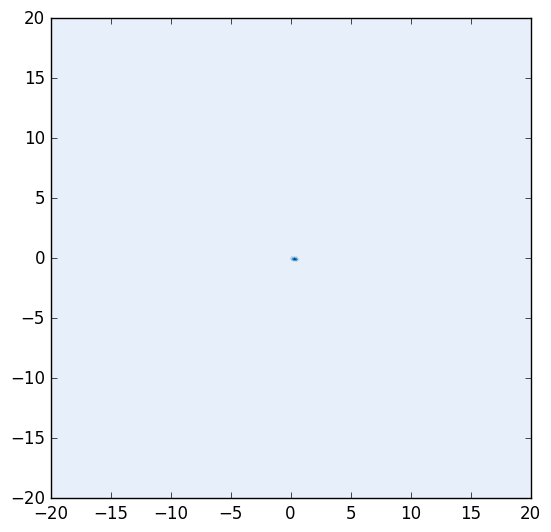}
	\end{subfigure}
	\begin{subfigure}[b]{0.08\textwidth}
		\centering
		\includegraphics[width=\textwidth]{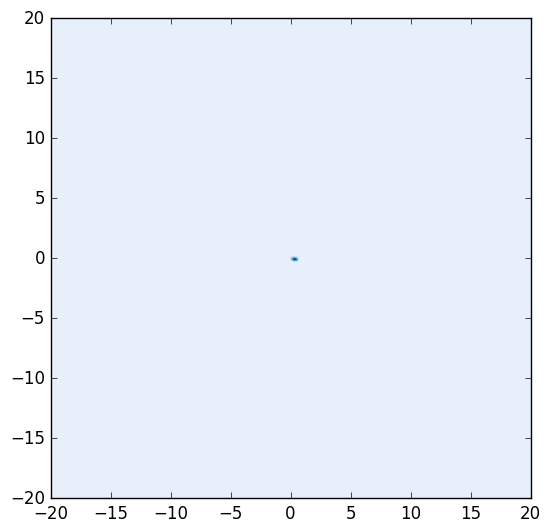}
	\end{subfigure}
	\begin{subfigure}[b]{0.08\textwidth}
		\centering
		\includegraphics[width=\textwidth]{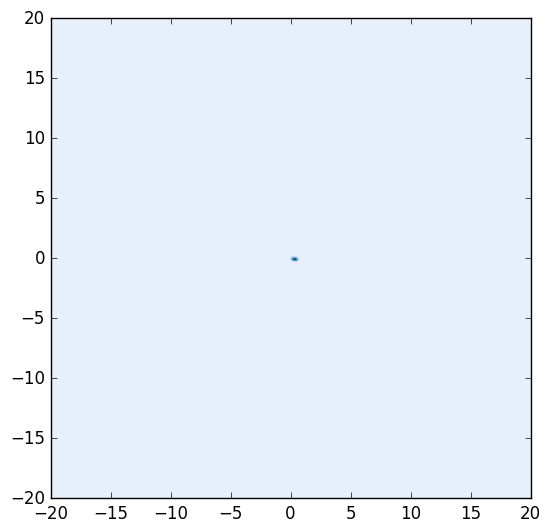}
	\end{subfigure}
	\begin{subfigure}[b]{0.08\textwidth}
		\centering
		\includegraphics[width=\textwidth]{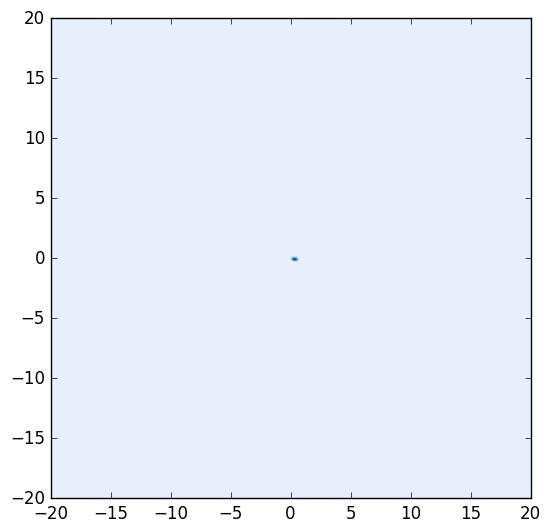}
	\end{subfigure}
	\caption*{\footnotesize (c) ConOpt ($\gamma = 1000$)}
	
	\begin{subfigure}[b]{0.08\textwidth}
		\centering
		\includegraphics[width=\textwidth]{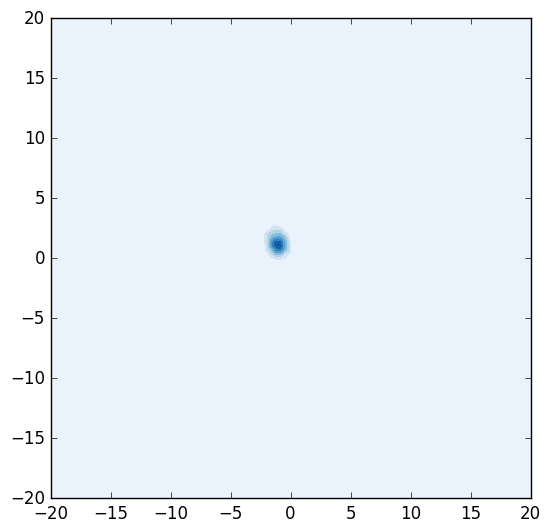}
	\end{subfigure}
	\begin{subfigure}[b]{0.08\textwidth}
		\centering
		\includegraphics[width=\textwidth]{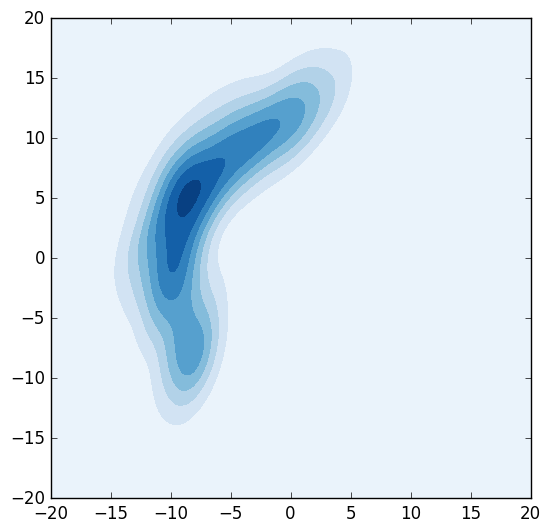}
	\end{subfigure}
	\begin{subfigure}[b]{0.08\textwidth}
		\centering
		\includegraphics[width=\textwidth]{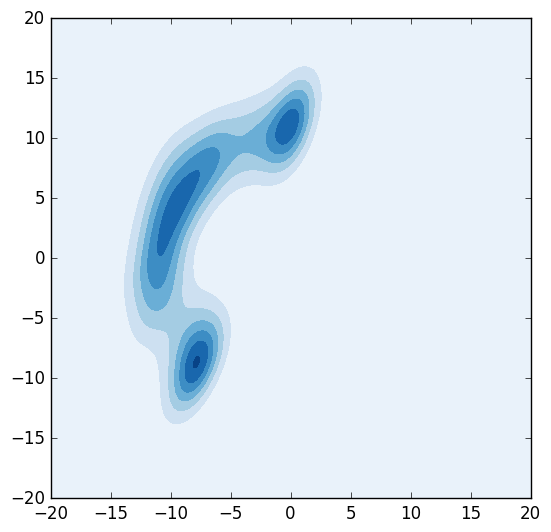}
	\end{subfigure}
	\begin{subfigure}[b]{0.08\textwidth}
		\centering
		\includegraphics[width=\textwidth]{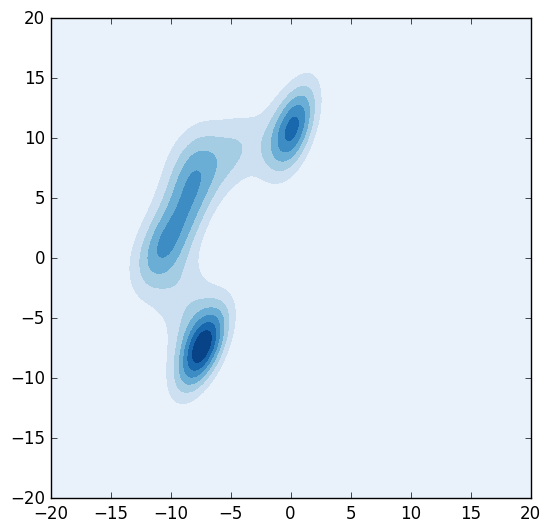}
	\end{subfigure}
	\begin{subfigure}[b]{0.08\textwidth}
		\centering
		\includegraphics[width=\textwidth]{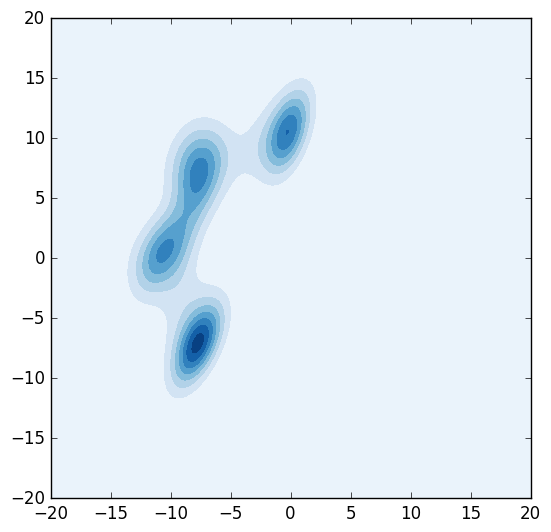}
	\end{subfigure}
	\begin{subfigure}[b]{0.08\textwidth}
		\centering
		\includegraphics[width=\textwidth]{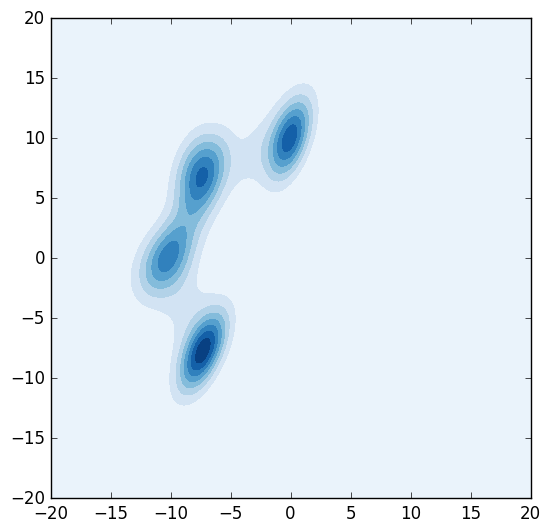}
	\end{subfigure}
	\caption*{\footnotesize (d) Ours ($\gamma = 10$)}
	
	\begin{subfigure}[b]{0.08\textwidth}
		\centering
		\includegraphics[width=\textwidth]{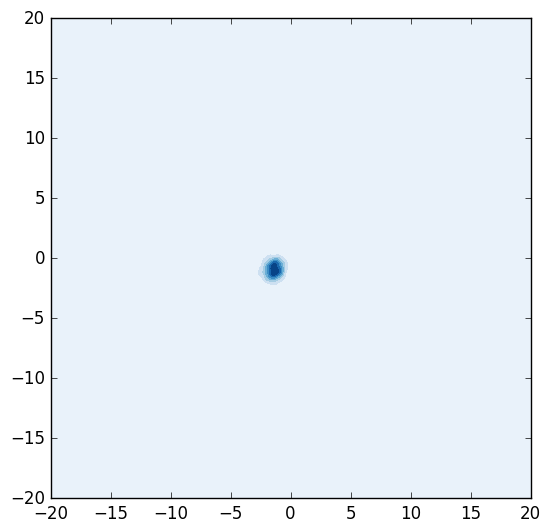}
	\end{subfigure}
	\begin{subfigure}[b]{0.08\textwidth}
		\centering
		\includegraphics[width=\textwidth]{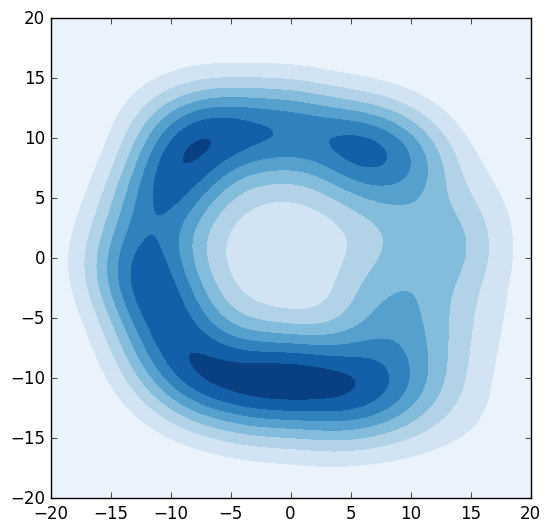}
	\end{subfigure}
	\begin{subfigure}[b]{0.08\textwidth}
		\centering
		\includegraphics[width=\textwidth]{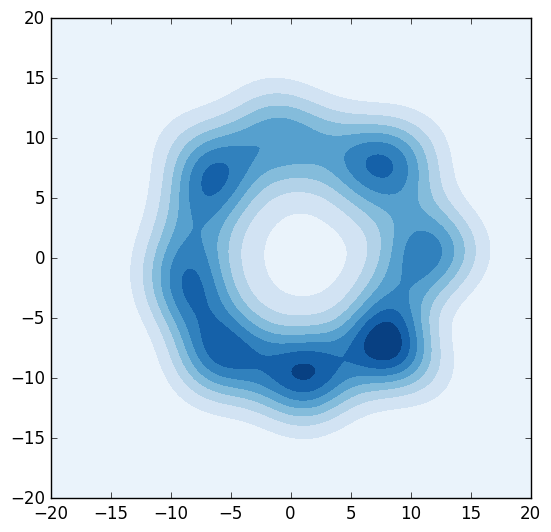}
	\end{subfigure}
	\begin{subfigure}[b]{0.08\textwidth}
		\centering
		\includegraphics[width=\textwidth]{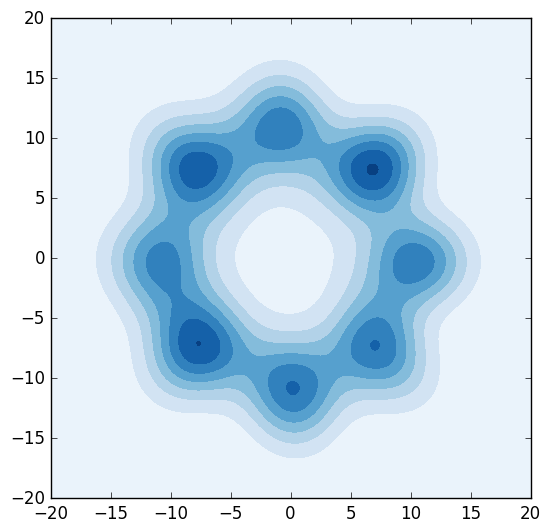}
	\end{subfigure}
	\begin{subfigure}[b]{0.08\textwidth}
		\centering
		\includegraphics[width=\textwidth]{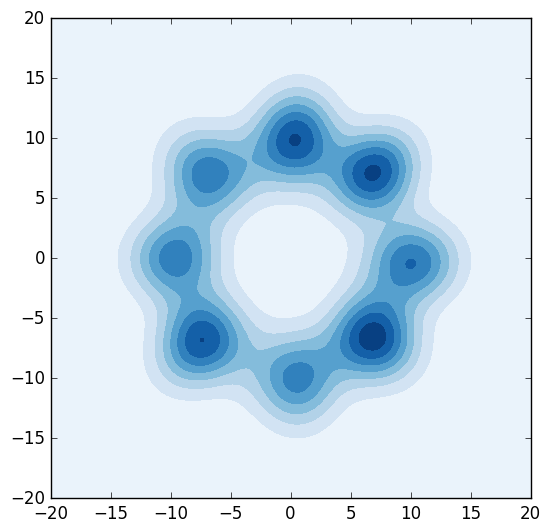}
	\end{subfigure}
	\begin{subfigure}[b]{0.08\textwidth}
		\centering
		\includegraphics[width=\textwidth]{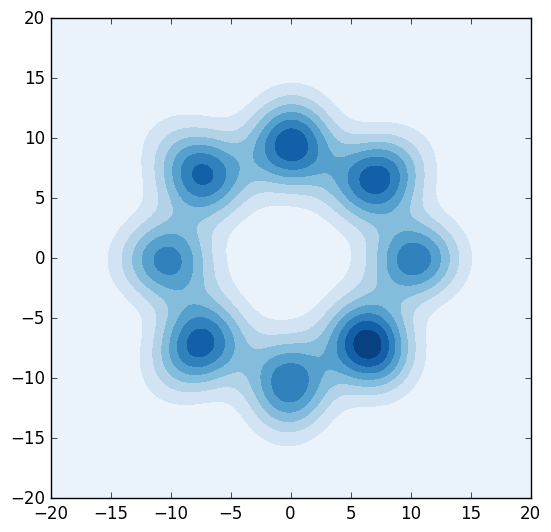}
	\end{subfigure}
	\caption*{\footnotesize (e) Ours ($\gamma = 1000$)}
	\caption{\small Comparison of SimGD (a), ConOpt (b,c) and Ours (d,e) on the mixture of Gaussians over iterations where $r=10$. From left to right, each row consists of the results after 0, 2000, 4000, 6000, 8000 and 10000 iterations. 
	} \label{gen_gmm_mu20_full}
\end{figure}

\newpage

\onecolumn

\subsection{Network architectures}
\label{sec_net_arc}

\begin{table}[H]
    \centering
    \begin{subtable}{.48\linewidth}
        \centering
        \begin{tabular}{c}
        \hline\hline
         $z \in \mathbb{R}^{128} \sim \mathcal{N}(0, I)$ \\
         \hline
         dense, $2 \times 2 \times M_f$\\
         \hline
         $4 \times 4$, stride=2, deconv. $M_f$ ReLU\\
         ResBlock $M_f$\\
         \hline
         $4 \times 4$, stride=2, deconv. $M_f$ ReLU\\
         ResBlock $M_f$\\
         \hline
         $4 \times 4$, stride=2, deconv. $M_f$ ReLU\\
         ResBlock $M_f$\\
         \hline
         $4 \times 4$, stride=2, deconv. 3 tanh \\
         \hline
         \hline
    \end{tabular}
    \caption{Generator}
    \end{subtable}
    \begin{subtable}{.48\linewidth}
        \centering
        \begin{tabular}{c}
        \hline\hline
         $x \in \mathbb{R}^{32 \times 32 \times 3}$ \\
         \hline
         $4 \times 4$, stride=2, conv. $M_f$ ReLU\\
         ResBlock $M_f$\\
         \hline
         $4 \times 4$, stride=2, conv. $M_f$ ReLU\\
         ResBlock $M_f$\\
         \hline
         $4 \times 4$, stride=2, conv. $M_f$ ReLU\\
         ResBlock $M_f$\\
         \hline
         $4 \times 4$, stride=2, conv. $M_f$ ReLU\\
         ResBlock $M_f$\\
         \hline
         dense $\to 1$ \\
         \hline
         \hline
    \end{tabular}
    \caption{Discriminator}
    \end{subtable}
    \caption{ResNet architectures v1 for CIFAR-10 where $M_f$ denotes the number of filters.}
    \vspace{2mm}
    \label{resnet_cifar10_v1}
\end{table}

\begin{table}[H]
    \centering
    \begin{subtable}{.48\linewidth}
        \centering
        \begin{tabular}{c}
        \hline\hline
         $z \in \mathbb{R}^{128} \sim \mathcal{N}(0, I)$ \\
         \hline
         dense, $4 \times 4 \times M_f$\\
         \hline
         $4 \times 4$, stride=2, deconv. $M_f$ ReLU\\
         ResBlock $M_f$\\
         \hline
         $4 \times 4$, stride=2, deconv. $M_f$ ReLU\\
         ResBlock $M_f$\\
         \hline
         $4 \times 4$, stride=2, deconv. $M_f$ ReLU\\
         ResBlock $M_f$\\
         \hline
         $3 \times 3$, stride=1, conv. 3 tanh \\
         \hline
         \hline
    \end{tabular}
    \caption{Generator}
    \end{subtable}
    \begin{subtable}{.48\linewidth}
        \centering
        \begin{tabular}{c}
        \hline\hline
         $x \in \mathbb{R}^{32 \times 32 \times 3}$ \\
         \hline
         $4 \times 4$, stride=2, conv. $M_f$ ReLU\\
         ResBlock $M_f$\\
         \hline
         $4 \times 4$, stride=2, conv. $M_f$ ReLU\\
         ResBlock $M_f$\\
         \hline
         $4 \times 4$, stride=2, conv. $M_f$ ReLU\\
         ResBlock $M_f$\\
         \hline
         $4 \times 4$, stride=2, conv. $M_f$ ReLU\\
         ResBlock $M_f$\\
         \hline
         dense $\to 1$ \\
         \hline
         \hline
    \end{tabular}
    \caption{Discriminator}
    \end{subtable}
    \caption{ResNet architectures v2 for CIFAR-10 where $M_f$ denotes the number of filters.}
    \vspace{2mm}
    \label{resnet_cifar10_v2}
\end{table}

\begin{table}[H]
    \centering
    \begin{subtable}{.48\linewidth}
        \centering
        \begin{tabular}{c}
        \hline\hline
         $z \in \mathbb{R}^{128} \sim \mathcal{N}(0, I)$ \\
         \hline
         dense, $4 \times 4 \times M_f$\\
         \hline
         $4 \times 4$, stride=2, deconv. $M_f$ ReLU\\
         ResBlock $M_f$\\
         \hline
         $4 \times 4$, stride=2, deconv. $M_f$ ReLU\\
         ResBlock $M_f$\\
         \hline
         $4 \times 4$, stride=2, deconv. $M_f$ ReLU\\
         ResBlock $M_f$\\
         \hline
         $4 \times 4$, stride=2, deconv. $M_f$ ReLU\\
         ResBlock $M_f$\\
         \hline
         $3 \times 3$, stride=1, conv. 3 tanh \\
         \hline
         \hline
    \end{tabular}
    \caption{Generator}
    \end{subtable}
    \begin{subtable}{.48\linewidth}
        \centering
        \begin{tabular}{c}
        \hline\hline
         $x \in \mathbb{R}^{64 \times 64 \times 3}$ \\
         \hline
         $4 \times 4$, stride=2, conv. $M_f$ ReLU\\
         ResBlock $M_f$\\
         \hline
         $4 \times 4$, stride=2, conv. $M_f$ ReLU\\
         ResBlock $M_f$\\
         \hline
         $4 \times 4$, stride=2, conv. $M_f$ ReLU\\
         ResBlock $M_f$\\
         \hline
         $4 \times 4$, stride=2, conv. $M_f$ ReLU\\
         ResBlock $M_f$\\
         \hline
         $4 \times 4$, stride=2, conv. $M_f$ ReLU\\
         ResBlock $M_f$\\
         \hline
         dense $\to 1$ \\
         \hline
         \hline
    \end{tabular}
    \caption{Discriminator}
    \end{subtable}
    \caption{ResNet architectures for CelebA where $M_f$ denotes the number of filters.}
    \vspace{2mm}
    \label{resnet_celeba}
\end{table}

\begin{table}[H]
    \centering
    \begin{subtable}{.48\linewidth}
        \centering
        \begin{tabular}{c}
        \hline\hline
         $z \in \mathbb{R}^{128} \sim \mathcal{N}(0, I)$ \\
         \hline
         dense, $4 \times 4 \times M_f$\\
         \hline
         $4 \times 4$, stride=2, deconv. $M_f$ ReLU\\
         ResBlock $M_f$\\
         \hline
         $4 \times 4$, stride=2, deconv. $M_f$ ReLU\\
         ResBlock $M_f$\\
         \hline
         $4 \times 4$, stride=2, deconv. $M_f$ ReLU\\
         ResBlock $M_f$\\
         \hline
         $4 \times 4$, stride=2, deconv. $M_f$ ReLU\\
         ResBlock $M_f$\\
         \hline
         $4 \times 4$, stride=2, deconv. $M_f$ ReLU\\
         ResBlock $M_f$\\
         \hline
         $3 \times 3$, stride=1, conv. 3 tanh \\
         \hline
         \hline
    \end{tabular}
    \caption{Generator}
    \end{subtable}
    \begin{subtable}{.48\linewidth}
        \centering
        \begin{tabular}{c}
        \hline\hline
         $x \in \mathbb{R}^{128 \times 128 \times 3}$ \\
         \hline
         $4 \times 4$, stride=2, conv. $M_f$ ReLU\\
         ResBlock $M_f$\\
         \hline
         $4 \times 4$, stride=2, conv. $M_f$ ReLU\\
         ResBlock $M_f$\\
         \hline
         $4 \times 4$, stride=2, conv. $M_f$ ReLU\\
         ResBlock $M_f$\\
         \hline
         $4 \times 4$, stride=2, conv. $M_f$ ReLU\\
         ResBlock $M_f$\\
         \hline
         $4 \times 4$, stride=2, conv. $M_f$ ReLU\\
         ResBlock $M_f$\\
         \hline
         $4 \times 4$, stride=2, conv. $M_f$ ReLU\\
         ResBlock $M_f$\\
         \hline
         dense $\to 1$ \\
         \hline
         \hline
    \end{tabular}
    \caption{Discriminator}
    \end{subtable}
    \caption{ ResNet architectures for ImageNet where $M_f$ denotes the number of filters.}
    \vspace{2mm}
    \label{resnet_imagenet}
\end{table}



\newpage

\onecolumn

\subsection{Generated images on CIFAR-10 with four methods: GAN, SN-GAN, ConOpt and JARE.} 
\label{appendix_cifar10}
\begin{figure} [H]
    \centering
    \includegraphics[width=\textwidth]{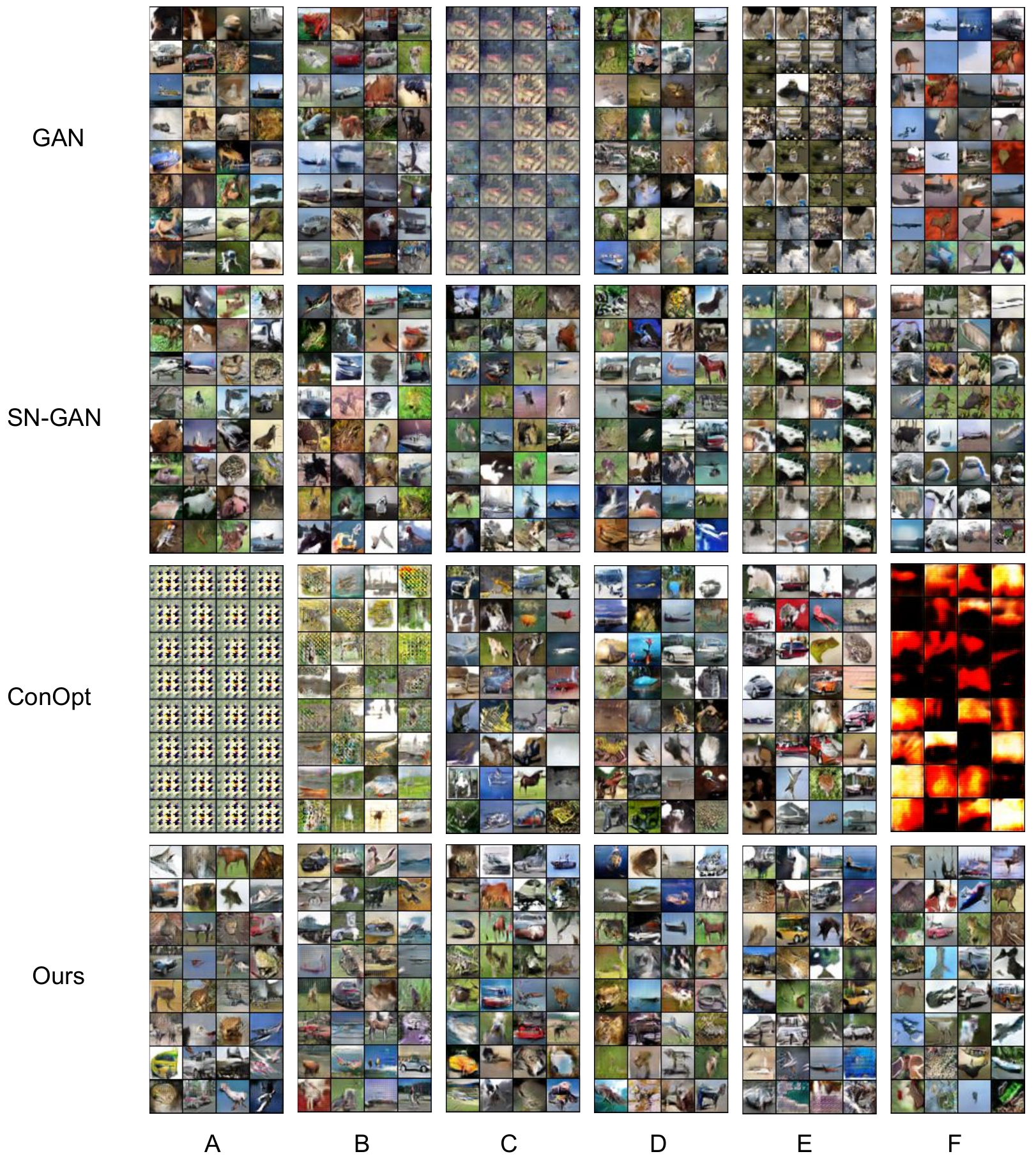}
    \caption{Generated images on CIFAR-10 with four training methods: standard GAN (or GAN), SN-GAN, ConOpt and JARE (Ours) in all the A-F settings. Best viewed in the electronic version by zooming in. We can see that only JARE is able to generate realistic images when training on CIFAR-10 across all six settings.}
    \label{cifar10_robustness}
\end{figure}



\begin{figure} [t]
    \centering
    \begin{subfigure}[b]{0.33\textwidth}
		\centering
		\includegraphics[width=\textwidth]{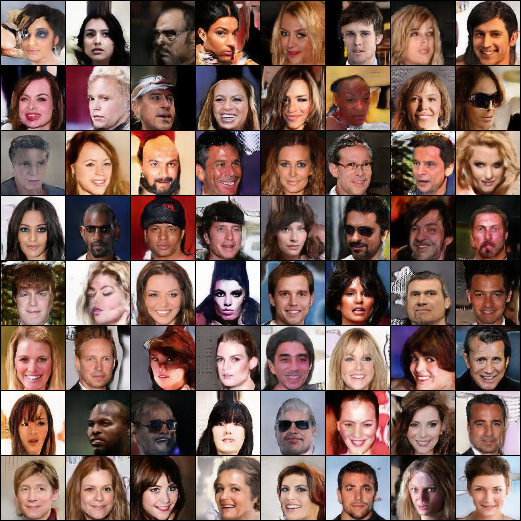}
		\caption{\footnotesize Random samples for CelebA}
	\end{subfigure}
	\quad
	\begin{subfigure}[b]{0.33\textwidth}
		\centering
		\includegraphics[width=\textwidth]{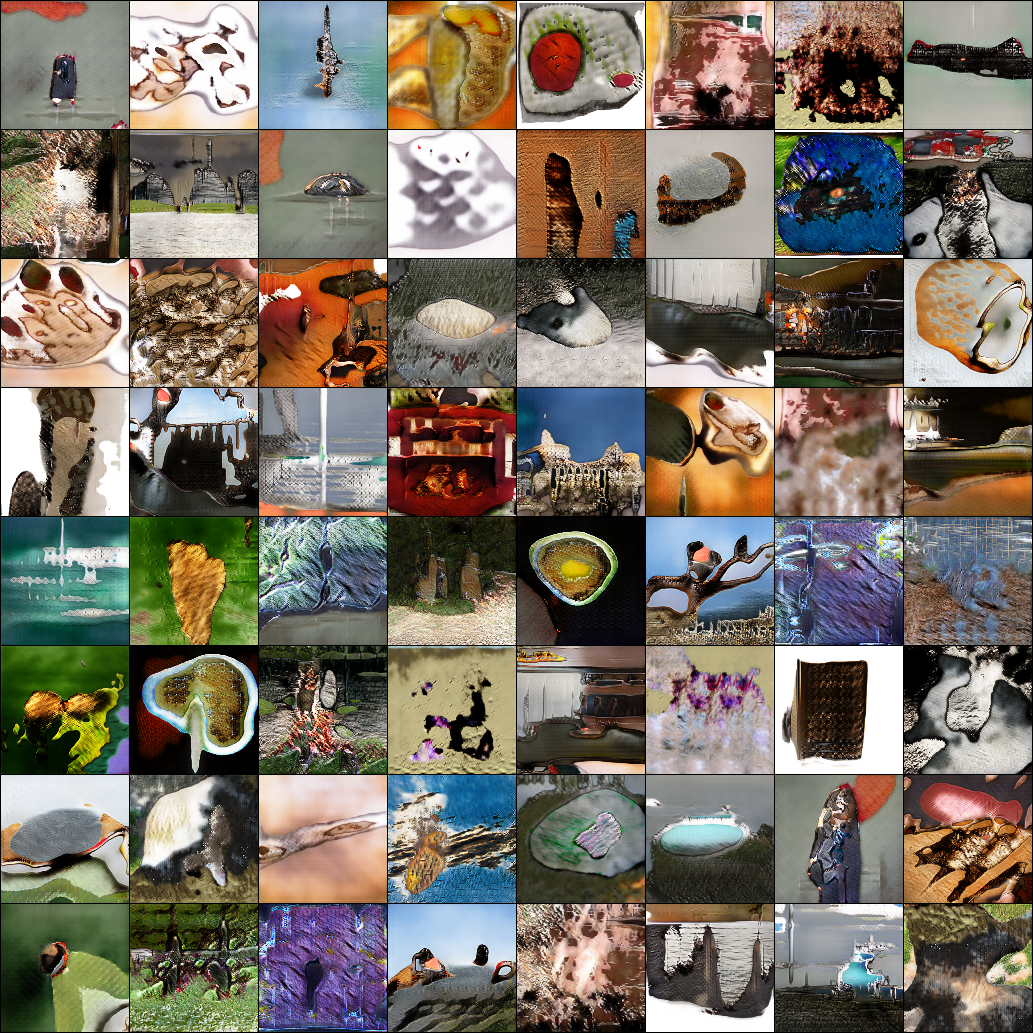}
		\caption{\footnotesize Random samples for ImageNet}
	\end{subfigure}
    \caption{\small Random samples generated by JAREs trained on CelebA and ImageNet, respectively, in an unsupervised manner.  For CelebA, the sample size is $64 \times 64$, and for ImageNet, the sample size is $128 \times 128$.}
    \label{visual_quality}
\end{figure}

\subsection{More Rresults on CelebA and ImageNet.}
In this experiment, we qualitatively evaluate the generated samples of JARE on the CelebA (with size of $64 \times 64$) \citep{liu2015faceattributes} and ILSVRC2012 (ImageNet, with size of $128 \times 128$) \citep{russakovsky2015imagenet} datasets. Due to the limitation of our computational budgets, we do not apply large hyperparameter searches. Instead, we use a similar training setup as for the CIFAR-10 experiments, with slightly different network architectures. Please see Tables \ref{resnet_celeba} and \ref{resnet_imagenet} in Appendix \ref{sec_net_arc} for details.

Figure \ref{visual_quality} (a) and (b) show the randomly generated samples of JARE trained on CelebA and ImageNet, respectively. We can see that for CelebA, JARE can produce realistic and diverse celebrity faces with various backgrounds.
For ImageNet, JARE can stabilize the training well while other training methods quickly collapse. While not completely realistic, it can generate visually convincing and diverse images from 1000 ImageNet classes in a completely unsupervised manner. The good results of JARE on CelebA and ImageNet demonstrate its ability of stabilizing the GAN training on more complex tasks.

\end{document}